\documentclass{article}

\usepackage[preprint]{neurips_2026}

\usepackage[utf8]{inputenc} % allow utf-8 input
\usepackage[T1]{fontenc}    % use 8-bit T1 fonts
\usepackage[bookmarks=false, backref=page]{hyperref} % hyperlinks
\usepackage{url}            % simple URL typesetting
\usepackage{booktabs}       % professional-quality tables
\usepackage{amsfonts}       % blackboard math symbols
\usepackage{nicefrac}       % compact symbols for 1/2, etc.
\usepackage{microtype}      % microtypography
\usepackage[table, dvipsnames]{xcolor}         % colors
\hypersetup{
  colorlinks=true,
  citecolor=NavyBlue!70!black,
  linkcolor=magenta!70!black,
}
\usepackage{amsmath}
\usepackage{array}
\usepackage{multirow}
\usepackage{subcaption}
\usepackage{float}
\usepackage{hhline}
\usepackage{textcomp}
\usepackage{tabularx}
\usepackage{microtype}

\usepackage{algorithm}
\usepackage{algorithmic}

\usepackage{tikz}
\usepackage{pgfplots}
\usepackage{pgfplotstable}
\pgfplotsset{compat=1.18}

\usetikzlibrary{shapes.geometric, arrows.meta, positioning, fit, backgrounds, calc, shadows, shapes.symbols}
\usepgfplotslibrary{groupplots, fillbetween}
\definecolor{grpBlue}{RGB}{70, 130, 180}
\definecolor{grpRed}{RGB}{205, 92, 92}
\definecolor{procGray}{RGB}{245, 245, 245}
\definecolor{jointGold}{RGB}{255, 223, 100}

\title{D5P4: Partition Determinantal Point Process for Diversity in Parallel Discrete Diffusion Decoding}

\makeatletter
\newcommand\blfootnote[1]{\g@addto@macro\@thanks{{\renewcommand\thefootnote{}\footnotetext{#1}}}}
\newcommand\pAND[1]{
\end{tabular}\\[#1]
\begin{tabular}[t]{c}\bf\rule{\z@}{0pt}\ignorespaces
}
\makeatother

\author{%
  Jonathan Lys\textsuperscript{1}\blfootnote{\textsuperscript{1}IMT Atlantique, Lab-STICC, UMR CNRS 6285, F-29238 Brest, France.
    \textsuperscript{2}Sony Europe Ltd. Stuttgart Technology Center, EUREC, Germany.
    Correspondence to: Jonathan Lys
  \texttt{<jonathan.lys@imt-atlantique.fr>}}
  \And
  Vincent Gripon\textsuperscript{1}
  \And
  Axel Marmoret\textsuperscript{1}
  \And
  Lukas Mauch\textsuperscript{2}
  \pAND{-5pt}
  Fabien Cardinaux\textsuperscript{2}
  \And
  Ghouthi Boukli Hacene\textsuperscript{2}
  \And
  Bastien Pasdeloup\textsuperscript{1}
}

\setcitestyle{numbers,square,comma,sort&compress}
\begin{document}

\definecolor{darkgray176}{RGB}{176,176,176}
\definecolor{lightgray204}{RGB}{204,204,204}
\definecolor{goldenrod1911910}{RGB}{191,191,0}
\definecolor{green01270}{RGB}{0,127,0}
\definecolor{darkorange25512714}{RGB}{255,127,14}
\definecolor{forestgreen4416044}{RGB}{44,160,44}
\definecolor{steelblue31119180}{RGB}{31,119,180}

\definecolor{darkWhite}{rgb}{0.96,0.96,0.96}
\definecolor{bluekeywords}{rgb}{0.13,0.13,1}
\definecolor{greencomments}{rgb}{0,0.5,0}
\definecolor{redstrings}{rgb}{0.9,0,0}

\definecolor{Comment}{RGB}{97,161,176}

\definecolor{btfGreen}{RGB}{51,160,44}
\definecolor{btfRed}{RGB}{190,60,90}

\definecolor{bleuUni}{RGB}{0, 157, 224}
\definecolor{marronUni}{RGB}{68, 58, 49}

\definecolor{bluecite}{HTML}{009DE0}

\definecolor{Paired-1}{RGB}{31,120,180}
\definecolor{Paired-2}{RGB}{166,206,227}
\definecolor{Paired-3}{RGB}{51,160,44}
\definecolor{Paired-4}{RGB}{178,223,138}
\definecolor{Paired-5}{RGB}{227,26,28}
\definecolor{Paired-6}{RGB}{251,154,153}
\definecolor{Paired-7}{RGB}{255,127,0}
\definecolor{Paired-8}{RGB}{253,191,111}
\definecolor{Paired-9}{RGB}{106,61,154}
\definecolor{Paired-10}{RGB}{202,178,214}
\definecolor{Paired-11}{RGB}{177,89,40}
\definecolor{Paired-12}{RGB}{255,255,153}
\definecolor{Accent-1}{RGB}{127,201,127}
\definecolor{Accent-2}{RGB}{190,174,212}
\definecolor{Accent-3}{RGB}{253,192,134}
\definecolor{Accent-4}{RGB}{255,255,153}
\definecolor{Accent-5}{RGB}{56,108,176}
\definecolor{Accent-6}{RGB}{240,2,127}
\definecolor{Accent-7}{RGB}{191,91,23}
\definecolor{Accent-8}{RGB}{102,102,102}
\definecolor{Spectral-1}{RGB}{158,1,66}
\definecolor{Spectral-2}{RGB}{213,62,79}
\definecolor{Spectral-3}{RGB}{244,109,67}
\definecolor{Spectral-4}{RGB}{253,174,97}
\definecolor{Spectral-5}{RGB}{254,224,139}
\definecolor{Spectral-6}{RGB}{255,255,191}
\definecolor{Spectral-7}{RGB}{230,245,152}
\definecolor{Spectral-8}{RGB}{171,221,164}
\definecolor{Spectral-9}{RGB}{102,194,165}
\definecolor{Spectral-10}{RGB}{50,136,189}
\definecolor{Spectral-11}{RGB}{94,79,162}
\definecolor{Set1-1}{RGB}{228,26,28}
\definecolor{Set1-2}{RGB}{55,126,184}
\definecolor{Set1-3}{RGB}{77,175,74}
\definecolor{Set1-4}{RGB}{152,78,163}
\definecolor{Set1-5}{RGB}{255,127,0}
\definecolor{Set1-6}{RGB}{255,255,51}
\definecolor{Set1-7}{RGB}{166,86,40}
\definecolor{Set1-8}{RGB}{247,129,191}
\definecolor{Set1-9}{RGB}{153,153,153}
\definecolor{Set1-10}{RGB}{0,0,0}
\definecolor{Set2-1}{RGB}{102,194,165}
\definecolor{Set2-2}{RGB}{252,141,98}
\definecolor{Set2-3}{RGB}{141,160,203}
\definecolor{Set2-4}{RGB}{231,138,195}
\definecolor{Set2-5}{RGB}{166,216,84}
\definecolor{Set2-6}{RGB}{255,217,47}
\definecolor{Set2-7}{RGB}{229,196,148}
\definecolor{Set2-8}{RGB}{179,179,179}
\definecolor{Dark2-1}{RGB}{27,158,119}
\definecolor{Dark2-2}{RGB}{217,95,2}
\definecolor{Dark2-3}{RGB}{117,112,179}
\definecolor{Dark2-4}{RGB}{231,41,138}
\definecolor{Dark2-5}{RGB}{102,166,30}
\definecolor{Dark2-6}{RGB}{230,171,2}
\definecolor{Dark2-7}{RGB}{166,118,29}
\definecolor{Dark2-8}{RGB}{102,102,102}
\definecolor{Reds-1}{RGB}{255,245,240}
\definecolor{Reds-2}{RGB}{254,224,210}
\definecolor{Reds-3}{RGB}{252,187,161}
\definecolor{Reds-4}{RGB}{252,146,114}
\definecolor{Reds-5}{RGB}{251,106,74}
\definecolor{Reds-6}{RGB}{239,59,44}
\definecolor{Reds-7}{RGB}{203,24,29}
\definecolor{Reds-8}{RGB}{165,15,21}
\definecolor{Reds-9}{RGB}{103,0,13}
\definecolor{Greens-1}{RGB}{247,252,245}
\definecolor{Greens-2}{RGB}{229,245,224}
\definecolor{Greens-3}{RGB}{199,233,192}
\definecolor{Greens-4}{RGB}{161,217,155}
\definecolor{Greens-5}{RGB}{116,196,118}
\definecolor{Greens-6}{RGB}{65,171,93}
\definecolor{Greens-7}{RGB}{35,139,69}
\definecolor{Greens-8}{RGB}{0,109,44}
\definecolor{Greens-9}{RGB}{0,68,27}
\definecolor{Blues-1}{RGB}{247,251,255}
\definecolor{Blues-2}{RGB}{222,235,247}
\definecolor{Blues-3}{RGB}{198,219,239}
\definecolor{Blues-4}{RGB}{158,202,225}
\definecolor{Blues-5}{RGB}{107,174,214}
\definecolor{Blues-6}{RGB}{66,146,198}
\definecolor{Blues-7}{RGB}{33,113,181}
\definecolor{Blues-8}{RGB}{8,81,156}
\definecolor{Blues-9}{RGB}{8,48,107}
\definecolor{mapcolor}{HTML}{1F78B4}   % Blue
\definecolor{refcolor}{HTML}{FF7F00}   % Orange
\definecolor{basecolor}{HTML}{2CA02C}  % Green

\pgfplotsset{
  discard if not/.style 2 args={
    x filter/.append code={
      \edef\tempa{\thisrow{#1}}
      \edef\tempb{#2}
      \ifx\tempa\tempb
      \else
      \def\pgfmathresult{inf}
      \fi
    }
  }
}

\pgfplotsset{
  % Hook to map the #7 argument to either title or ylabel
  setup title or ylabel/.style={title={#1}},
  base seaborn style/.style={
    grid=major,
    grid style={solid, draw=black!15},
    axis background/.style={fill=white},
    minor tick num=0,
    axis lines=box,
  },
  appendix seaborn style/.style={
    base seaborn style,
    label style={font=\scriptsize},
    tick label style={font=\scriptsize},
    legend style={font=\scriptsize, row sep=-3pt, fill opacity=0.7, draw opacity=1, text opacity=1},
    title style={font=\scriptsize, yshift=-1.2ex}, % Bring title closer
    setup title or ylabel/.style={title={##1}},
  },
  main seaborn style/.style={
    base seaborn style,
    label style={font=\small},
    tick label style={font=\small},
    legend style={font=\small, fill opacity=0.7, draw opacity=1, text opacity=1},
    ylabel style={font=\small},
    setup title or ylabel/.style={ylabel={##1}},
  }
}

% Alias for backward compatibility if needed, though we updated usage
\pgfplotsset{seaborn whitegrid/.style={appendix seaborn style}}

% Now takes 7 arguments (1 optional):
% #1: style (default: appendix seaborn style)
% #2: y-axis variable
% #3: width
% #4: height
% #5: xlabel
% #6: legend position
% #7: title/caption (used as ylabel in main paper style)
\newcommand{\createplot}[7][appendix seaborn style]{%
  \begin{tikzpicture}
    \begin{axis}[
        #1,
        width=#3,
        height=#4,
        xlabel={#5},
        setup title or ylabel={#7},
        xtick={1, 1.5, 2, 2.5, 3},
        legend pos=#6,
      ]

      \addplot[
        color=basecolor,
        mark=triangle*,
        solid,
        line width=1.0pt,
        mark options={scale=0.8, solid},
        discard if not={method}{baseline}
      ] table [x=cfg, y=#2, col sep=comma] {data/sweeps/cfg_results_all.csv};
      \addlegendentry{Baseline}

      \addplot[
        color=refcolor,
        mark=square*,
        dashed,
        line width=1.0pt,
        mark options={scale=0.8, solid},
        discard if not={method}{greedybs}
      ] table [x=cfg, y=#2, col sep=comma] {data/sweeps/cfg_results_all.csv};
      \addlegendentry{GreedyBS}

      \addplot[
        color=mapcolor,
        mark=*,
        solid,
        line width=1.0pt,
        mark options={scale=0.8, solid},
        discard if not={method}{d5p4}
      ] table [x=cfg, y=#2, col sep=comma] {data/sweeps/cfg_results_all.csv};
      \addlegendentry{D$5$P$4$}

    \end{axis}
  \end{tikzpicture}
}

\maketitle
\vspace{-1pt}

% \begin{abstract}
%   Discrete diffusion models are promising alternatives to autoregressive approaches for text generation, yet their decoding methods remain
%   under-studied. Standard decoding methods for autoregressive models, such as beam search, do not directly apply to iterative denoising,
%   and existing diffusion decoding techniques provide limited control over in-batch diversity. To bridge this gap, we introduce a
%   generalized beam-search framework for discrete diffusion that generates candidates in parallel and supports modular beam-selection
%   objectives. As a diversity-focused instantiation, we propose \textbf{D5P4}, which formulates the selection step as MAP inference over a
%   Determinantal Point Process. Leveraging a scalable greedy solver, D5P4 enables an explicit
%   trade-off between model probability and target diversity with near-zero compute overhead. Experiments on free-form generation and
%   question answering demonstrate that D5P4 improves diversity over strong baselines while maintaining competitive generation
%   quality.\footnote{\href{https://anonymous.4open.science/r/d5p4}{https://anonymous.4open.science/r/d5p4/}}
% \end{abstract}

\begin{abstract}
  Discrete diffusion models are promising alternatives to autoregressive approaches for text generation, yet their decoding methods remain
  under-studied.
  Standard autoregressive search procedures, such as beam search, do not directly apply to iterative denoising, where hypotheses are
  complete intermediate sequences rather than left-to-right prefixes.
  Furthermore, existing diffusion decoding procedures only provide limited control over the diversity and coverage of retained hypotheses.
  In this work, we introduce \textbf{D5P4}, a beam-style decoding method tailored to discrete diffusion models, which casts intermediate
  beam selection as MAP inference under a partitioned Determinantal Point Process.
  This yields a model-internal batch objective that balances quality and diversity without external verifiers.
  Experiments on open-ended generation, question answering, and mathematical reasoning show that D5P4 improves diversity and pass@$k$
  coverage while matching or surpassing baseline quality and fidelity{\renewcommand{\thefootnote}{\textasteriskcentered}\footnote{Code:
      \href{https://anonymous.4open.science/r/d5p4}{https://anonymous.4open.science/r/d5p4/}.
  This work used GENCI-IDRIS resources (2025-AD011015252R1)}}.
  D5P4 also improves decoding efficiency, reducing post-forward latency and peak memory while scaling to large candidate pools.
\end{abstract}

\section{Introduction}

Discrete diffusion models have recently emerged as competitive alternatives to autoregressive language models (ARMs), achieving strong
performance on a range of text generation tasks~\cite{sahoo2024simple, nie2025large, ye2025dream}. By refining sequences in parallel through
iterative denoising, they depart from traditional left-to-right decoding but introduce a fundamental structural challenge: the need to
coordinate updates across the entire sequence simultaneously. While ARMs benefit from mature search algorithms like beam search to maintain
high-quality outputs~\cite{Lowerre1976Harpy}, these procedures do not directly transfer to diffusion models, where hypotheses are not partial
prefixes but complete intermediate states.

This algorithmic gap is compounded by the increasing importance of diversity and coverage in modern generation. Recent evidence suggests
that while reinforcement learning and supervised fine-tuning can improve top-1 performance, they often saturate coverage (pass@$k$) and
reduce solution diversity~\cite{yue_does_2025}. Similarly, in conditioned diffusion, strong guidance sharpens fidelity at the expense of sample
variety~\cite{sadat2024cads}. Together, these observations suggest that a search algorithm should not choose candidates solely by independent
sequence scores and instead account for the collective composition of the set to ensure the efficient exploration of the solution space.

To address these challenges, we introduce \emph{Partition Determinantal Point Processes for Diversity in Parallel Discrete Diffusion Decoding}
(\textbf{D5P4}), a beam-style decoding method tailored to the iterative structure of discrete diffusion. Our approach equips parallel
denoising with step-wise structured set selection over the intermediate sequences. By utilizing a Determinantal Point Process (DPP)
kernel, D5P4 combines quality signals from diffusion logits with diversity signals from model-internal hidden representations. Rather
than directly sampling from the DPP, we perform approximate Maximum A Posteriori (MAP) inference over this kernel using a scalable
fast greedy solver to select high-quality and diverse beams. We complement this selection procedure with a structural partition constraint on
candidate selection that explicitly prevents lineage collapse, a phenomenon in which hypotheses degenerate toward a single ancestry, as
identified in diverse beam search~\cite{vijayakumar2016diverse}.

This work moves beyond independent sampling to provide a principled paradigm for structured search in non-sequential models. We validate
this pipeline across diverse tasks and datasets, demonstrating that D5P4 improves diversity and coverage (pass@$k$) in open-ended
generation, question answering, and mathematical reasoning. Notably, our method matches or outperforms compute-matched baselines that rely
on external models, despite D5P4 being fully model-internal and avoiding auxiliary scorers. Scalability analysis further confirms the
efficiency of our framework in end-to-end decoding pipelines, where D5P4 achieves measurable speed and memory improvements over existing baselines.

\section{Related Work}
\subsection{Discrete diffusion language models}
Discrete diffusion models originate from early formulations of diffusion processes over discrete variables~\citep{sohl2015deep} and
categorical spaces via argmax-based flows~\citep{hoogeboom2021argmax}.
D3PM~\citep{austin2021structured} formalized these as structured Markov processes, unifying training via an ELBO objective and
accommodating diverse corruption mechanisms (e.g., uniform, absorbing). Absorbing-state diffusion underpins masked diffusion language
models (MDLMs), which leverage continuous diffusion advances~\citep{ho2020denoising} and Transformer
architectures~\citep{peebles2023scalable}. Recent MDLMs, including simplified recipes~\citep{sahoo2024simple}, scaled
variants~\citep{von2025scaling,nie2024scaling,ye2025dream}, and instruction-following models like LLaDA~\citep{nie2025large}, now approach
autoregressive performance.

A defining trait of MDLMs is parallel decoding, where tokens are refined jointly. While this
decouples decoding depth from sequence length, naive parallel refinement often degrades generation quality, as token updates fail to
explicitly model inter-token dependencies. As a result, prior work typically frames parallel decoding as a speed–quality trade-off. Several
approaches accelerate inference through speculative execution or diffusion-specific key–value
caching~\citep{wu2025fastdllmtrainingfreeaccelerationdiffusion,israel2025accelerating,ma2025dkv,wang2025diffusion}, but largely focus on
throughput, not joint quality or diversity across parallel hypotheses.
\subsection{Sampling and selection for text generation}
In autoregressive models, decoding can be interpreted as approximate search over full sequences. Beam search~\cite{Lowerre1976Harpy}
maintains multiple partial hypotheses, but shared prefixes often lead to rapid collapse into a single ancestral path. Nucleus
sampling~\citep{holtzman2020curious} promotes diversity via stochastic truncation, yet lacks explicit coordination between
hypotheses. Other methods introduce diversity at selection time, including Diverse Beam Search~\citep{vijayakumar2016diverse}, stochastic
beam search with Gumbel-Top-$k$ sampling~\citep{kool2019stochastic}, and reranking approaches such as Maximal Marginal
Relevance~\citep{mmr98}, which explicitly penalize similarity among outputs.
% Recent work on test-time compute scaling reinforces a ``generate many, then select'' paradigm, showing that performance can improve
% substantially by selecting from large candidate pools~\citep{beeching2024scaling}. For example, \citet{kang_scalable_2025} propose
% self-certainty as a lightweight selection criterion that does not rely on external verifiers. In the context of discrete diffusion,
% \citet{dang2025inference} introduce Particle Gibbs sampling for diffusion language models, enabling reward-guided inference via
% MCMC/SMC-style resampling over full denoising trajectories. However, this approach does not model interactions between candidate solutions
% and is therefore not directly comparable to our method. While an adaptation to incorporate such interactions may be possible, the resulting
% procedure would incur substantial additional test-time cost, scaling with the number of particles, diffusion steps, sampling iterations,
% and repeated reward evaluations. As a consequence, any direct quantitative comparison would conflate fundamentally different compute
% regimes and optimization objectives.
Recent studies on test-time scaling demonstrate that performance improves by selecting from large candidate
pools~\citep{beeching2024scaling,kang_scalable_2025}. A related line of work uses particle-based inference to steer generation toward
sequence-level targets. Twisted SMC decodes from an unnormalized target induced by a potential in ARMs~\citep{zhao2024probabilistic}, while
Feynman--Kac and SMC-based methods apply similar weighting and resampling ideas through rewards, guidance likelihoods, or importance
weights~\citep{uehara2025inference,skreta2025feynmankac,kit2025debiasing}. Closest to our work, PG-DLM~\citep{dang2025inference} employs
reward-guided MCMC/SMC resampling over full trajectories for discrete diffusion models. However, because it operates at the trajectory
level and requires multiple refinement passes, it is not directly comparable in terms of computational overhead. Crucially, while these
methods optimize for independent sequence-level targets, D5P4 introduces a \emph{batch-level} objective where candidate selection also
depends on the global composition of the set. This allows us to explicitly select a diverse ensemble of high-quality hypotheses, trading
off individual sequence-level quality for batch-level diversity.
% This allows us to explicitly select a diverse ensemble of high-quality hypotheses, leveraging batch-level diversity to maximize the
% exploration of the solution space and significantly boost coverage (pass@k) without increasing the total compute budget

\subsection{Diversity-aware decoding}
Recent analyses of reinforcement learning and supervised fine-tuning (SFT) in ARMs~\cite{dang_assessing_2025,yue_does_2025} observe a
consistent reduction in output diversity, particularly in reasoning-focused settings. GEM~\cite{li2025preserving} study this phenomenon in the
context of SFT and propose a mitigation strategy at training time that improves coverage and benefits test-time scaling.

These findings suggest that gains in alignment and fidelity are not necessarily accompanied by broader coverage, further motivating
explicit diversity-aware decoding mechanisms.

A similar issue arises in continuous diffusion, where strong classifier-free guidance (CFG) is known to induce mode collapse, an effect that
also appears in discrete diffusion for text~\cite{schiff2024simple}. Existing strategies for recovering diversity in continuous diffusion
typically act by modifying the sampling dynamics, for example through noise injection~\cite{sadat2024cads} or partial guidance
schedules~\cite{Kynkaanniemi2024}. These methods improve the diversity--quality trade-off, but they do not optimize diversity among the
retained hypotheses.

Most closely related to our selection objective, Determinantal Beam Search~\cite{meister2021determinantal} formulates beam selection using
DPPs, encouraging diverse yet high-scoring hypotheses via a log-determinant objective and a string-based similarity measure. However, this
method is inherently autoregressive. D5P4 instead adapts the DPP modeling approach to parallel discrete
diffusion by selecting among complete intermediate denoising states, using model-internal representations and MAP inference.

\section{Methods}

\subsection{Discrete Diffusion Preliminaries}\label{subsec:preliminary}

We briefly review the discrete diffusion inference framework underlying MDLM~\cite{sahoo2024simple}
and LLaDA~\cite{nie2025large}, introducing the notation needed for our decoding method.
Let $\mathcal{V}$ denote a vocabulary of tokens represented as one-hot vectors, including a special
mask token $\mathbf{m}\in\mathcal{V}$. A sequence of length $L$ is denoted
$\mathbf{x}\in\mathcal{V}^L$. The diffusion process operates on latent variables
$\mathbf{z}_t\in\mathcal{V}^L$, indexed by a continuous time $t\in[0,1]$, with
$\mathbf{z}_1=\mathbf{m}^L$ a fully masked sequence and $\mathbf{z}_0=\mathbf{x}_0$ the clean data.
A noise schedule $\alpha_t\in[0,1]$ controls the proportion of unmasked tokens along the trajectory.

At inference time, both MDLM and LLaDA start from a fully masked sequence and iteratively refine it.
For a transition from timestep $t$ to $s$, with $0\le s<t\le 1$, the model produces logits
$p_\theta(\cdot\mid \mathbf{z}_t)$ over $\mathcal{V}^L$, and the next state is obtained via a
projection operator
\begin{equation}
  \mathbf{z}_s = \Pi_{t,s}\big(p_\theta(\cdot \mid \mathbf{z}_t)\big) \;.
  \label{eq:logits2seq}
\end{equation}
The operator $\Pi_{t,s}$ encapsulates the model-specific sampling and re-masking rule. LLaDA samples
a fully denoised sequence and then re-masks a fixed number of tokens, selected either uniformly, by
low confidence, or via a hybrid sampling strategy (tempered confidence). In contrast, MDLM enforces the
target masking ratio in logit space. In instruction-following settings, LLaDA further excludes prompt
tokens from the remasking budget while keeping them in the conditioning context.

This unified inference view is sufficient for our purposes: at each denoising step, the decoder has
access to candidate logits and hidden representations, on top of which we perform structured
selection. Additional background on discrete diffusion training is deferred to
Appendix~\ref{app:diffusion_prelim}.

\subsection{Beam-Style Decoding for Discrete Diffusion}\label{subsec:beam_diffusion}

While parallelism enables scalable sampling, it does not by itself address the intractability of sequence-level search. Effective
approximations therefore require maintaining and selectively refining a limited set of high-quality hypotheses. We thus formulate a
beam-style decoding approach for discrete diffusion, in which parallel sampling is structured through intermediate selection steps.

\begin{figure}[t!]
  \centering
  \resizebox{\linewidth}{!}{
    \begin{tikzpicture}[
    %x=1.8cm, y=0.8cm,
    x=2.4cm, y=0.8cm,
    >=Latex,
    % Node Styles
    seq/.style={
      rectangle,
      draw=black!60,
      rounded corners=2pt,
      minimum width=1.2cm,
      minimum height=0.5cm,
      align=center,
      fill=white,
      font=\scriptsize,
      drop shadow={opacity=0.3},
      thin
    },
    op_circle/.style={
      circle,
      draw=black!50,
      fill=procGray,
      minimum size=0.6cm,
      inner sep=0pt,
      font=\tiny\bfseries
    },
    joint_block/.style={
      rectangle,
      draw=black!60,
      fill=jointGold!30,
      rounded corners=4pt,
      minimum width=1.5cm,
      align=center,
      thick,
    },
    arrow_style/.style={
      ->,
      thick,
      black!60
    }
  ]

  % Separation Line
  \draw[dashed, black!20] (-1, 0) -- (5, 0) node[pos=0.9, above, font=\scriptsize, text=black!40] {Lineage independence};

  % ==========================================
  % STAGE 1: INPUT SEQUENCES (Step t)
  % ==========================================

  % --- Group 1 (Blue) ---
  % Highlight Index 2 (Middle) as winner
  \node[seq, fill=grpBlue!30, draw=grpBlue, thick] (g1_1) at (0, 3) {Seq $\mathbf{z}_{1,1}$};
  \node[seq, fill=grpBlue!10] (g1_2) at (0, 2) {Seq $\mathbf{z}_{1,2}$};
  \node[seq, fill=grpBlue!10] (g1_3) at (0, 1) {Seq $\mathbf{z}_{1,3}$};

  % --- Group 2 (Red) ---
  % Highlight Index 3 (Bottom) as winner
  \node[seq, fill=grpRed!10] (g2_1) at (0, -1) {Seq $\mathbf{z}_{2,1}$};
  \node[seq, fill=grpRed!10] (g2_2) at (0, -2) {Seq $\mathbf{z}_{2,2}$};
  \node[seq, fill=grpRed!30, draw=grpRed, thick] (g2_3) at (0, -3) {Seq $\mathbf{z}_{2,3}$};

  % Labels
  \node[above=0.2cm of g1_1] {Step $t$};

  % FIXED LABEL POSITIONING:
  % We use 'anchor=south' on a 90-degree rotated node.
  % Visually, 'south' is the right side of the text (the baseline), facing the node.
  % We position this relative to the .west anchor of the middle node.
  \node[rotate=90, anchor=south, font=\tiny, text=grpBlue] at ($(g1_2.west) + (-0.15, 0)$) {Group 1};

  % Centered relative to g2_2 (middle of group 2)
  \node[rotate=90, anchor=south, font=\tiny, text=grpRed] at ($(g2_2.west) + (-0.15, 0)$) {Group 2};

  % ==========================================
  % STAGE 2: MODEL PASS (p_theta)
  % ==========================================

  \foreach \n in {g1_1, g1_2, g1_3, g2_1, g2_2, g2_3} {
    \node[op_circle, right=1cm of \n] (mod_\n) {$p_\theta$};
    \draw[arrow_style] (\n) -- (mod_\n);
  }

  % ==========================================
  % STAGE 3: JOINT SCORING BLOCK
  % ==========================================

  % Calculate center for the large block
  \coordinate (top_mod) at (mod_g1_1.north);
  \coordinate (bot_mod) at (mod_g2_3.south);
  \path (top_mod) -- (bot_mod) coordinate[midway] (mid_block);

  \node[joint_block, right=1.2cm of mid_block, minimum height=6cm, anchor=west] (joint) {Joint\\Scoring\\and\\Groupwise\\Selection\\via DPP\\objective};
  % \node[joint_block, right=1.5cm of mid_block, minimum height=6cm, anchor=west] (joint) {\rotatebox{90}{\shortstack[c]{Joint Scoring\\\&\\Groupwise Selection}}};

  % Wiring inputs to Joint Block
  \foreach \n in {g1_1, g1_2, g1_3, g2_1, g2_2, g2_3} {
    \draw[arrow_style] (mod_\n) -- (mod_\n -| joint.west);
  }

  % ==========================================
  % STAGE 4: WINNERS & EXPANSION
  % ==========================================

  % Winner 1 (From Blue Group, Index 2)
  \node[seq, fill=grpBlue!30, draw=grpBlue, thick, right=1.2cm of joint.east |- g1_2] (win1) {Seq $\mathbf{z}_{1,1}$};

  % Winner 2 (From Red Group, Index 3) - Aligned with bottom row
  \node[seq, fill=grpRed!30, draw=grpRed, thick, right=1.2cm of joint.east |- g2_2] (win2) {Seq $\mathbf{z}_{2,3}$};

  % Arrows from Joint Block
  \draw[arrow_style, grpBlue, dashed] (joint.east |- g1_1) -- (win1.west);
  \draw[arrow_style, grpRed, dashed] (joint.east |- g2_3) -- (win2.west);

  % --- EXPANSION (Branching) ---
  \node[op_circle, right=0.6cm of win1] (exp1) {$\Pi$};
  \node[op_circle, right=0.6cm of win2] (exp2) {$\Pi$};

  \draw[arrow_style] (win1) -- (exp1);
  \draw[arrow_style] (win2) -- (exp2);

  % ==========================================
  % STAGE 5: NEXT GENERATION (Step t-1)
  % ==========================================

  % Group 1 expanded
  \node[seq, fill=grpBlue!20, right=0.8cm of exp1] (next1_2) {Seq $\mathbf{z}_{1,2}'$};
  \node[seq, fill=grpBlue!20, above=0.1cm of next1_2] (next1_1) {Seq $\mathbf{z}_{1,1}'$};
  \node[seq, fill=grpBlue!20, below=0.1cm of next1_2] (next1_3) {Seq $\mathbf{z}_{1,3}'$};

  % Group 2 expanded
  \node[seq, fill=grpRed!20, right=0.8cm of exp2] (next2_2) {Seq $\mathbf{z}_{2,2}'$};
  \node[seq, fill=grpRed!20, above=0.1cm of next2_2] (next2_1) {Seq $\mathbf{z}_{2,1}'$};
  \node[seq, fill=grpRed!20, below=0.1cm of next2_2] (next2_3) {Seq $\mathbf{z}_{2,3}'$};

  % Connect Expander to Next Gen
  \foreach \n in {next1_1, next1_2, next1_3} \draw[arrow_style, thin] (exp1) -- (\n.west);
  \foreach \n in {next2_1, next2_2, next2_3} \draw[arrow_style, thin] (exp2) -- (\n.west);

  \node[above=0.2cm of next1_1] {Step $s$};

  % ==========================================
  % ANNOTATIONS
  % ==========================================

  % Highlight logic text
  \node[font=\tiny, text=grpBlue!80!black, align=center, above=0.1cm of win1] {Selected 1};
  \node[font=\tiny, text=grpRed!80!black, align=center, below=0.1cm of win2] {Selected 2};

\end{tikzpicture}
  }
  \caption{
    Overview of our diffusion search algorithm. $n$ partially denoised sequences at time $t$ are fed to the model $p_\theta$, which outputs
    logits and embeddings for the Joint Scoring \& Groupwise Selection block. The logits of the $k$ jointly-selected sequences in each
    group are then expanded through $w$ independent applications of the projection operator $\Pi$, producing $n$ sequences at time $s$.
  }
  \label{fig:diff_bs}
\end{figure}

\paragraph{Branching and Scoring.} Let $k$ denote the number of retained beams and $w$ the branching factor, yielding a candidate pool of
size $n = k \cdot w$. At each diffusion step $t$, we select $k$ beams and generate $w$ descendants per beam by $w$ independent applications
of the stochastic projection operator $\Pi_{t,s}$ to the denoising logits $p_\theta(\cdot \mid \mathbf{z}_t)$. Candidates are evaluated
via a scoring function $q: \mathcal{V}^L \rightarrow \mathbb{R}^+$. As discrete diffusion lacks monotonic prefix
likelihoods, we utilize sequence-level normalized negative entropy as a proxy for generative quality, a choice validated by our ablation studies
(Appendix~\ref{app:ablations}). This yields a quality score vector $\mathbf{Q} \in \left(\mathbb{R}^+\right)^n$ for the candidate pool.

\paragraph{Transversal Partition.} Selecting the top-$k$ candidates based solely on $\mathbf{Q}$ can lead to ancestral collapse, in which
near-duplicate sequences are selected, thereby underutilizing the available parallel budget~\citep{vijayakumar2016diverse}. To mitigate
this issue, we impose a \textit{transversal partition} constraint. Specifically, the $n$ candidates are partitioned into $k$ groups, each
with the $w$ descendants of a common parent beam. We ablate the effect of this constraint in Appendix~\ref{app:partition-ablation}.

\subsection{DPP Modeling and Greedy MAP Inference}\label{subsec:dpp}

While the transversal partition ensures that retained beams originate from distinct parents, a simple selection strategy that chooses the
highest-scoring candidate within each group (transversal greedy beam search, or per-group argmax) remains agnostic to interactions between
candidates across groups.

To further control redundancy across groups, we score the selected set jointly using a Determinantal Point Process (DPP) kernel.
This modeling choice provides a principled set-level objective that favors high-quality and mutually dissimilar
candidates~\cite{kulesza2012determinantal}. The resulting discrete diffusion beam-search procedure is illustrated in Figure~\ref{fig:diff_bs}.

A DPP defines a distribution over subsets $S \subseteq \{1,\dots,n\}$ of the candidate pool:
\begin{equation}\label{eq:dpp_prop}
  \mathbb{P}(S) \propto \det(\mathbf{L}_S) \;,
\end{equation}
where $\mathbf{L} \in \mathbb{R}^{n \times n}$ is a positive semidefinite kernel and $\mathbf{L}_S$ is the principal submatrix indexed by
$S$. Intuitively, the determinant penalizes redundancy: if two candidates are very similar, their corresponding rows and columns in
$\mathbf{L}_S$ become nearly linearly dependent, which decreases $\det(\mathbf{L}_S)$. We develop this intuition and provide a local
approximation showing that maximization of the DPP objective induces pairwise cosine minimization in Appendix~\ref{app:dpp-theory}.

\paragraph{Kernel construction.}

We build the $\mathbf{L}$-kernel directly from diffusion model outputs. Sequence-level quality scores are derived from the token logits,
while pairwise interactions are computed from flattened hidden representations immediately before the unembedding layer. These design choices
are validated in Appendix~\ref{app:ablations}. We consider the following two parameterizations:
\begin{equation}
  \mathbf{L} =
  \begin{cases}
    \mathrm{diag}(\mathbf{Q}) + \beta \boldsymbol{\mathcal{K}},
    & \text{\scriptsize(additive)} \\
    \mathrm{diag}(e^{\mathbf{Q}/\beta}) \, \boldsymbol{\mathcal{K}} \, \mathrm{diag}(e^{\mathbf{Q}/\beta})
    & \text{\scriptsize(multiplicative)}
  \end{cases} \;,
  \label{eq:variants}
\end{equation}
where $\mathbf{Q}$ contains the quality scores, $\boldsymbol{\mathcal{K}}$ encodes pairwise interactions, and $\beta$
controls the quality--diversity trade-off. The multiplicative form is the standard DPP parameterization, the additive form is closer
to~\cite{meister2021determinantal}.
In practice, $\boldsymbol{\mathcal{K}}$ is a cosine kernel: $\boldsymbol{\mathcal{K}}_{ij} = \langle \mathbf{e}_i, \mathbf{e}_j \rangle$.
Since it is a Gram matrix and our quality score is positive, both formulations guarantee that the resulting DPP kernel is PSD.

\paragraph{Greedy MAP inference.}\label{subsec:greedy_map}

Our goal is to identify the most representative mode of the set-level distribution at each denoising step.
We formulate selection as MAP inference, which provides a stable, deterministic criterion that avoids the trajectory
drift inherent in stochastic sampling. This formulation natively enforces the transversal partition, which is not directly supported by
standard eigendecomposition, while remaining highly scalable. We optimize the following objective:
\begin{equation}
  S^\star = \arg\max_{S \in \mathcal{T}_k} \det(\mathbf{L}_S) \;,
\end{equation}
where $\mathcal{T}_k$ denotes the family of size-$k$ transversal subsets, \emph{i.e.}, subsets containing exactly one element from each
group. This recovers standard top-$k$ beam selection as a degenerate quality-only case with diagonal $\mathbf{L}$ (no interaction term),
but more generally optimizes quality and diversity at the set level.

The resulting combinatorial problem is NP-hard, so we use the fast greedy MAP algorithm of~\cite{chen_fast_2018}. Consequently, D5P4 avoids
the $O(n^3)$ cost of traditional DPP sampling~\cite{kulesza2012determinantal} in favor of an efficient search tailored to iterative
refinement. We adapt the algorithm to the partition-constrained setting and add a multi-initialization strategy: the solver is initialized
from the groupwise argmax of each parent group, and all initializations are explored in parallel on the GPU. The original greedy solver has
complexity $\mathcal{O}(k^2 n)$; our partition-aware multi-start variant has complexity $\mathcal{O}(k^3 n)$. In practice, however, this
additional work is parallelized and remains highly efficient, as shown in Section~\ref{sec:scalability}.

The combination of the discrete diffusion decoding framework, the set selection strategy using
the partition-DPP kernel, and the greedy MAP solver constitutes \textbf{D5P4}.

\section{Experiments and Results}

The detailed configuration used in each experiment can be found in Section~\ref{subsec:exp-conf} of the Appendix. We first note that MDLM
can only be used for open-ended generation, not for conditional settings, such as the question answering and mathematical reasoning.
We therefore use LLaDA in those settings.

% \paragraph{Metrics.} We evaluate our methods across several dimensions: generation quality, task-specific accuracy, and diversity, both at
% the batch and sequence levels. We use reference-less and reference-based metrics, depending on the task. For open-ended generation
% quality, we rely on \textbf{perplexity} and distribution-level similarity (\textbf{MAUVE}). For reference-based accuracy, we measure
% performance via corpus-level \textbf{BLEU} and sequence-level \textbf{F1}. Finally, generation diversity is quantified across all
% tasks using \textbf{self-cosine similarity}, \textbf{self-BLEU}, and n-gram uniqueness (\textbf{Distinct-n} and its length-robust variant
% \textbf{EAD}). Coverage is measured with \textbf{pass@k} for math benchmarks and \textbf{F1@k} for question answering. A comprehensive
% description of all metrics is provided in Section~\ref{app:metrics} of the Appendix.

\paragraph{Metrics.} We evaluate our methods across several complementary dimensions.
For open-ended generation quality, we rely on \textbf{perplexity} and distribution fidelity (\textbf{MAUVE}).
When references are available, we use average \textbf{F1} to check for deviation from the target task distribution.
At the batch level, we evaluate the coverage by quantifying the presence of a correct solution within the set of retained proposals through
\textbf{pass@$\mathbf{k}$} in mathematical reasoning and \textbf{max F1} in question answering.
Batch diversity is quantified semantically using \textbf{self-cosine similarity} with the Jina Embeddings v2
model~\cite{günther2023jina} and lexically via \textbf{self-BLEU} and $n$-gram uniqueness
(\textbf{Distinct-$\mathbf{n}$} and its length-robust variant \textbf{EAD}). A more detailed analysis of the metrics is provided in
Section~\ref{app:metrics} of the Appendix.

\paragraph{Baselines.}
We distinguish two comparison regimes, both ensuring a strictly compute-matched (diffusion model forward passes) and proposal-matched (final
output pool size) comparison.

The first is the \emph{search} regime, where all methods perform beam-style decoding with the same intermediate candidate budget,
but differ in the subset-selection rule used to retain candidates. This includes our DPP-based objective, \textbf{D5P4}, as well as selection
objectives such as \textbf{Greedy Beam Search}, which keeps the highest-scoring candidates independently, and a partition-constrained
variant of \textbf{Diverse Beam Search}, implemented as transversal MMR.
The second regime is the \emph{$k$-best of $N$} (\textbf{BoN}) regime, where $N$ independent sequences are generated first and then
ranked to select the final $k$ outputs. This regime decouples generation from selection and includes three distinct scoring levels.
Perplexity BoN uses the perplexity of the final LLaMA 3 model, providing a high-performance signal but making the baseline
significantly more expensive and not self-contained. Internal BoN uses the internal score at the last step as a quality measure. It serves
as a proxy for the perplexity BoN but allows for a fair comparison with the search methods. Lastly, an F1-BoN utilizes the oracle F1-score and
provides an upper bound of achievable performance from the whole candidate pool. Precise definitions of all baselines and configurations
are provided in Appendix~\ref{app:baselines}.

\subsection{Open-Ended Generation}\label{subsec:open-ended-gen}

\begin{figure}[ht!]
  \centering
  \includegraphics[width=\linewidth]{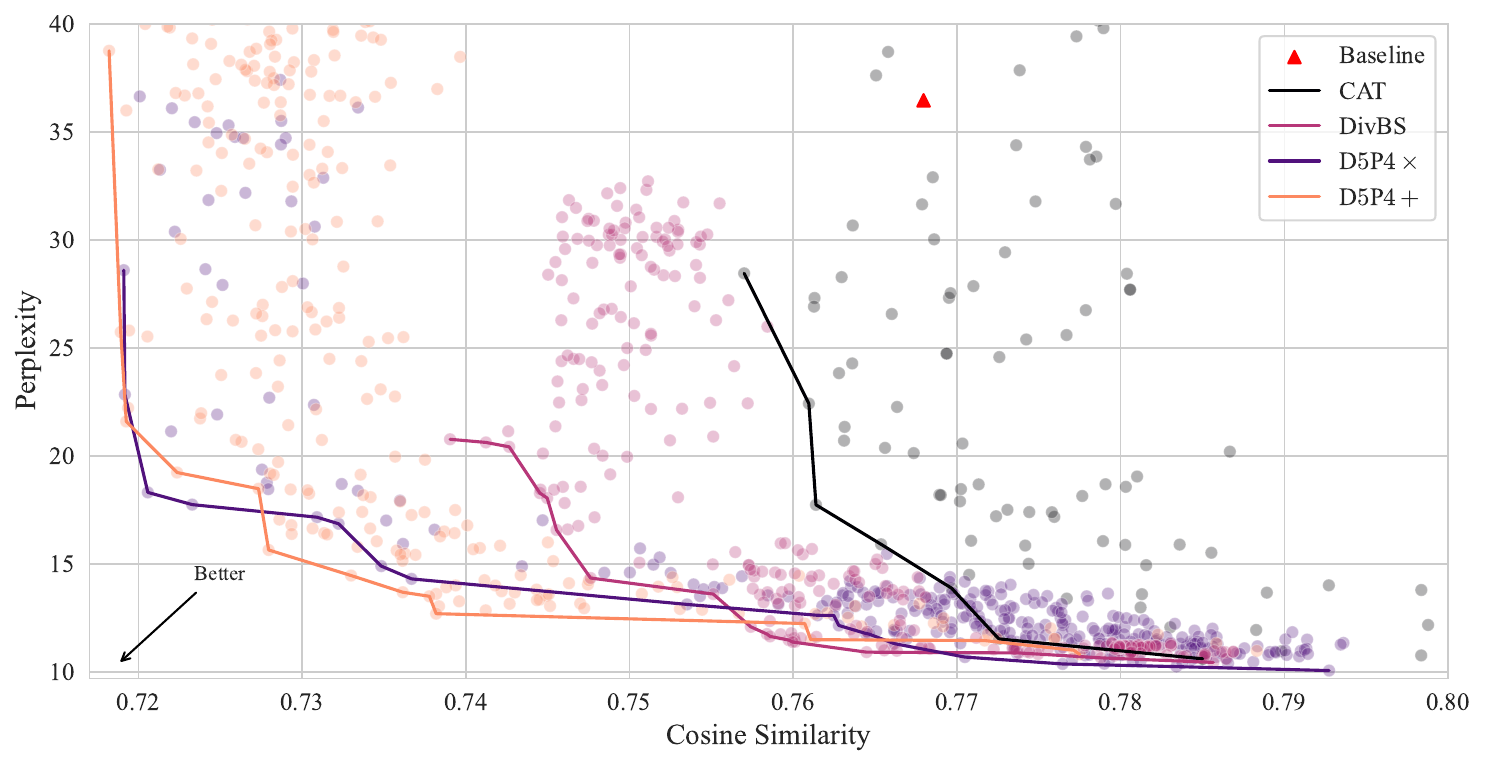}
  \caption{Quality--diversity trade-off in open-ended MDLM generation. Lower perplexity indicates higher quality, and lower cosine
    similarity indicates higher diversity. D5P4 achieves the most favorable Pareto fronts and delays the onset of quality collapse under
  increasing diversity pressure.}
  \label{fig:pareto}
\end{figure}

We first consider open-ended generation with MDLM~\cite{sahoo2024simple}. Figure~\ref{fig:pareto} compares D5P4, in both its additive
(D5P4$+$) and multiplicative (D5P4$\times$) variants (see Equation~\ref{eq:variants}), against independent generation of $k$ sequences
(Baseline), beam-search with tempered
categorical sampling (CAT), and Diverse Beam Search (DivBS), which is implemented as a partition-constrained MMR search. For each method,
we systematically sweep its diversity-control parameter:
temperature for CAT, diversity penalty $\alpha_{\text{div}}$ for DivBS, and interaction strength $\beta$ for D5P4.
Crucially, when these respective parameters are set to zero, all evaluated search methods reduce to standard greedy beam search.

Several trends emerge from Figure~\ref{fig:pareto}. First, all search-based methods outperform the independent baseline, achieving lower perplexity
and lower cosine similarity, showing that even lightweight search improves the diversity-quality trade-off over naive sampling.

More importantly, all methods exhibit the same qualitative two-regime behavior. As the diversity parameter is increased, there is first a
favorable regime in which cosine similarity decreases while perplexity remains stable. Beyond a method-dependent threshold,
perplexity rises sharply, indicating that the diversity objective has become too strong, pushing decoding toward low-likelihood regions.

This transition occurs earliest for the temperature-based method (CAT), whose quality deteriorates abruptly once its limited control range
is exceeded. DivBS follows, with a smoother but still earlier breakdown than D5P4. Both D5P4$\times$ and D5P4$+$ postpone this collapse the
furthest and achieve the best Pareto fronts, with similar breakdown locations. Overall, D5P4 provides the most controlled
diversity-quality trade-off, with later and more predictable failure modes than CAT or DivBS.

Figure~\ref{fig:control_metrics} further characterizes the effect of the diversity-control parameter through two complementary evaluations.
In the left panel (Figure~\ref{fig:mauve_inter}), we measure D5P4$+$'s distribution-level fidelity across varying values of the interaction parameter
$\beta$, using MAUVE and MAUVE* with FineWeb samples~\cite{penedo2024finewebdatasetsdecantingweb} as the reference distribution.
In the right panel (Figure~\ref{tab:method_corr}), we quantify how strongly the control parameter influences the end metrics by reporting
correlations between the control parameters and the end metrics across all methods, including the D5P4 variants.

\begin{figure}[ht!]
  \centering
  \begin{subfigure}[b]{0.52\linewidth}
    \centering
    \begin{tikzpicture}
  \pgfplotsset{set layers}

  \begin{axis}[
      width=\linewidth,
      height=0.7\linewidth,
      axis lines=left,
      x axis line style={->},
      grid=none,
      xlabel={$\beta$},
      xlabel style={at={(ticklabel* cs:1)}, anchor=north west},
      xmode=log,
      log basis x=10,
      xtick={1,10,100,1000},
      xmin=.5, xmax=2000,
      minor xtick={},
      tick align=outside,
      tick pos=left,
      tick pos=bottom,
      legend style={
        at={(0.5, 0.97)},
        anchor=north,
        font=\scriptsize,
        fill=white,
        fill opacity=0.9,
        text opacity=1,
      },
      legend columns=2,
      label style={font=\footnotesize},
      tick label style={font=\footnotesize},
      ymin=0.85,
      ymax=1.03,
      ytick={0.85, 0.90, 0.95, 1.00}, % stop at 1.
    ]

    % Baseline mauve
    \addplot[
      Paired-1,
      dashed,
      thick,
      domain=1:1000,
      forget plot,
    ] {0.8824};

    % Baseline mauve*
    \addplot[
      Paired-7,
      dashed,
      thick,
      domain=1:1000,
      forget plot,
    ] {0.9134};

    \addlegendimage{Paired-1, dashed, thick}
    \addlegendentry{Ref. MAUVE~~}
    \addlegendimage{Paired-7, dashed, thick}
    \addlegendentry{Ref. MAUVE*}

    % mauve scores
    \addplot[
      Paired-1,
      thick,
      mark=*,
      mark size=1.5pt,
    ] coordinates {
      (1, 0.8669)
      (3, 0.8811)
      (10, 0.9195)
      (30, 0.9366)
      (100, 0.9515)
      (300, 0.9474)
      (1000, 0.8746)
    };
    \addlegendentry{MAUVE}

    % mauve* scores
    \addplot[
      Paired-7,
      thick,
      mark=*,
      mark size=1.5pt,
    ] coordinates {
      (1, 0.9024)
      (3, 0.8900)
      (10, 0.9302)
      (30, 0.9502)
      (100, 0.9615)
      (300, 0.9574)
      (1000, 0.9078)
    };
    \addlegendentry{MAUVE*}

  \end{axis}
\end{tikzpicture}
    \caption{Distribution-level fidelity (MAUVE)}
    \label{fig:mauve_inter}
  \end{subfigure}
  \hfill
  \begin{subfigure}[b]{0.44\linewidth}
    \centering
    \renewcommand{\arraystretch}{1.3}
    \resizebox{\linewidth}{!}{
      \begin{tabular}{cccc}
        \multirow{2}{*}{Method} & \multirow{2}{*}{Parameter} & \multicolumn{2}{c}{Correlation} \\
        &                            & PPL & COS \\
        \toprule
        CAT                     & $\log$ temp.         & 0.974 & -0.795 \\
        DivBS                   & $\log \alpha_{div}$        & 0.944 & -0.570 \\
        \midrule
        D5P4$+$                  & \multirow{2}{*}{$\log \beta_{inter}$} & 0.970 & \textbf{-0.925} \\
        D5P4$\times$             &                                       & \textbf{-0.976} & 0.866 \\
        \bottomrule
      \end{tabular}
    }
    \vspace{0.3cm} % compensate for the figure bottom margin
    \caption{Parameter controllability}
    \label{tab:method_corr}
  \end{subfigure}
  \caption{Effect of the diversity-control parameter on generation fidelity. (Left) Distribution-level fidelity, measured by MAUVE and
    MAUVE* as a function of the D5P4 interaction parameter $\beta$, showing that an intermediate regime yields the best fidelity. (Right)
    Correlation between each method's control parameter and the resulting quality--diversity metrics. D5P4 shows the strongest overall
  coupling between its control parameter and the end metrics.}
  \label{fig:control_metrics}
\end{figure}

The results reveal two consistent patterns. First, MAUVE and MAUVE* are maximized in an intermediate regime of $\beta$. Values that are too
small or too large fall below the baseline, whereas an intermediate range improves upon it, indicating an optimal operating regime for more
sample-efficient modeling of the target distribution. Second, the control parameter is strongly correlated with both perplexity and cosine
similarity across all methods, confirming that it provides a meaningful handle on the quality--diversity trade-off. These correlations are
strongest for D5P4, especially with respect to diversity, indicating that its interaction parameter induces a particularly direct and
monotonic response. We note that D5P4$+$ has superior control over the cosine similarity than D5P4$\times$. For this reason, we focus
on D5P4$+$ for the remainder of the paper.

Together, these results indicate that the diversity parameter does not merely affect individual end metrics, but directly shapes the
fidelity of the generated distribution. In particular, D5P4 exhibits a particularly strong and monotonic response to its control parameter,
and admits a favorable intermediate regime in which quality and diversity are balanced most effectively.

\subsection{Question Answering}

We next evaluate D5P4 on question answering with LLaDA. We compare two best-of-N (BoN) variants, an internal
entropy-based selector and an external perplexity-based selector, against Diverse Beam Search, D5P4, and an augmented version, D5P4$^P$. The
latter applies classifier-free guidance only during the first half of the denoising process to improve sample diversity, a strategy inspired
by partial CFG schedules in continuous diffusion~\citep{Kynkaanniemi2024}. Notably, the perplexity BoN requires an additional forward pass
with the autoregressive model, making it computationally more expensive and difficult to compare directly to the model-internal methods.
We report quality, coverage, and diversity metrics in Table~\ref{tab:qa-LLaDA} on the TruthfulQA~\cite{lin2021truthfulqa} and Arc
Challenge~\cite{arch:challenge} datasets. Comprehensive results, including performance on CommonSenseQA, are provided in
Appendix~\ref{app:qa-LLaDA-full}.

\begin{table}[ht!]
  \centering
  % \caption{Question answering performance comparison on TruthfulQA and Arc Challenge. \textbf{BN-P}: PPL best-of-$n$
  %   (privileged reference since it requires an additional forward pass with an external AR model), \textbf{BN-I}: internal entropy-based
  % best-of-$n$,
  %   \textbf{DBS}: Diverse Beam Search, \textbf{D5P4$^P$}: D5P4 with partial CFG. \textbf{S-COS} and \textbf{S-BLEU} denote the self-cosine
  %   similarity and self-BLEU, respectively. \textbf{D1-3} is the average of Distinct-1/2/3. \textbf{SDE} denotes the average standard error
  % across all methods. Bold and underline indicate the best and second-best results among model-internal (non-privileged) methods.}
  \caption{Question answering performance on TruthfulQA and Arc-Challenge. \colorbox{gray!10}{\textbf{BN-P}}: Perplexity best-of-$n$
    (privileged reference using an external AR model for selection); \textbf{BN-I}: internal entropy-based best-of-$n$; \textbf{DBS}:
    Diverse Beam Search; \textbf{D5P4$^P$}: D5P4 with partial CFG. Diversity is measured via \textbf{S-COS} (self-cosine similarity),
    \textbf{S-BLEU} (self-BLEU), and \textbf{D1-3} (average Distinct-1/2/3). \textbf{SDE} denotes the average standard error across all
  methods. \textbf{Bold} and \underline{underline} indicate the best and second-best results among \textbf{model-internal} methods.}
  \resizebox{\linewidth}{!}{\newcommand{\intermethod}{12pt}
\setlength{\tabcolsep}{4pt}
\newcolumntype{C}{>{\centering\arraybackslash}p{1.3cm}}
\renewcommand{\arraystretch}{1.3}

\begin{tabular}{ l | >{\columncolor{gray!10}}c c c c c | c || >{\columncolor{gray!10}}c c c c c | c }
  \hline
  \multirow{2}{*}{\rule{0pt}{3ex}\textbf{Metric}} & \multicolumn{6}{c||}{\textbf{TruthfulQA}}  & \multicolumn{6}{c}{\textbf{ARC-Challenge}}  \\
  \hhline{~|-----|-||-----|-|}
  \rule{0pt}{3ex} & BN-P & BN-I & DBS & D5P4 & D5P4$^P$ & SDE & BN-P & BN-I & DBS & D5P4 & D5P4$^P$ & SDE \rule[-1.5ex]{0pt}{0pt} \\
  \hline \hline

  %\multicolumn{15}{l}{\textit{Quality}} \\
  \textbf{Quality/Fidelity} & & & & & & & & & & & & \\
  $\downarrow$ PPL & 5.24 & 6.64 & \textbf{6.37} & \underline{6.40} & 6.72 & $\pm$0.14 & 4.93 & 5.86 & \underline{5.78} & \textbf{5.78} & 5.82 & $\pm$0.07 \\
  $\uparrow$ F1 (\%) & 17.31 & \underline{19.39} & 19.07 & 19.02 & \textbf{20.71} & $\pm$0.17 & 5.01 & \underline{5.35} & 5.22 & 5.24 & \textbf{5.46} & $\pm$0.06 \\
  \hline

  \textbf{Coverage} & & & & & & & & & & & & \\
  $\uparrow$ max F1 (\%) & 19.16 & 22.00 & 22.55 & \underline{22.71} & \textbf{24.84} & $\pm$0.33 & 5.85 & 6.38 & 6.36 & \underline{6.42} & \textbf{6.74} & $\pm$0.12 \\
  \hline

  \textbf{Diversity} & & & & & & & & & & & & \\
  $\downarrow$ S-COS (\%) & 96.40 & 96.12 & 95.36 & \underline{95.16} & \textbf{94.88} & $\pm$0.07 & 97.65 & 97.50 & 97.00 & \underline{96.87} & \textbf{96.76} & $\pm$0.03 \\
  $\downarrow$ S-BLEU (\%) & 46.27 & 45.57 & \underline{40.74} & \textbf{39.00} & 41.64 & $\pm$0.39 & 55.04 & 53.54 & \underline{49.03} & \textbf{47.71} & 49.70 & $\pm$0.28 \\
  $\uparrow$ D1-3 (\%) & 53.54 & 54.66 & \underline{57.00} & \textbf{57.54} & 56.37 & $\pm$0.20 & 47.42 & 48.82 & \underline{50.86} & \textbf{51.30} & 50.10 & $\pm$0.16 \\
  $\uparrow$ EAD (\%) & 34.57 & 36.03 & 37.52 & \textbf{37.78} & \underline{37.53} & $\pm$0.13 & 30.17 & 31.39 & \underline{32.56} & \textbf{32.78} & 32.24 & $\pm$0.11 \\
  \hline
\end{tabular}

}
  \label{tab:qa-LLaDA}
\end{table}

Across all datasets, D5P4 and its variants achieve better quality and diversity performance compared to both iterative search and post-hoc
selection baselines. Notably, our methods outperform both the model-internal (\textbf{BN-I}) and privileged perplexity (\textbf{BN-P})
best-of-$n$ references on all but one accuracy/diversity metric, trailing only on the specific perplexity signal utilized by
\textbf{BN-P} for selection. While Diverse Beam Search (\textbf{DBS}) also employs a partition constraint, D5P4 provides significant and
consistent gains in coverage and intra-batch diversity. Mechanically, the DPP objective acts as a unified selection criterion that promotes
both semantic (cosine similarity) and lexical diversity (self-BLEU, Distinct-n, and EAD), which directly translates into the improved
solution discovery observed in the max F1 results. Finally, the augmented \textbf{D5P4$^P$} further pushes semantic alignment and coverage,
though it involves a minor trade-off in lexical diversity compared to the base model.

We further investigate a regime where diversity control is critical: high classifier-free guidance (CFG), which causes diversity
degradation in diffusion models~\citep{sadat2024cads, schiff2024simple}. Increasing CFG with LLaDA on TruthfulQA shows
that higher guidance consistently erodes diversity (Figure~\ref{fig:cfg_cosine}). Notably, Appendix~\ref{app:cfg} shows that best-of-$n$
methods fail to alleviate this collapse, providing lower accuracy and diversity than proposal-matched independent sampling. Greedy Beam
Search (GBS) maintains competitive accuracy but yields the lowest diversity, failing to match even independent samples.

In contrast, D5P4 preserves substantially higher diversity across all metrics while matching the generation quality and accuracy of GBS.
This demonstrates that structured decoding effectively decouples guidance-induced alignment from mode collapse, allowing for diverse high-fidelity
generation. Ultimately, these results confirm the importance of a set-level DPP objective in alleviating CFG-induced collapse,
outperforming both single-objective search and post-hoc selection.

\begin{figure}[ht!]
  \centering
  \begin{tikzpicture}
  \begin{axis}[
      main seaborn style,
      width=\linewidth,
      height=0.25\linewidth,
      xlabel={CFG},
      setup title or ylabel={$\Delta$ COS (vs IND)},
      xtick={1, 1.5, 2, 2.5, 3},
      ymin=-0.015, ymax=0.015,
      restrict y to domain*=-0.015:0.015,
      ymajorgrids=true,
      extra y ticks={0},
      extra y tick style={grid=major, grid style={dashed, black}},
      legend style={
        at={(0.5,1.05)},
        anchor=south,
        legend columns=-1,
        font=\scriptsize,
        draw=none,
        /tikz/every even column/.append style={column sep=5pt}
      },
    ]

    \addplot[
      color=basecolor,
      mark=triangle*,
      solid,
      line width=1.0pt,
      mark options={scale=0.8, solid},
    ] coordinates {
      (1.0, 0.0) (1.5, 0.0) (2.0, 0.0) (2.5, 0.0) (3.0, 0.0)
    };
    \addlegendentry{Indep.}

    \addplot[
      color=refcolor,
      mark=square*,
      dashed,
      line width=1.0pt,
      mark options={scale=0.8, solid},
    ] coordinates {
      (1.0, 0.032909) (1.5, 0.002248) (2.0, 0.002939) (2.5, 0.003065) (3.0, 0.000868)
    };
    \addlegendentry{Greedy Beam}

    \addplot[
      color=black!60,
      mark=diamond*,
      densely dotted,
      line width=1.0pt,
      mark options={scale=0.8, solid},
    ] coordinates {
      (1.0, 0.015027) (1.5, 0.005550) (2.0, 0.004404) (2.5, 0.003997) (3.0, 0.002279)
    };
    \addlegendentry{PPL BoN}

    \addplot[
      color=mapcolor,
      mark=*,
      solid,
      line width=1.0pt,
      mark options={scale=0.8, solid},
    ] coordinates {
      (1.0, -0.003140) (1.5, -0.012122) (2.0, -0.007765) (2.5, -0.005547) (3.0, -0.006712)
    };
    \addlegendentry{D5P4}

  \end{axis}
\end{tikzpicture}
  %\caption{Mitigation of CFG diversity collapse. $\Delta$ cosine similarity vs. independent sampling.}
  \caption{Mitigation of CFG diversity collapse. We report the difference in self-cosine similarity against
  proposal-matched independent samples. Lower values indicate higher diversity.}
  \label{fig:cfg_cosine}
\end{figure}

\subsection{Mathematical Reasoning (GSM8K)}

We next evaluate D5P4 on the GSM8K~\cite{cobbe2021gsm8k} benchmark, using pass@$k$ to measure solution accuracy and
coverage. Table~\ref{tab:gsm8k} compares D5P4 against independent sampling, a perplexity-based (using an external LLaMA
model) and entropy-based best-of-$N$, Greedy and Diverse beam searches, under confidence and stochastic
remasking (from~\cite{chen2025optimizing}). Full results are provided in Section~\ref{app:gsm8k_full}.

\begin{table}[h!]
  \centering
  \caption{Mathematical reasoning performance on GSM8K (compute budget: $n=16$, proposal budget: $k=4$).
    \colorbox{gray!10}{\textbf{BN-P}}: Perplexity best-of-$n$
    (privileged reference); \textbf{BN-I}: internal-entropy best-of-$n$; \textbf{IND}:
    proposal-matched independent sampling; \textbf{GBS}: Greedy Beam Search; \textbf{DBS}: Diverse Beam Search.
  \textbf{Bold} and \underline{underline} indicate the best and second-best results among \textbf{model-internal} methods.}
  \resizebox{\linewidth}{!}{\setlength{\tabcolsep}{4pt}
\renewcommand{\arraystretch}{1.3}

\begin{tabular}{ l | >{\columncolor{gray!10}}c c c c c c || >{\columncolor{gray!10}}c c c c c c }
  \hline
  \multirow{2}{*}{\rule{0pt}{3ex}\textbf{Metric}} & \multicolumn{6}{c||}{\textbf{Confidence}}  & \multicolumn{6}{c}{\textbf{Stochastic}}  \\
  \hhline{~|------||------|}
  \rule{0pt}{3ex} & BN-P & BN-I & IND & GBS & DBS & D5P4 & BN-P & BN-I & IND & GBS & DBS & D5P4 \rule[-1.5ex]{0pt}{0pt} \\
  \hline \hline
  \rule{0pt}{4ex}$\uparrow$ Pass@1 & 46.5 & \underline{44.4} & \textbf{44.5} & \underline{44.4} & \underline{44.4} & \underline{44.4} & 51.8 & 46.8 & 45.5 & \underline{47.3} & \textbf{47.4} & \underline{47.3} \\
  $\uparrow$ Pass@2 & 51.6 & 49.6 & 49.9 & 50.8 & \underline{51.9} & \textbf{52.5} & 63.6 & 59.7 & 58.8 & 58.9 & \underline{60.7} & \textbf{61.2} \\
  $\uparrow$ Pass@4 & 56.3 & 54.0 & 54.5 & 56.3 & \underline{58.2} & \textbf{59.3} & 72.6 & 70.1 & 69.4 & 68.3 & \underline{71.2} & \textbf{72.2} \rule[-2ex]{0pt}{0pt} \\
  \hline
  \rule{0pt}{4ex}$\downarrow$ S-BLEU & 87.3 & 85.8 & 85.2 & 83.4 & \underline{80.8} & \textbf{79.1} & 72.7 & 68.3 & \underline{66.7} & 73.4 & 67.5 & \textbf{65.2} \\
  \hline
\end{tabular}

}
  \label{tab:gsm8k}
\end{table}

D5P4 demonstrates a fundamental trade-off advantage, maintaining high individual quality while significantly boosting set coverage. While
average accuracy (pass@1) remains on par with or superior to internal baselines in the confidence setting, the primary strength of D5P4
lies in its improved coverage as $k$ increases. D5P4 excels at maximizing pass@4, the primary goal for this setting. It outperforms all
internal baselines, trailing only the external PPL selector in the stochastic regime. Notably, that selector uses a LLaMA 3 model with high
baseline accuracy on this benchmark; D5P4 approaches this performance using only internal scores. Crucially, D5P4 consistently yields the
lowest Self-BLEU scores, confirming that its coverage gains are driven by genuine diversity. These results show that D5P4 effectively
explores unique candidate solutions, ensuring the decoded set contains a broader range of reasoning paths without sacrificing individual
sample quality.

\section{Scalability and Computational Efficiency}\label{sec:scalability}

We finally assess D5P4's computational efficiency from two perspectives. Table~\ref{tab:scalability_combined} (left) reports
end-to-end decoding measurements on LLaDA at the largest supported batch size (6 groups $\times$ 10 candidates per
GPU, 2-H200 node), which captures the practical effect of D5P4 inside the full distributed pipeline. On the right, we isolate selection
costs by comparing algorithms at a high-complexity operating point from the synthetic scaling study (Appendix~\ref{app:scaling}), excluding
model forward passes.

\begin{table}[ht!]
  \centering
  \caption{Scalability and efficiency analysis on LLaDA. (Left) End-to-end decoding performance: total time per step (forward pass
    $\approx$ 929ms) and peak GPU memory at maximum supported batch size (6 groups of 10 per GPU). (Right) Runtime and raw
    subdeterminant value comparison of subset selection algorithms on synthetic DPP kernels (32 groups $\times$ 32). (*) DPP sampling is not
  transversal.}
  \label{tab:scalability_combined}
  \vspace{1em}
  \begin{minipage}[c]{0.63\linewidth}
    \centering
    {\footnotesize \textbf{End-to-End Decoding Performance}} \\
    \vspace{2pt}
    \resizebox{\linewidth}{!}{
      \begin{tabular}{lccc|c}
        \toprule
        Method & Selection & Cat. samp. & Total & Mem (GB) \\
        \midrule
        Baseline & \textbf{22.7ms} & 86.1ms & 108.8ms & 132.8 \\
        DivBS & 58.7ms & \textbf{9.4ms} & 68.1ms & \textbf{79.3}\\
        D5P4 & 46.6ms & \textbf{9.5ms} & \textbf{56.1ms} & \textbf{79.3} \\
        \bottomrule
      \end{tabular}
    }
  \end{minipage}
  \hfill
  \begin{minipage}[c]{0.35\linewidth}
    \centering
    {\footnotesize \textbf{Search Efficiency}} \\
    \vspace{2pt}
    \resizebox{\linewidth}{!}{
      \begin{tabular}{lcc}
        \toprule
        Method & Sel. time & Raw val. \\
        \midrule
        Random & 0.1ms & -25.465 \\ %-0.907 \\
        DPP (*) & 547.8ms & -25.453 \\ %-0.875 \\
        DivBS & 29.5ms & -24.625 \\ %0.665 \\
        D5P4 & \textbf{2.3ms} & \textbf{-24.483} \\ %1.021 \\
        \bottomrule
      \end{tabular}
    }
  \end{minipage}
\end{table}

% In the end-to-end setting, D5P4 leaves model forward latency essentially unchanged. Although the selection step is slower than the
% baseline, this cost is more than offset by a large reduction in categorical sampling time, reducing the total post-forward overhead. D5P4
% also substantially lowers peak memory usage, from 132.8\,GB to 79.3\,GB.
% Relative to DivBS, D5P4 achieves comparable end-to-end behavior while using a faster selector.
% The synthetic benchmark leads to the same conclusion from the search side. At the representative 32 $\times$ 32 point, D5P4 achieves both
% the best normalized objective and the lowest selection time among the non-trivial diversity-aware methods, outperforming DivBS while
% remaining far more practical than the CPU-based DPP reference. Combined with the full trends reported in Appendix~\ref{app:scaling}, these
% results indicate that D5P4 remains efficient even in large-batch, high-complexity decoding regimes.

In the end-to-end setting, D5P4 achieves superior hardware efficiency through its branching mechanism. While the selection step introduces
marginal overhead, this is fundamentally compensated by a significant (nearly $10\times$) reduction in categorical sampling time. By
shifting the $n$ sampling operations into $w$ independent samples from $k$ sources, the method drastically minimizes the computational
burden typically associated with high-dimensional distribution projections. This also results in a 40\% reduction in peak memory usage, as
the memory footprint required to materialize categorical probability tensors is significantly curtailed.

The synthetic benchmark (Table~\ref{tab:scalability_combined}, right) further distinguishes D5P4's algorithmic efficiency. At the
high-complexity $32 \times 32$ operating point, D5P4 achieves the highest raw objective value while being over $12\times$ faster than
Diverse Beam Search and orders of magnitude faster than the exact DPP reference. Coupled with the scaling trends in
Appendix~\ref{app:scaling}, these results demonstrate that D5P4 provides a highly efficient, GPU-resident solution for diversity-aware
decoding that remains practical even in high-throughput, large-batch regimes.

\section{Conclusion}

% In this work, we introduce D5P4, a decoding framework for discrete diffusion language models that couples parallel denoising with
% principled, diversity-aware selection. D5P4 casts beam selection as MAP inference in a Partition Determinantal Point Process, yielding a
% simple and interpretable mechanism for trading off model-implied quality and in-batch diversity. Crucially, both signals are computed from
% the diffusion model itself (sequence-level entropy for quality and hidden-state representations for semantic similarity) to avoid reliance
% on costly external scorers at inference time.

% Across open-ended generation with MDLM and question answering with LLaDA, D5P4 improves diversity without sacrificing competitive quality,
% and consistently provides a stronger quality-diversity Pareto trade-off than widely used alternatives such as diverse beam search and
% temperature-based sampling. These gains come with minimal overhead: our greedy MAP solver is efficient, multi-GPU friendly, and preserves
% the throughput advantages of parallel decoding. More broadly, D5P4 suggests a practical route to test-time scaling for diffusion LMs that
% increases coverage through structured candidate selection rather than additional model calls. Future work could extend this principle to
% other diffusion formulations and modalities, and explore richer kernel designs or task-aware objectives while retaining the same efficient
% selection backbone.

In this work, we introduced D5P4, a principled decoding framework for discrete diffusion language models that addresses the critical gap
between parallel denoising and structured sequence search. By casting beam selection as MAP inference within a partition Determinantal
Point Process, we provide a mechanism to balance model-implied quality with batch-level diversity. A key advantage of D5P4 is its
self-contained nature; it derives both quality signals and diversity signals directly from the diffusion model, eliminating the need for
expensive external scorers during inference.

Our experiments on open-ended generation show that D5P4 achieves a superior quality-diversity Pareto front compared to diverse beam search
and stochastic sampling. On question answering and mathematical reasoning benchmarks (TruthfulQA, ARC-Challenge, CommonSenseQA, and GSM8K),
D5P4 provides critical improvements in coverage and variety. These gains are supported by high efficiency: our method reduces post-forward
latency and peak memory in end-to-end pipelines and outperforms existing selection baselines on large-scale synthetic kernels.

% Broadly, D5P4 suggests a practical route to test-time scaling for diffusion LLMs that increases coverage through structured candidate
% selection. Future research may explore extending these principles to discrete image or multimodal models, while investigating task-specific
% kernel designs to further refine the balance between alignment and exploration. We believe this approach paves the way for more robust and
% controllable non-autoregressive text generation.

Broadly, our results suggest that test-time scaling for diffusion LLMs should not only generate more candidates, but also preserve
meaningful variation among them throughout denoising. This matters beyond benchmark performance: diversity-aware decoding can help reduce
mode collapse and output homogenization, especially when generated data are reused for self-training or distillation, where loss of coverage
may be amplified over successive generations. D5P4 is one instantiation of this principle, using model-internal signals to promote coverage
without external verifiers, while remaining limited by the reliability of these proxies and by the choice of kernel and diversity scale.
Future work may explore task-aware kernels and extensions to image or multimodal discrete diffusion models.

\bibliography{references}

\newpage
\appendix

\section{Discrete Diffusion Preliminaries}
\label{app:diffusion_prelim}

This section complements Section~\ref{subsec:preliminary} with additional background on the discrete
diffusion formulation used by MDLM~\cite{sahoo2024simple} and LLaDA~\cite{nie2025large}.

\paragraph{Training objective.}
Let $p_\theta(\mathbf{x}\mid \mathbf{z}_t,t)$ denote the predicted categorical distribution over clean
tokens conditioned on the noisy state $\mathbf{z}_t$. Following \citet{sahoo2024simple}, the
Rao--Blackwellized variational bound yields a simplified objective depending only on the transition
rates. Under the commonly used linear schedule $\alpha_t = 1-t$, the loss takes the form:
\begin{equation}
  \mathcal{L}(\theta)
  = -\mathbb{E}_{t,\mathbf{x},\mathbf{z}_t}
  \left[
    \frac{1}{t}
    \sum_{i:\mathbf{z}_t^i=\mathbf{m}}
    \log p_\theta(x^i\mid \mathbf{z}_t,t)
  \right] \;.
\end{equation}
This objective is thus identical to that of LLaDA, unifying the training objectives of the two models.

\paragraph{Inference dynamics.}
At inference time, both MDLM and LLaDA start from a fully masked sequence
$\mathbf{z}_1=\mathbf{m}^L$ and iteratively denoise it.\\
For a transition from timestep $t$ to $s$, with $0\le s<t\le 1$, the next state is obtained as:
\begin{equation}
  \mathbf{z}_s = \Pi_{t,s}\big(p_\theta(\cdot \mid \mathbf{z}_t)\big) \;.
\end{equation}
The projection operator $\Pi_{t,s}$ captures the model-specific sampling and re-masking strategy.

In LLaDA, one first samples a fully denoised sequence from the logits and then re-masks a fixed
number of positions, chosen either uniformly, by low-confidence token prediction, or by a mixture of the two.
Notably, this specific strategy is adapted from E-SMC. It samples the unmasking positions from
the tempered top confidence values. In the zero-temperature limit, it is equivalent to sampling from the argmax
positions (confidence), while in the infinite-temperature limit, it is equivalent to uniform sampling over those positions.
In MDLM, the masking ratio is enforced directly in logit space so that the expected number of masked tokens matches the target schedule. In
instruction-following mode, LLaDA excludes prompt tokens from the remasking budget, although they remain part of the conditioning context.

These implementation differences do not affect our decoding formulation. D5P4 only assumes that, at each denoising step, the model provides
a batched set of candidate logits and hidden representations from which we construct the selection objective. In both models, the hidden
representations used to construct the selection kernel consist of the flattened sequence embeddings extracted immediately before the final
Transformer unembedding layer.

\section{Experimental details}

We present all experimental details: model configuration, metrics, datasets and baselines.

\subsection{Detailed Experimental Setup}\label{subsec:exp-conf}

To ensure a fair comparison, we use a controlled compute budget within each experiment and keep the proposal budget constant across
compared methods. All search-based methods operate on the same grouped decoding structure, so differences reflect the selection strategy
rather than additional model evaluations. In practice, groups can be distributed across GPUs or nodes to preserve parallel execution and
accelerate inference. This does not change the end results.

We adopt standard diffusion schedules: the number of refinement steps is set equal to the sequence length for open-ended generation and to
the target answer length for question-answering tasks.

In the open-ended generation experiments (Figure~\ref{fig:pareto}), we use $4$ elements per group and $8$ groups, yielding a total batch
size of $32$. Each reported point corresponds to $8$ independent runs, while the independent baseline aggregates results over $100$ runs.
The MAUVE study (Figure~\ref{fig:mauve_inter}) uses the same configuration.
In these settings, we maintain a sequence length of $1024$ tokens. In the question answering experiments, we evaluate on three benchmarks:
TruthfulQA~\cite{lin2021truthfulqa}, ARC-Challenge~\cite{arch:challenge}, and CommonSenseQA~\cite{talmor_commonsenseqa_2019}. All QA
results are obtained with a global batch size of $9$, organized as $3$ groups of $3$ candidates each, and an answer length set to 128
tokens. For mathematical reasoning, we use $4$ groups of $4$ elements each with an answer length of 256 tokens.

All LLaDA experiments utilize the ``pure diffusion'' regime, where the sequence length matches the block size. We fix the number of denoising
steps to equal the sequence length, resulting in a schedule that unmasks exactly one token per step. For MDLM, we follow the standard
schedule, ensuring the expected masking ratio aligns with the target linear schedule. Following the results of Figure~\ref{fig:pareto},
and apart from the additional math experiment, where the temperature is set to zero, we keep the categorical sampling with $T=1$, as higher
values lead to very high perplexity values.

\subsection{Metrics}\label{app:metrics}

We add details to the definition of the metrics used in the paper and to the additional ones used in the exhaustive results that follow.

\paragraph{Quality/Fidelity Metrics.}
We report \textbf{Perplexity (PPL)}, which measures sequence likelihood under an external autoregressive evaluator (GPT-2 Large for MDLM
and Llama-3 for LLaDA). We additionally report \textbf{Empirical Entropy}. This metric is computed over the token space by replacing model
probabilities with empirical frequencies, i.e., each token probability is estimated as its frequency divided by the total number of
generated tokens in the evaluated sample set. It therefore measures the effective support of the generated token distribution.

In open-ended generation, we report \textbf{MAUVE} and \textbf{MAUVE*}, which quantify distribution-level similarity between generated and
reference text.
The default setting from~\cite{pillutla_mauve_2021} uses a GPT2 model to compute features.
In question answering, we report cosine similarity with the reference answers (\textbf{COS}), with a Jina
embeddings v2 model~\cite{günther2023jina}. This model follows a BERT-style architecture, supports long input
sequences (up to 8192 tokens), and is designed for general-purpose text embeddings for long documents.

\paragraph{Accuracy/Exact Match.}
For question answering, we evaluate correctness using \textbf{BLEU} and token-level \textbf{F1-score} to check for deviation from the
target distribution. We note, however, that exact-match-style metrics are not equally informative across datasets. In particular,
\textbf{CommonSenseQA} is used primarily as a sanity check, since its answers are typically very short; in this regime, BLEU and
token-level F1 have limited resolution and should be interpreted with caution relative to the more semantically informative alignment and
diversity metrics.

\paragraph{Coverage as Batch-Max Performance.}
We define coverage metrics as the \textbf{maximum performance achieved across a batch} of $k$ samples, which evaluates the model's ability
to explore the solution space effectively. In mathematics benchmarks, \textbf{pass@$k$} is treated as an indicator metric that is $1$ if at
least one candidate in the batch is correct, and $0$ otherwise. We interpret this specifically as the \textit{maximum binary accuracy}
across the batch.

For soft-match tasks such as question answering, we generalize the pass@$k$ logic by taking the maximum score achieved within the batch:
\begin{equation}
  \text{Metric@}k = \max_{i \in \{1, \dots, k\}} \text{Metric}(s_i, \text{ref})
\end{equation}
where each sample $s_i$ is compared against the reference. This ``best-of-$k$'' approach measures the quality of the most accurate
hypothesis the model is capable of generating within a given budget; specifically, we report maximum F1 (\textbf{F1@$k$}) and maximum BLEU
(\textbf{BLEU@$k$}).

Importantly, when evaluating these metrics over overcomplete sets (e.g., computing pass@2 from a pool of 4 samples), the value is computed
as an average over all possible combinations of size $k$. In this regime, the metric represents the probability of drawing at least one
correct sample (or the expected maximum score) when randomly choosing $k$ items from the pool of candidates. Consequently, \textbf{pass@1}
is equivalent to the average accuracy across all generated samples.

\paragraph{Diversity Metrics.}
We report the average \textbf{in-batch cosine similarity} using embeddings from the Jina v2 model, where lower similarity
indicates higher semantic diversity. To measure lexical diversity, we report \textbf{Self-BLEU}, \textbf{Distinct-$n$}, and \textbf{EAD}
(Expectation-Adjusted Distinct). Note that self-BLEU and self-cosine are inter-sample metrics, while distinct-$n$ and EAD are intra-sample metrics.

% We evaluate reference-based question answering on three complementary benchmarks: TruthfulQA, ARC-C, and CommonSenseQA.

% \paragraph{TruthfulQA.}
% TruthfulQA is designed to measure whether language models reproduce common misconceptions or instead generate truthful answers. We use it
% as a short-form generative question answering benchmark and evaluate both answer quality and diversity across the generated candidate set.

% \paragraph{ARC-C.}
% ARC-C is the challenge split of the AI2 Reasoning Challenge and contains science questions with higher reasoning difficulty than the easy
% split. We include ARC-C to complement the other QA benchmarks with a setting that places greater emphasis on knowledge-intensive and
% multi-step reasoning under the same decoding pipeline.

% \paragraph{CommonSenseQA.}
% CommonSenseQA targets everyday commonsense reasoning. It complements TruthfulQA and ARC-C by focusing on plausible world knowledge and
% common sense rather than factual truthfulness alone.

% Note that we include CommonSenseQA primarily as a sanity-check benchmark for short-form commonsense answering. Because the reference answers are
% typically very short, exact-match-style metrics such as BLEU and token-level F1 are less expressive than on longer-form generative QA
% benchmarks. We therefore interpret this dataset mainly through consistency with the broader trends in alignment and diversity, rather than
% through small absolute differences in exact-match scores alone.

\subsection{Datasets}

We evaluate reference-based question answering across three benchmarks that represent distinct challenges for discrete diffusion models:
truthfulness, logical reasoning, and commonsense knowledge.

\paragraph{TruthfulQA~\cite{lin2021truthfulqa}.}
TruthfulQA is an adversarial benchmark designed to evaluate whether models reproduce human misconceptions. We
utilize the generative task to measure the model's ability to maintain \textbf{factual truthfulness} while exploring the generation space.
Unlike standard QA, this setting requires navigating away from high-likelihood but false attractors in the distribution.

\paragraph{ARC-Challenge~\cite{arch:challenge}.}
The ARC-C split consists of science questions that require \textbf{multi-step reasoning} and external knowledge
retrieval. We include ARC-C to test the robustness of our decoding pipeline in knowledge-intensive settings where the semantic overlap
between a "correct" answer and a "plausible" one is often narrow, placing a higher premium on precision in the solution space.

\paragraph{CommonSenseQA (CSQA)~\cite{talmor_commonsenseqa_2019}.}
CSQA targets everyday reasoning about world dynamics. We employ CSQA primarily as a \textbf{sanity check}; because the target answers are
typically single words or short phrases, exact-match metrics (BLEU, F1) exhibit limited resolution. Consequently, we interpret CSQA results
primarily through their consistency with broader coverage and diversity trends, rather than gains in token-level overlap.

\subsection{Baselines}\label{app:baselines}

\subsubsection{Single-objective baselines}

\paragraph{Best-of-$N$ (Oversample).}
This baseline generates $N > k$ sequences independently and retains the top-$k$ according to a ranking function.
In its strongest form, this ranking relies on an external scorer, verifier, or task-specific evaluation metric (e.g., exact match),
which provides a high-quality but expensive/oracle selection signal, the latter serving as an upper-bound reference under matched compute.
In practice, we instantiate this baseline using externally-evaluated perplexity (or a corresponding internal proxy), F1 score, or even
answer accuracy in the math setting. Note that this baseline natively implements a partition constraint, since all proposals stem from independent
ancestors.

\paragraph{Standard (Greedy) Beam Search.}
This baseline performs quality-only step-wise selection in the search regime. At each step, candidates are ranked independently according
to the score $\mathbf{Q}$, and the top elements are retained under the partition constraint. This ignores redundancy across beams and
serves as the canonical quality-centric decoding baseline.

\subsubsection{Diversity-promoting baselines}

\paragraph{Diverse Beam Search (Transversal MMR).}
We adapt diverse beam search to the discrete diffusion setting using a partition-constrained Maximal Marginal Relevance (MMR) objective.
At each selection step, candidates are scored by combining their individual quality with a penalty proportional to their similarity to
already selected elements. The selection proceeds greedily under the transversal constraint (one item per group).

To better match the exploration capacity of D5P4, we employ multiple parallel initializations of the greedy procedure and retain the subset
with the highest cumulative MMR objective. This reduces sensitivity to initialization and makes the comparison closer to our method.

\paragraph{Standard DPP Sampling (Non-Transversal Only).}
For completeness, we consider sampling from a DPP defined by the kernel $\mathbf{L}$ using spectral sampling (via the DPPy~\cite{dppy} library).
In this setting, subsets of size $k$ are sampled directly from the unconstrained DPP distribution.

This baseline is restricted to the non-transversal setting, as exact sampling under partition constraints is not supported. It provides
a useful reference for stochastic diversity-oriented selection, but differs from our approach, which performs deterministic MAP inference
under a constrained objective.

% \paragraph{Diversity-Promoting Baselines}

% \textbf{Diverse Beam Search (Transversal MMR).}
% Since discrete diffusion lacks the partial hypotheses used in autoregressive diverse beam search~\citep{vijayakumar2016diverse}, we adapt
% this baseline as a partition-constrained Maximal Marginal Relevance (MMR)~\citep{mmr98} objective.
% We employ a greedy subset selection strategy where candidate scores are penalized by their similarity to the currently selected set. To
% ensure robustness, we run the greedy procedure from multiple initializations and select the subset with the highest cumulative score. This
% strategy also matches the initialization strategy in our method.

% \textbf{Standard DPP Sampling (Non-Transversal Only).}
% To isolate the impact of our proposed partition constraints, we use the \texttt{DPPy} library \citep{dppy} to sample exactly $k$ items from
% the unconstrained kernel $\mathbf{L}$. This serves as a baseline for pure DPP sampling partition constraints: exact transversal sampling is
% not possible in this setting.

\section{Extended results}

In this section, we provide extended experimental results and comprehensive visualizations to complement the main text. Specifically, we
detail the parameter dynamics governing the quality--diversity trade-offs for all evaluated search methods in the open-ended experiment.
We also present the exhaustive question-answering metrics across all datasets (Section~\ref{app:qa-LLaDA-full}), and supply the complete
suite of evaluation metrics demonstrating D5P4's effectiveness at mitigating CFG-induced mode collapse (Section~\ref{app:cfg})
and on mathematical reasoning (Section~\ref{app:gsm8k_full}).

\subsection{Extended Quality--Diversity Trade-offs in MDLM generations}

\begin{figure}[ht]
  \centering

  \begin{subfigure}{0.49\linewidth}
    \centering
    \includegraphics[width=\linewidth]{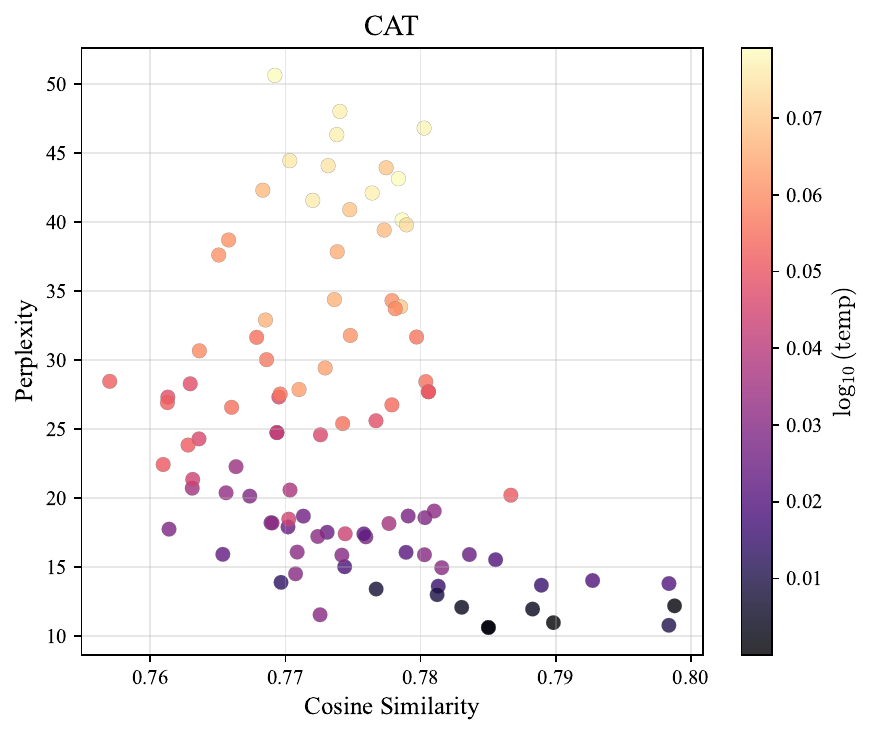}
    \caption{Categorical temperature scaling}
  \end{subfigure}
  \hfill
  \begin{subfigure}{0.49\linewidth}
    \centering
    \includegraphics[width=\linewidth]{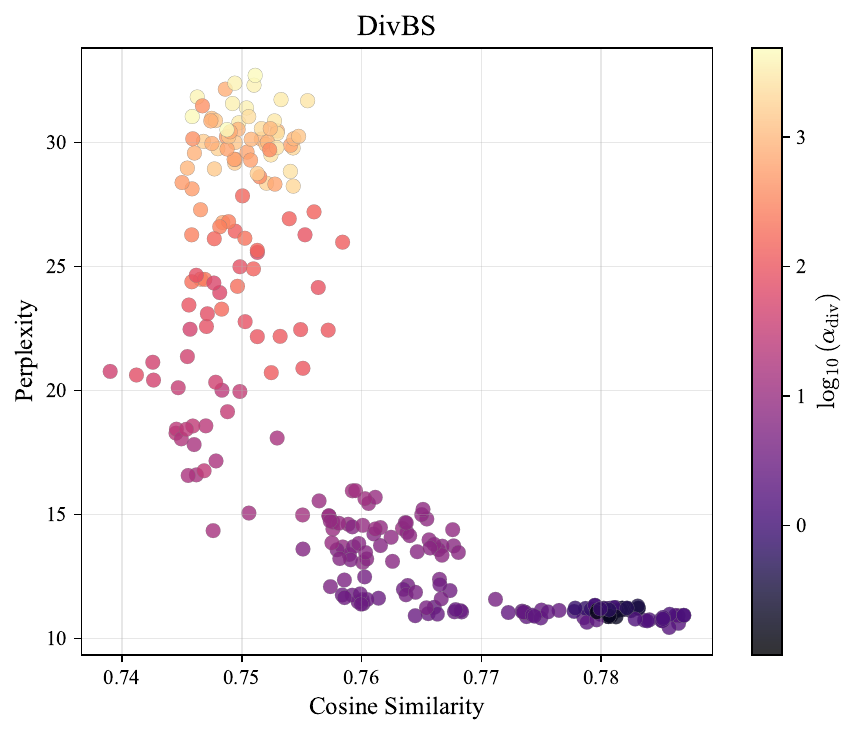}
    \caption{DivBS with varying $\alpha_{div}$}
  \end{subfigure}

  \vspace{1.5em}

  \begin{subfigure}{0.49\linewidth}
    \centering
    \includegraphics[width=\linewidth]{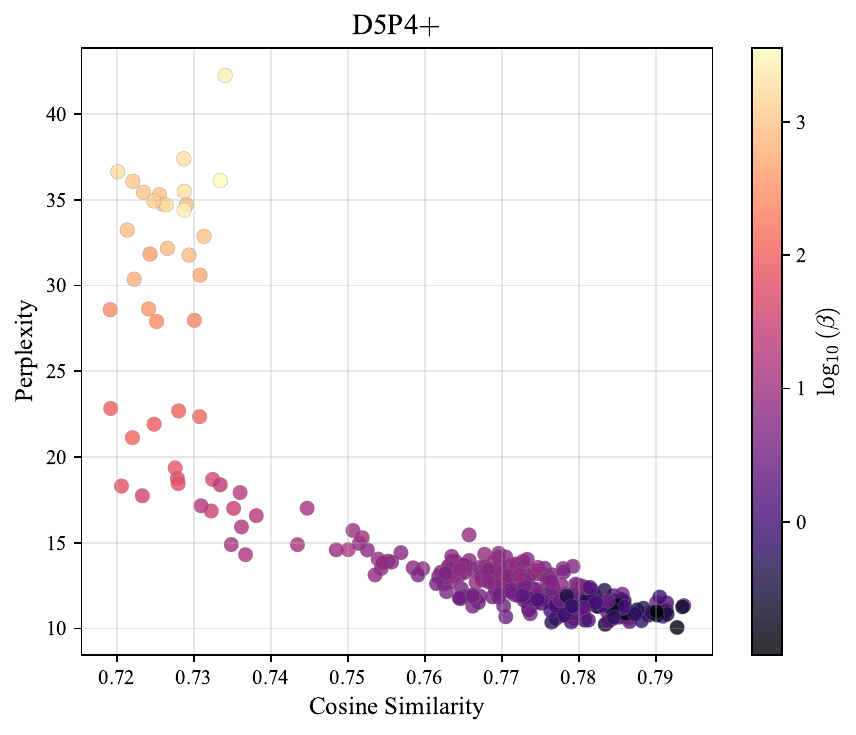}
    \caption{D5P4$+$ with varying $\beta$}
  \end{subfigure}
  \hfill
  \begin{subfigure}{0.49\linewidth}
    \centering
    \includegraphics[width=\linewidth]{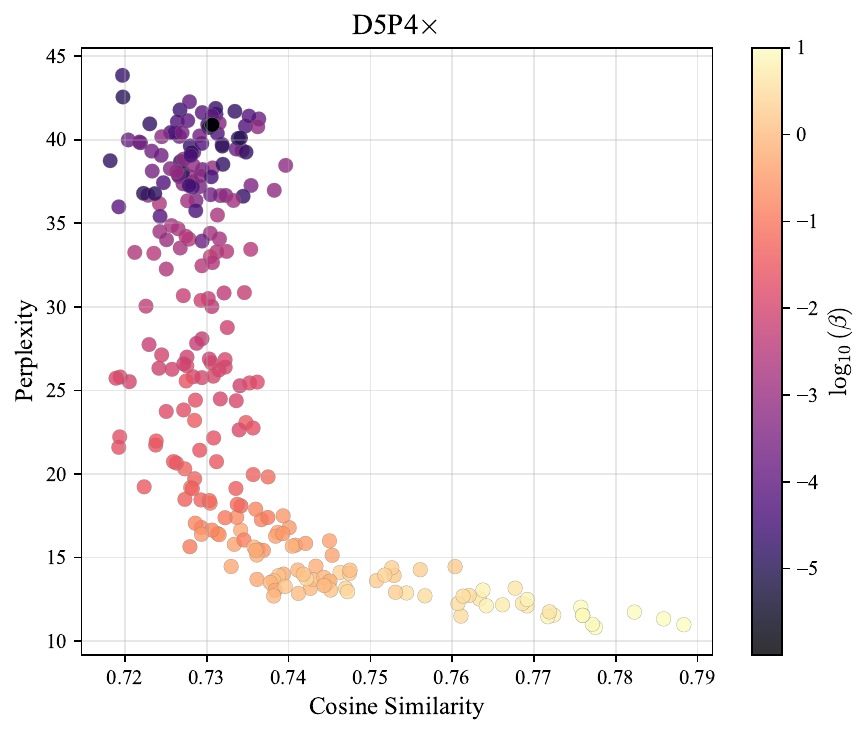}
    \caption{D5P4$\times$ with varying $\beta$}
  \end{subfigure}

  \caption{Dynamics of the parameters controlling diversity.}
  \label{fig:dynamics}
\end{figure}

We observe that the point density is non-uniform across the Pareto fronts, a byproduct of the underlying hyperparameter sampling
distribution. This density imbalance biases the calculation of the Pearson correlation, potentially misrepresenting how specific parameters
govern the target metrics. To mitigate this artifact, we employ a stratified resampling approach: we partition the logarithmic parameter
range into equal-width bins and compute the mean metric value within each bin. The final correlations are then calculated over these
bin-averaged values, ensuring that over-represented regions of the hyperparameter space do not disproportionately influence the analysis.

\subsection{Exhaustive QA metrics}\label{app:qa-LLaDA-full}

Tables~\ref{tab:qa-full-truthful_qa}, \ref{tab:qa-full-ai2_arc}, and \ref{tab:qa-full-commonsense_qa} detail the comprehensive performance
metrics across all evaluated question-answering datasets, expanding upon the summarized results provided in the main text. To ensure a
rigorous comparison, we report the complete suite of quality and diversity metrics alongside their variance estimates. These
extended results confirm that the improvements in lexical and semantic diversity achieved by D5P4 are statistically significant and do not
come at the expense of task-specific accuracy. We compare our approach against a proposal-matched independent baseline ($k$-Ind.),
the perplexity BoN (BN-PPL), its internal-entropy proxy (BoN-INT), Diverse Beam Search (DBS) and an oracle F1 BoN.

\begin{table}[ht!]
  \centering
  \renewcommand{\arraystretch}{1.2}
  \caption{Exhaustive metrics for \textbf{TruthfulQA}. Bold indicates best non-oracle method.}
  \label{tab:qa-full-truthful_qa}
  \resizebox{\linewidth}{!}{
    \begin{tabular}{l | c >{\columncolor{gray!10}}c c c >{\columncolor{blue!10}}c >{\columncolor{blue!10}}c | c}
      \hline
      \rule{0pt}{3ex}\textbf{Metric} & $k$-Ind. & BN-PPL & BoN-INT & DBS & D5P4 & D5P4$^P$ & BoN-F1 (oracle) \\
      \hline \hline
      $\downarrow$ PPL & 6.46 $\pm$ 0.14 & \textbf{5.24 $\pm$ 0.10} & 6.64 $\pm$ 0.15 & 6.37 $\pm$ 0.14 & 6.40 $\pm$ 0.14 & 6.72 $\pm$ 0.17
      & 7.01 $\pm$ 0.16 \\
      $\uparrow$ Entropy & 3.73 $\pm$ 0.01 & \textbf{3.75 $\pm$ 0.01} & 3.69 $\pm$ 0.01 & 3.71 $\pm$ 0.01 & 3.71 $\pm$ 0.01 & 3.60 $\pm$
      0.01 & 3.67 $\pm$ 0.01 \\
      $\uparrow$ F1 (\%) & 18.32 $\pm$ 0.16 & 17.31 $\pm$ 0.15 & 19.39 $\pm$ 0.18 & 19.07 $\pm$ 0.17 & 19.02 $\pm$ 0.17 & \textbf{20.71
      $\pm$ 0.20} & 21.20 $\pm$ 0.18 \\
      $\uparrow$ F1@k (\%) & 20.74 $\pm$ 0.30 & 19.16 $\pm$ 0.28 & 22.00 $\pm$ 0.33 & 22.55 $\pm$ 0.34 & 22.71 $\pm$ 0.34 & \textbf{24.84
      $\pm$ 0.39} & 23.25 $\pm$ 0.33 \\
      $\uparrow$ BLEU (\%) & 4.40 & 4.32 & 4.51 & 4.52 & 4.50 & \textbf{4.55} & 4.88 \\
      $\uparrow$ BLEU@k (\%) & 7.97 $\pm$ 0.24 & 7.18 $\pm$ 0.23 & 8.87 $\pm$ 0.27 & 9.19 $\pm$ 0.28 & 9.26 $\pm$ 0.27 & \textbf{11.55
      $\pm$ 0.36} & 9.35 $\pm$ 0.27 \\
      $\uparrow$ R-COS (\%) & 87.36 $\pm$ 0.12 & 87.29 $\pm$ 0.12 & 87.41 $\pm$ 0.12 & 87.53 $\pm$ 0.12 & 87.56 $\pm$ 0.12 & \textbf{87.82
      $\pm$ 0.12} & 87.60 $\pm$ 0.12 \\
      %$\downarrow$ WD (\%) & 57.85 $\pm$ 0.22 & 57.94 $\pm$ 0.22 & 57.74 $\pm$ 0.22 & 57.69 $\pm$ 0.22 & 57.69 $\pm$ 0.22 & \textbf{57.40
      % $\pm$ 0.23} & 57.47 $\pm$ 0.22 \\
      $\uparrow$ Distinct-1 (\%) & 34.75 $\pm$ 0.12 & 33.15 $\pm$ 0.12 & 34.64 $\pm$ 0.13 & 36.08 $\pm$ 0.14 & \textbf{36.33 $\pm$ 0.14} &
      36.13 $\pm$ 0.15 & 34.74 $\pm$ 0.13 \\
      $\uparrow$ Distinct-2 (\%) & 60.66 $\pm$ 0.21 & 58.51 $\pm$ 0.22 & 59.69 $\pm$ 0.23 & 62.43 $\pm$ 0.22 & \textbf{63.06 $\pm$ 0.21} &
      61.57 $\pm$ 0.23 & 59.52 $\pm$ 0.22 \\
      $\uparrow$ Distinct-3 (\%) & 70.82 $\pm$ 0.25 & 68.95 $\pm$ 0.25 & 69.65 $\pm$ 0.27 & 72.49 $\pm$ 0.25 & \textbf{73.24 $\pm$ 0.24} &
      71.40 $\pm$ 0.27 & 69.32 $\pm$ 0.27 \\
      $\uparrow$ EAD (\%) & 36.19 $\pm$ 0.12 & 34.57 $\pm$ 0.12 & 36.03 $\pm$ 0.13 & 37.52 $\pm$ 0.14 & \textbf{37.78 $\pm$ 0.14} & 37.53
      $\pm$ 0.15 & 36.10 $\pm$ 0.13 \\
      $\downarrow$ S-BLEU (\%) & 44.07 $\pm$ 0.38 & 46.27 $\pm$ 0.37 & 45.57 $\pm$ 0.41 & 40.74 $\pm$ 0.38 & \textbf{39.00 $\pm$ 0.37} &
      41.64 $\pm$ 0.43 & 46.30 $\pm$ 0.41 \\
      $\downarrow$ S-COS (\%) & 96.08 $\pm$ 0.06 & 96.40 $\pm$ 0.06 & 96.12 $\pm$ 0.06 & 95.36 $\pm$ 0.07 & 95.16 $\pm$ 0.07 &
      \textbf{94.88 $\pm$ 0.08} & 96.19 $\pm$ 0.06 \\
      \hline
    \end{tabular}
  }
\end{table}

\begin{table}[ht!]
  \centering
  \renewcommand{\arraystretch}{1.2}
  \caption{Exhaustive metrics for \textbf{ARC-C}. Bold indicates best non-oracle method.}
  \label{tab:qa-full-ai2_arc}
  \resizebox{\linewidth}{!}{
    \begin{tabular}{l | c >{\columncolor{gray!10}}c c c >{\columncolor{blue!10}}c >{\columncolor{blue!10}}c | c}
      \hline
      \rule{0pt}{3ex}\textbf{Metric} & $k$-Ind. & BN-PPL & BoN-INT & DBS & D5P4 & D5P4$^P$ & BoN-F1 (oracle) \\
      \hline \hline
      $\downarrow$ PPL & 5.84 $\pm$ 0.07 & \textbf{4.93 $\pm$ 0.05} & 5.86 $\pm$ 0.08 & 5.78 $\pm$ 0.07 & 5.78 $\pm$ 0.07 & 5.82 $\pm$ 0.09
      & 5.94 $\pm$ 0.07 \\
      $\uparrow$ Entropy & 3.76 $\pm$ 0.01 & \textbf{3.77 $\pm$ 0.01} & 3.73 $\pm$ 0.01 & 3.75 $\pm$ 0.01 & 3.75 $\pm$ 0.01 & 3.67 $\pm$
      0.01 & 3.75 $\pm$ 0.01 \\
      $\uparrow$ F1 (\%) & 5.16 $\pm$ 0.06 & 5.01 $\pm$ 0.06 & 5.35 $\pm$ 0.06 & 5.22 $\pm$ 0.06 & 5.24 $\pm$ 0.06 & \textbf{5.46 $\pm$
      0.07} & 6.34 $\pm$ 0.07 \\
      $\uparrow$ F1@k (\%) & 6.14 $\pm$ 0.12 & 5.85 $\pm$ 0.10 & 6.38 $\pm$ 0.13 & 6.36 $\pm$ 0.12 & 6.42 $\pm$ 0.12 & \textbf{6.74 $\pm$
      0.14} & 7.13 $\pm$ 0.13 \\
      $\uparrow$ BLEU (\textperthousand) & 4.08 & 4.20 & 4.18 & 4.10 & 4.21 & \textbf{4.26} & 5.14 \\
      $\uparrow$ BLEU@k (\textperthousand) & 11.59 $\pm$ 0.62 & 10.64 $\pm$ 0.55 & 12.60 $\pm$ 0.70 & 12.70 $\pm$ 0.69 & 12.97 $\pm$ 0.70 &
      \textbf{15.00 $\pm$ 0.81} & 13.76 $\pm$ 0.71 \\
      $\uparrow$ R-COS (\%) & 76.12 $\pm$ 0.11 & 76.11 $\pm$ 0.11 & 76.08 $\pm$ 0.11 & 76.23 $\pm$ 0.11 & \textbf{76.28 $\pm$ 0.11} & 76.25
      $\pm$ 0.11 & 76.35 $\pm$ 0.11 \\
      %$\downarrow$ WD (\%) & 69.85 $\pm$ 0.16 & \textbf{69.80 $\pm$ 0.16} & 69.83 $\pm$ 0.16 & 69.86 $\pm$ 0.16 & 69.86 $\pm$ 0.16 & 69.90
      % $\pm$ 0.16 & 69.45 $\pm$ 0.16 \\
      $\uparrow$ Distinct-1 (\%) & 29.26 $\pm$ 0.11 & 27.88 $\pm$ 0.11 & 29.11 $\pm$ 0.11 & 30.20 $\pm$ 0.12 & \textbf{30.40 $\pm$ 0.12} &
      29.94 $\pm$ 0.12 & 29.13 $\pm$ 0.11 \\
      $\uparrow$ Distinct-2 (\%) & 54.11 $\pm$ 0.17 & 51.84 $\pm$ 0.18 & 53.38 $\pm$ 0.18 & 55.70 $\pm$ 0.17 & \textbf{56.20 $\pm$ 0.17} &
      54.76 $\pm$ 0.18 & 53.48 $\pm$ 0.18 \\
      $\uparrow$ Distinct-3 (\%) & 64.86 $\pm$ 0.20 & 62.55 $\pm$ 0.20 & 63.96 $\pm$ 0.21 & 66.68 $\pm$ 0.19 & \textbf{67.31 $\pm$ 0.19} &
      65.61 $\pm$ 0.21 & 64.03 $\pm$ 0.20 \\
      $\uparrow$ EAD (\%) & 31.61 $\pm$ 0.11 & 30.17 $\pm$ 0.11 & 31.39 $\pm$ 0.11 & 32.56 $\pm$ 0.12 & \textbf{32.78 $\pm$ 0.12} & 32.24
      $\pm$ 0.12 & 31.43 $\pm$ 0.11 \\
      $\downarrow$ S-BLEU (\%) & 52.16 $\pm$ 0.28 & 55.04 $\pm$ 0.28 & 53.54 $\pm$ 0.30 & 49.03 $\pm$ 0.27 & \textbf{47.71 $\pm$ 0.27} &
      49.70 $\pm$ 0.30 & 53.55 $\pm$ 0.29 \\
      $\downarrow$ S-COS (\%) & 97.44 $\pm$ 0.03 & 97.65 $\pm$ 0.03 & 97.50 $\pm$ 0.03 & 97.00 $\pm$ 0.04 & 96.87 $\pm$ 0.04 &
      \textbf{96.76 $\pm$ 0.04} & 97.54 $\pm$ 0.03 \\
      \hline
    \end{tabular}
  }
\end{table}

\begin{table}[ht!]
  \centering
  \renewcommand{\arraystretch}{1.2}
  \caption{Exhaustive metrics for \textbf{CommonSenseQA}. Bold indicates best non-oracle method.}
  \label{tab:qa-full-commonsense_qa}
  \resizebox{\linewidth}{!}{
    \begin{tabular}{l | c >{\columncolor{gray!10}}c c c >{\columncolor{blue!10}}c >{\columncolor{blue!10}}c | c}
      \hline
      \rule{0pt}{3ex}\textbf{Metric} & $k$-Ind. & BN-PPL & BoN-INT & DBS & D5P4 & D5P4$^P$ & BoN-F1 (oracle) \\
      \hline \hline
      $\downarrow$ PPL & 11.35 $\pm$ 0.22 & \textbf{8.86 $\pm$ 0.16} & 11.81 $\pm$ 0.23 & 11.24 $\pm$ 0.22 & 11.23 $\pm$ 0.22 & 12.74 $\pm$
      0.30 & 11.35 $\pm$ 0.22 \\
      $\uparrow$ Entropy & 3.51 $\pm$ 0.01 & \textbf{3.56 $\pm$ 0.01} & 3.47 $\pm$ 0.01 & 3.49 $\pm$ 0.01 & 3.49 $\pm$ 0.01 & 3.33 $\pm$
      0.01 & 3.51 $\pm$ 0.01 \\
      $\uparrow$ F1 (\textperthousand) & 12.30 $\pm$ 0.40 & 11.80 $\pm$ 0.40 & 12.70 $\pm$ 0.40 & 12.20 $\pm$ 0.40 & 12.20 $\pm$ 0.40 &
      \textbf{13.70 $\pm$ 0.40} & 18.50 $\pm$ 0.40 \\
      $\uparrow$ F1@k (\%) & 1.80 $\pm$ 0.07 & 1.63 $\pm$ 0.07 & 1.80 $\pm$ 0.07 & 1.87 $\pm$ 0.07 & 1.90 $\pm$ 0.07 & \textbf{2.14 $\pm$
      0.09} & 2.37 $\pm$ 0.08 \\
      $\uparrow$ BLEU (\textperthousand) & 0.11 & 0.12 & 0.12 & 0.11 & 0.11 & \textbf{0.13} & 0.15 \\
      $\uparrow$ BLEU@k (\textperthousand) & 4.50 $\pm$ 0.15 & 3.98 $\pm$ 0.13 & 4.69 $\pm$ 0.15 & 4.90 $\pm$ 0.16 & 4.99 $\pm$ 0.16 &
      \textbf{7.45 $\pm$ 0.39} & 5.01 $\pm$ 0.15 \\
      $\uparrow$ R-COS (\%) & 74.19 $\pm$ 0.08 & 74.12 $\pm$ 0.08 & 74.18 $\pm$ 0.08 & 74.39 $\pm$ 0.08 & 74.41 $\pm$ 0.08 & \textbf{74.53
      $\pm$ 0.09} & 74.41 $\pm$ 0.08 \\
      %$\downarrow$ WD (\%) & 72.91 $\pm$ 0.12 & 72.90 $\pm$ 0.12 & 72.83 $\pm$ 0.12 & 72.85 $\pm$ 0.11 & 72.86 $\pm$ 0.11 & \textbf{72.75
      % $\pm$ 0.12} & 72.60 $\pm$ 0.12 \\
      $\uparrow$ Distinct-1 (\%) & 34.09 $\pm$ 0.10 & 32.34 $\pm$ 0.10 & 33.88 $\pm$ 0.11 & 35.26 $\pm$ 0.11 & \textbf{35.52 $\pm$ 0.11} &
      35.33 $\pm$ 0.12 & 33.95 $\pm$ 0.11 \\
      $\uparrow$ Distinct-2 (\%) & 56.97 $\pm$ 0.22 & 54.96 $\pm$ 0.21 & 55.67 $\pm$ 0.23 & 58.62 $\pm$ 0.22 & \textbf{59.36 $\pm$ 0.23} &
      56.56 $\pm$ 0.24 & 56.59 $\pm$ 0.22 \\
      $\uparrow$ Distinct-3 (\%) & 65.87 $\pm$ 0.27 & 64.21 $\pm$ 0.26 & 64.18 $\pm$ 0.28 & 67.48 $\pm$ 0.27 & \textbf{68.37 $\pm$ 0.27} &
      64.71 $\pm$ 0.30 & 65.46 $\pm$ 0.27 \\
      $\uparrow$ EAD (\%) & 37.70 $\pm$ 0.11 & 36.00 $\pm$ 0.11 & 37.29 $\pm$ 0.12 & 38.88 $\pm$ 0.12 & \textbf{39.18 $\pm$ 0.12} & 38.68
      $\pm$ 0.13 & 37.51 $\pm$ 0.12 \\
      $\downarrow$ S-BLEU (\%) & 51.70 $\pm$ 0.41 & 53.84 $\pm$ 0.39 & 54.42 $\pm$ 0.43 & 48.27 $\pm$ 0.42 & \textbf{46.86 $\pm$ 0.43} &
      51.91 $\pm$ 0.47 & 52.51 $\pm$ 0.41 \\
      $\downarrow$ S-COS (\%) & 96.52 $\pm$ 0.04 & 96.94 $\pm$ 0.04 & 96.76 $\pm$ 0.04 & 95.91 $\pm$ 0.05 & 95.77 $\pm$ 0.05 &
      \textbf{95.59 $\pm$ 0.06} & 96.62 $\pm$ 0.04 \\
      \hline
    \end{tabular}
  }
\end{table}

\subsection{Exhaustive GSM8K metrics}\label{app:gsm8k_full}

Tables~\ref{tab:gsm8k-full-low_confidence}, \ref{tab:gsm8k-full-selection_temperature}, and \ref{tab:gsm8k-full-zero} provide the complete
set of performance metrics for GSM8K under varying remasking strategies. Confidence utilizes LLaDA’s default self-confidence remasking.
Stochastic remasking uses tempered confidence scores ($T_{sel}=0.1$), following~\cite{chen2025optimizing}, while keeping the categorical
distribution unit-untempered. Tempered confidence argmax restores the full configuration used in~\cite{chen2025optimizing}, using
the logits argmax instead of sampling from the categorical distribution and still varying unmasking order. We compare our methods to the
PPL BoN, its internal proxy, Greedy Beam Search (GBS) and Diverse Beam Search (DBS). The oracle (BN-A) uses the mathematical accuracy
of the answer as the selection signal. This makes the reported pass@4 effectively measure the pass@16 of the full independent samples.

\begin{table}[ht!]
  \centering
  \renewcommand{\arraystretch}{1.2}
  \caption{Exhaustive metrics for GSM8K (Confidence). Bold indicates best practical method.}
  \label{tab:gsm8k-full-low_confidence}
  \resizebox{\linewidth}{!}{
    \begin{tabular}{l | c >{\columncolor{gray!10}}c c c c >{\columncolor{blue!10}}c | c}
      \hline
      \rule{0pt}{3ex}\textbf{Metric} & Indep. & BN-P & BN-I & GBS & DBS & D5P4 & BN-A (oracle) \\
      \hline \hline
      $\uparrow$ Entropy & \textbf{3.79 $\pm$ 0.00} & 3.78 $\pm$ 0.00 & 3.79 $\pm$ 0.00 & 3.79 $\pm$ 0.00 & 3.79 $\pm$ 0.00 & 3.79 $\pm$
      0.00 & 3.79 $\pm$ 0.00 \\
      $\uparrow$ F1 (\%) & \textbf{29.68 $\pm$ 0.04} & 29.56 $\pm$ 0.04 & 29.65 $\pm$ 0.04 & 29.68 $\pm$ 0.04 & 29.67 $\pm$ 0.04 & 29.67
      $\pm$ 0.04 & 29.72 $\pm$ 0.04 \\
      $\uparrow$ F1@k (\%) & 31.05 $\pm$ 0.07 & 30.83 $\pm$ 0.07 & 30.98 $\pm$ 0.08 & 31.19 $\pm$ 0.07 & 31.39 $\pm$ 0.07 & \textbf{31.48
      $\pm$ 0.07} & 31.09 $\pm$ 0.07 \\
      $\uparrow$ BLEU (\%) & 7.95 & 7.97 & 7.93 & \textbf{7.98} & 7.98 & 7.98 & 8.01 \\
      $\uparrow$ BLEU@k (\%) & 8.52 $\pm$ 0.04 & 8.48 $\pm$ 0.04 & 8.49 $\pm$ 0.04 & 8.70 $\pm$ 0.04 & 8.85 $\pm$ 0.04 & \textbf{8.93 $\pm$
      0.04} & 8.58 $\pm$ 0.04 \\
      $\uparrow$ Distinct-1 (\%) & 9.75 $\pm$ 0.02 & 9.33 $\pm$ 0.02 & 9.64 $\pm$ 0.02 & 9.89 $\pm$ 0.02 & 10.15 $\pm$ 0.02 & \textbf{10.28
      $\pm$ 0.02} & 9.74 $\pm$ 0.02 \\
      $\uparrow$ Distinct-2 (\%) & 22.78 $\pm$ 0.04 & 21.46 $\pm$ 0.04 & 22.46 $\pm$ 0.04 & 23.53 $\pm$ 0.04 & 24.67 $\pm$ 0.04 &
      \textbf{25.27 $\pm$ 0.04} & 22.72 $\pm$ 0.04 \\
      $\uparrow$ Distinct-3 (\%) & 31.27 $\pm$ 0.06 & 29.38 $\pm$ 0.06 & 30.82 $\pm$ 0.06 & 32.65 $\pm$ 0.06 & 34.56 $\pm$ 0.06 &
      \textbf{35.62 $\pm$ 0.06} & 31.18 $\pm$ 0.06 \\
      $\uparrow$ EAD (\%) & 10.74 $\pm$ 0.02 & 10.28 $\pm$ 0.02 & 10.62 $\pm$ 0.02 & 10.89 $\pm$ 0.02 & 11.18 $\pm$ 0.02 & \textbf{11.33
      $\pm$ 0.02} & 10.73 $\pm$ 0.02 \\
      $\downarrow$ S-BLEU (\%) & 85.19 $\pm$ 0.07 & 87.28 $\pm$ 0.07 & 85.81 $\pm$ 0.08 & 83.44 $\pm$ 0.07 & 80.77 $\pm$ 0.07 &
      \textbf{79.13 $\pm$ 0.08} & 85.30 $\pm$ 0.07 \\
      $\uparrow$ Pass@1 (\%) & \textbf{44.5 $\pm$ 0.3} & 46.5 $\pm$ 0.3 & 44.4 $\pm$ 0.3 & 44.4 $\pm$ 0.3 & 44.4 $\pm$ 0.3 & 44.4 $\pm$ 0.3
      & 55.9 $\pm$ 0.3 \\
      $\uparrow$ Pass@2 (\%) & 49.9 $\pm$ 0.3 & 51.6 $\pm$ 0.3 & 49.6 $\pm$ 0.4 & 50.8 $\pm$ 0.3 & 51.9 $\pm$ 0.3 & \textbf{52.5 $\pm$ 0.3}
      & 59.8 $\pm$ 0.3 \\
      $\uparrow$ Pass@4 (\%) & 54.5 $\pm$ 0.3 & 56.3 $\pm$ 0.3 & 54.0 $\pm$ 0.4 & 56.3 $\pm$ 0.3 & 58.2 $\pm$ 0.3 & \textbf{59.3 $\pm$ 0.3}
      & 64.3 $\pm$ 0.3 \\
      \hline
    \end{tabular}
  }
\end{table}

\begin{table}[ht!]
  \centering
  \renewcommand{\arraystretch}{1.2}
  \caption{Exhaustive metrics for GSM8K (Stochastic). Bold indicates best practical method.}
  \label{tab:gsm8k-full-selection_temperature}
  \resizebox{\linewidth}{!}{
    \begin{tabular}{l | c >{\columncolor{gray!10}}c c c c >{\columncolor{blue!10}}c | c}
      \hline
      \rule{0pt}{3ex}\textbf{Metric} & Indep. & BN-P & BN-I & GBS & DBS & D5P4 & BN-A (oracle) \\
      \hline \hline
      $\uparrow$ Entropy & \textbf{3.81 $\pm$ 0.00} & 3.79 $\pm$ 0.00 & 3.80 $\pm$ 0.00 & 3.79 $\pm$ 0.00 & 3.79 $\pm$ 0.00 & 3.79 $\pm$
      0.00 & 3.81 $\pm$ 0.00 \\
      $\uparrow$ F1 (\%) & 29.49 $\pm$ 0.03 & 29.10 $\pm$ 0.03 & 29.66 $\pm$ 0.04 & \textbf{29.73 $\pm$ 0.03} & 29.62 $\pm$ 0.03 & 29.62
      $\pm$ 0.03 & 29.64 $\pm$ 0.03 \\
      $\uparrow$ F1@k (\%) & 32.28 $\pm$ 0.07 & 31.39 $\pm$ 0.07 & 32.40 $\pm$ 0.08 & 32.14 $\pm$ 0.07 & 32.76 $\pm$ 0.07 & \textbf{32.85
      $\pm$ 0.07} & 32.41 $\pm$ 0.07 \\
      $\uparrow$ BLEU (\%) & 7.98 & 7.90 & 7.94 & \textbf{8.08} & 8.04 & 8.06 & 8.13 \\
      $\uparrow$ BLEU@k (\%) & 9.67 $\pm$ 0.04 & 9.22 $\pm$ 0.04 & 9.60 $\pm$ 0.04 & 9.50 $\pm$ 0.04 & 9.83 $\pm$ 0.04 & \textbf{9.98 $\pm$
      0.04} & 9.82 $\pm$ 0.04 \\
      $\uparrow$ Distinct-1 (\%) & 11.23 $\pm$ 0.02 & 10.18 $\pm$ 0.01 & 11.20 $\pm$ 0.02 & 10.69 $\pm$ 0.02 & 11.17 $\pm$ 0.01 &
      \textbf{11.32 $\pm$ 0.01} & 11.08 $\pm$ 0.01 \\
      $\uparrow$ Distinct-2 (\%) & 30.36 $\pm$ 0.04 & 26.92 $\pm$ 0.03 & 29.78 $\pm$ 0.04 & 27.63 $\pm$ 0.04 & 29.90 $\pm$ 0.04 &
      \textbf{30.67 $\pm$ 0.04} & 29.79 $\pm$ 0.04 \\
      $\uparrow$ Distinct-3 (\%) & 44.32 $\pm$ 0.05 & 39.29 $\pm$ 0.05 & 43.30 $\pm$ 0.06 & 39.80 $\pm$ 0.05 & 43.47 $\pm$ 0.05 &
      \textbf{44.77 $\pm$ 0.05} & 43.41 $\pm$ 0.05 \\
      $\uparrow$ EAD (\%) & 12.39 $\pm$ 0.02 & 11.26 $\pm$ 0.01 & 12.34 $\pm$ 0.02 & 11.78 $\pm$ 0.02 & 12.31 $\pm$ 0.02 & \textbf{12.48
      $\pm$ 0.02} & 12.23 $\pm$ 0.02 \\
      $\downarrow$ S-BLEU (\%) & 66.70 $\pm$ 0.07 & 72.69 $\pm$ 0.07 & 68.29 $\pm$ 0.08 & 73.41 $\pm$ 0.07 & 67.48 $\pm$ 0.07 &
      \textbf{65.16 $\pm$ 0.07} & 67.93 $\pm$ 0.07 \\
      $\uparrow$ Pass@1 (\%) & 45.5 $\pm$ 0.2 & 51.8 $\pm$ 0.3 & 46.8 $\pm$ 0.3 & 47.3 $\pm$ 0.3 & \textbf{47.4 $\pm$ 0.2} & 47.3 $\pm$ 0.2
      & 73.8 $\pm$ 0.2 \\
      $\uparrow$ Pass@2 (\%) & 58.8 $\pm$ 0.3 & 63.6 $\pm$ 0.3 & 59.7 $\pm$ 0.3 & 58.9 $\pm$ 0.3 & 60.7 $\pm$ 0.3 & \textbf{61.2 $\pm$ 0.3}
      & 78.9 $\pm$ 0.2 \\
      $\uparrow$ Pass@4 (\%) & 69.4 $\pm$ 0.3 & 72.6 $\pm$ 0.3 & 70.1 $\pm$ 0.3 & 68.3 $\pm$ 0.3 & 71.2 $\pm$ 0.3 & \textbf{72.2 $\pm$ 0.3}
      & 83.6 $\pm$ 0.2 \\
      \hline
    \end{tabular}
  }
\end{table}

\begin{table}[ht!]
  \centering
  \renewcommand{\arraystretch}{1.2}
  \caption{Exhaustive metrics for GSM8K (Temp. conf. argmax). Bold indicates best practical method.}
  \label{tab:gsm8k-full-zero}
  \resizebox{\linewidth}{!}{
    \begin{tabular}{l | c >{\columncolor{gray!10}}c c c c >{\columncolor{blue!10}}c | c}
      \hline
      \rule{0pt}{3ex}\textbf{Metric} & Indep. & BN-P & BN-I & GBS & DBS & D5P4 & BN-A (oracle) \\
      \hline \hline
      $\uparrow$ Entropy & \textbf{3.80 $\pm$ 0.00} & 3.79 $\pm$ 0.00 & 3.80 $\pm$ 0.00 & 3.79 $\pm$ 0.00 & 3.79 $\pm$ 0.00 & 3.79 $\pm$
      0.00 & 3.80 $\pm$ 0.00 \\
      $\uparrow$ F1 (\%) & 29.50 $\pm$ 0.04 & 29.10 $\pm$ 0.04 & 29.62 $\pm$ 0.04 & \textbf{29.72 $\pm$ 0.04} & 29.61 $\pm$ 0.04 & 29.57
      $\pm$ 0.04 & 29.61 $\pm$ 0.04 \\
      $\uparrow$ F1@k (\%) & 32.31 $\pm$ 0.08 & 31.40 $\pm$ 0.08 & 32.32 $\pm$ 0.08 & 32.10 $\pm$ 0.08 & 32.69 $\pm$ 0.08 & \textbf{32.79
      $\pm$ 0.08} & 32.39 $\pm$ 0.08 \\
      $\uparrow$ BLEU (\%) & 7.99 & 7.89 & 7.93 & \textbf{8.08} & 8.05 & 8.03 & 8.12 \\
      $\uparrow$ BLEU@k (\%) & 9.70 $\pm$ 0.04 & 9.22 $\pm$ 0.04 & 9.58 $\pm$ 0.04 & 9.49 $\pm$ 0.04 & 9.84 $\pm$ 0.05 & \textbf{9.91 $\pm$
      0.05} & 9.80 $\pm$ 0.04 \\
      $\uparrow$ Distinct-1 (\%) & 11.22 $\pm$ 0.02 & 10.16 $\pm$ 0.01 & 11.18 $\pm$ 0.02 & 10.65 $\pm$ 0.02 & 11.14 $\pm$ 0.02 &
      \textbf{11.30 $\pm$ 0.02} & 11.08 $\pm$ 0.02 \\
      $\uparrow$ Distinct-2 (\%) & 30.32 $\pm$ 0.04 & 26.89 $\pm$ 0.04 & 29.71 $\pm$ 0.05 & 27.42 $\pm$ 0.04 & 29.80 $\pm$ 0.04 &
      \textbf{30.57 $\pm$ 0.04} & 29.75 $\pm$ 0.04 \\
      $\uparrow$ Distinct-3 (\%) & 44.23 $\pm$ 0.06 & 39.24 $\pm$ 0.06 & 43.15 $\pm$ 0.06 & 39.40 $\pm$ 0.06 & 43.30 $\pm$ 0.06 &
      \textbf{44.64 $\pm$ 0.06} & 43.35 $\pm$ 0.06 \\
      $\uparrow$ EAD (\%) & 12.39 $\pm$ 0.02 & 11.25 $\pm$ 0.01 & 12.31 $\pm$ 0.02 & 11.74 $\pm$ 0.02 & 12.29 $\pm$ 0.02 & \textbf{12.46
      $\pm$ 0.02} & 12.22 $\pm$ 0.02 \\
      $\downarrow$ S-BLEU (\%) & 66.70 $\pm$ 0.08 & 72.77 $\pm$ 0.08 & 68.56 $\pm$ 0.08 & 74.06 $\pm$ 0.08 & 67.82 $\pm$ 0.08 &
      \textbf{65.31 $\pm$ 0.08} & 67.97 $\pm$ 0.08 \\
      $\uparrow$ Pass@1 (\%) & 45.2 $\pm$ 0.3 & 51.1 $\pm$ 0.3 & 46.2 $\pm$ 0.3 & 47.2 $\pm$ 0.3 & \textbf{47.2 $\pm$ 0.3} & 47.1 $\pm$ 0.3
      & 73.4 $\pm$ 0.3 \\
      $\uparrow$ Pass@2 (\%) & 58.5 $\pm$ 0.3 & 62.9 $\pm$ 0.3 & 58.9 $\pm$ 0.3 & 58.6 $\pm$ 0.3 & 60.3 $\pm$ 0.3 & \textbf{60.9 $\pm$ 0.3}
      & 78.6 $\pm$ 0.3 \\
      $\uparrow$ Pass@4 (\%) & 69.2 $\pm$ 0.3 & 72.0 $\pm$ 0.3 & 69.4 $\pm$ 0.3 & 67.5 $\pm$ 0.3 & 70.7 $\pm$ 0.3 & \textbf{71.8 $\pm$ 0.3}
      & 83.5 $\pm$ 0.3 \\
      \hline
    \end{tabular}
  }
\end{table}

\subsection{Comprehensive Metrics for CFG Collapse Mitigation using D5P4}\label{app:cfg}

This section provides the exhaustive metrics for our high classifier-free guidance (CFG) study. We evaluate \textbf{D5P4} against
\textbf{independent sampling}, \textbf{Greedy Beam Search (GBS)}, and the \textbf{perplexity BoN} and \textbf{internal BoN} baselines.
To provide a clear view of the diversity-quality trade-offs as the CFG scale increases, the results are structured to separate the unguided
reference from the guidance trajectory: Individual plots establish the base performance for each method without guidance. Continuous graphs
track the impact of increasing the CFG scale on quality, accuracy, and diversity metrics. These figures detail the evolution of lexical and
semantic diversity metrics, illustrating the effectiveness of the DPP objective in mitigating mode collapse under high guidance regimes.

\begin{figure}[H]
  \centering
  \begin{subfigure}{0.49\linewidth}
    \centering
    \includegraphics[width=\linewidth]{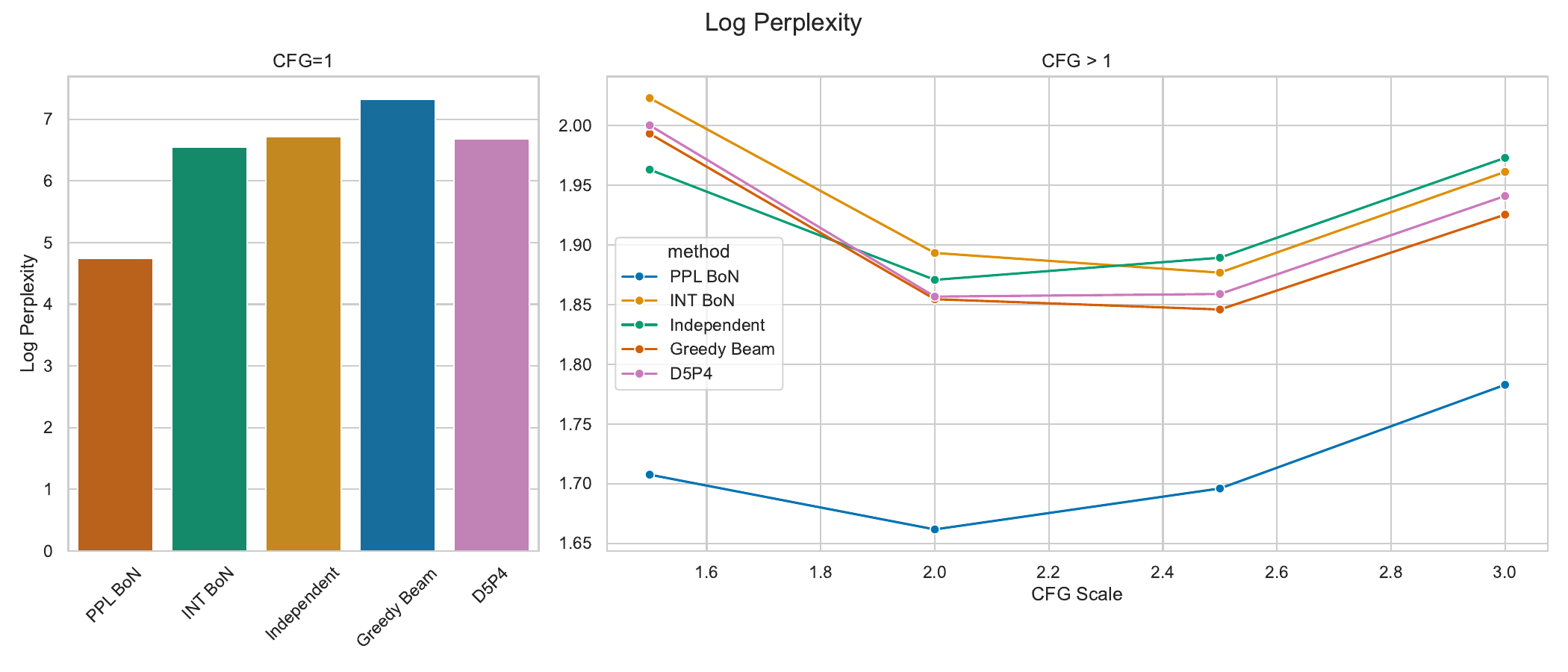}
  \end{subfigure}
  \hfill
  \begin{subfigure}{0.49\linewidth}
    \centering
    \includegraphics[width=\linewidth]{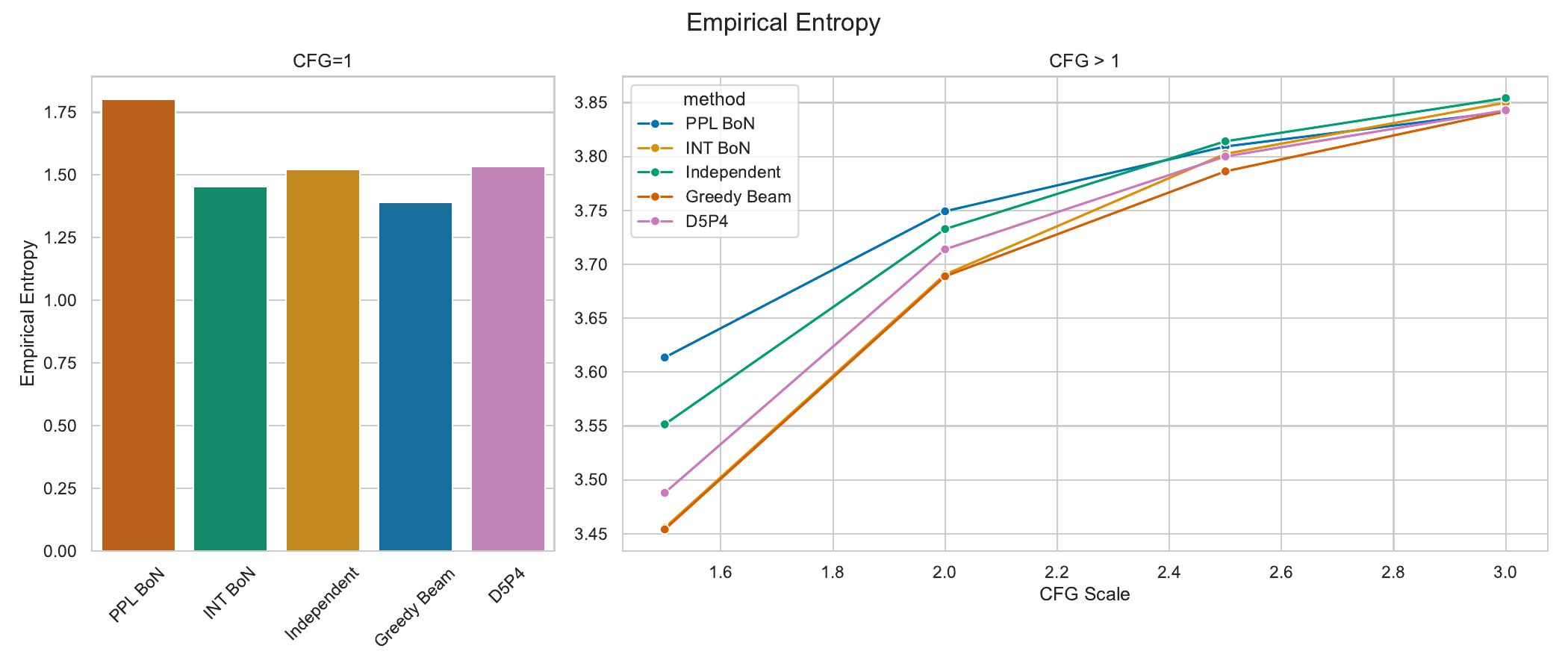}
  \end{subfigure}

  \begin{subfigure}{0.49\linewidth}
    \centering
    \includegraphics[width=\linewidth]{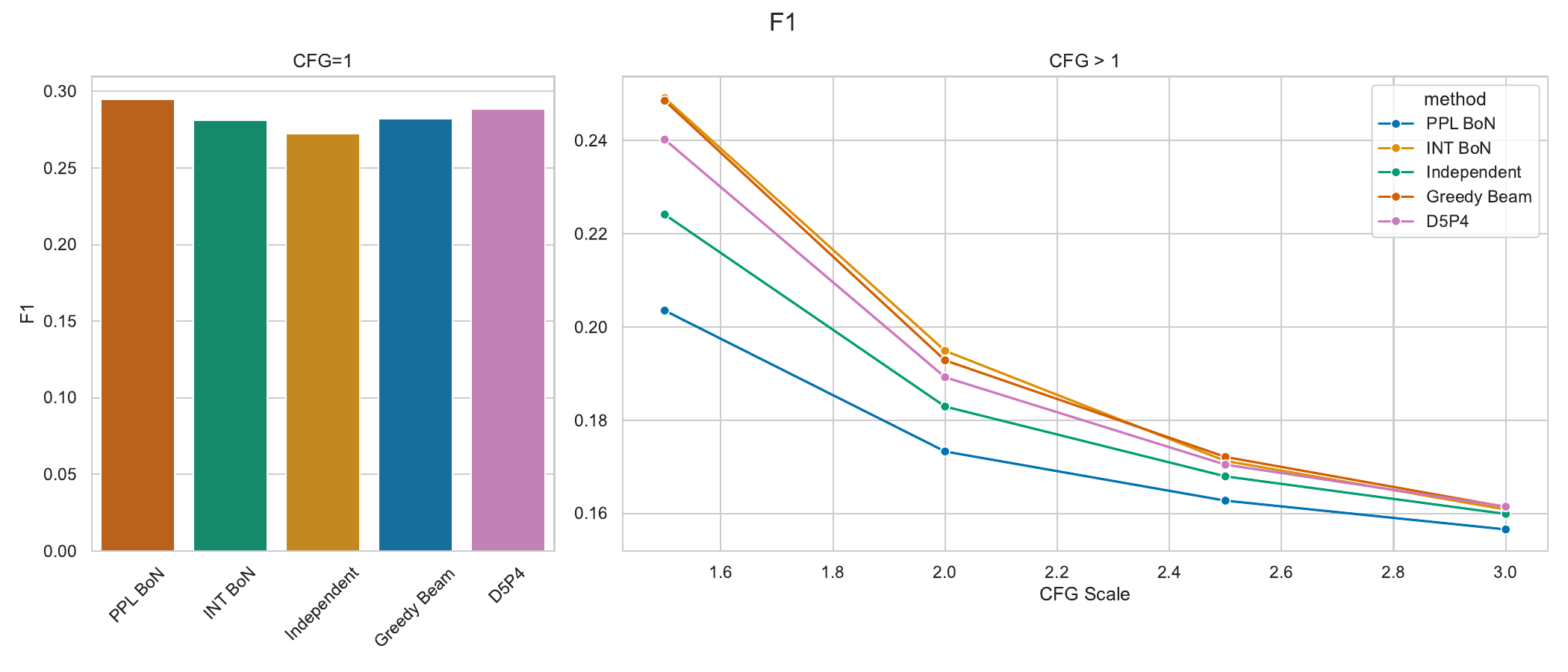}
  \end{subfigure}
  \hfill
  \begin{subfigure}{0.49\linewidth}
    \centering
    \includegraphics[width=\linewidth]{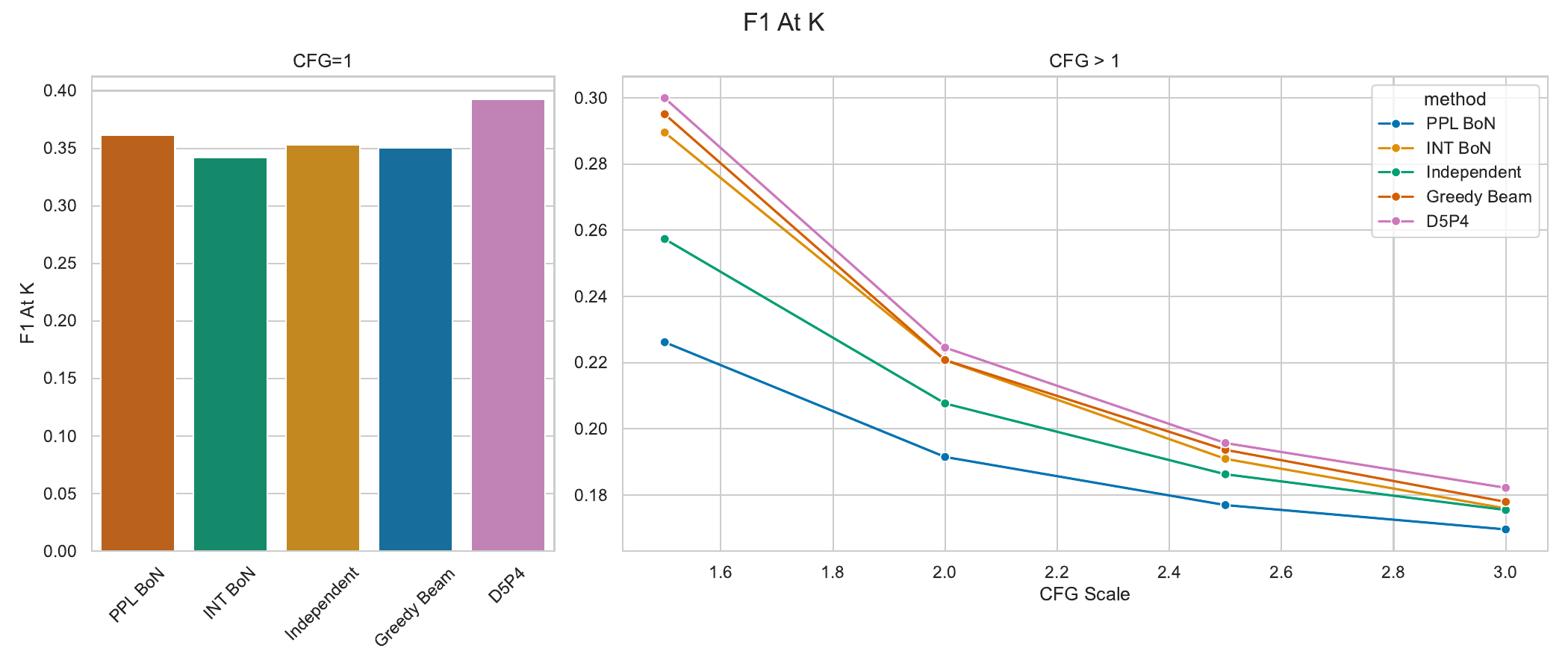}
  \end{subfigure}

  \begin{subfigure}{0.49\linewidth}
    \centering
    \includegraphics[width=\linewidth]{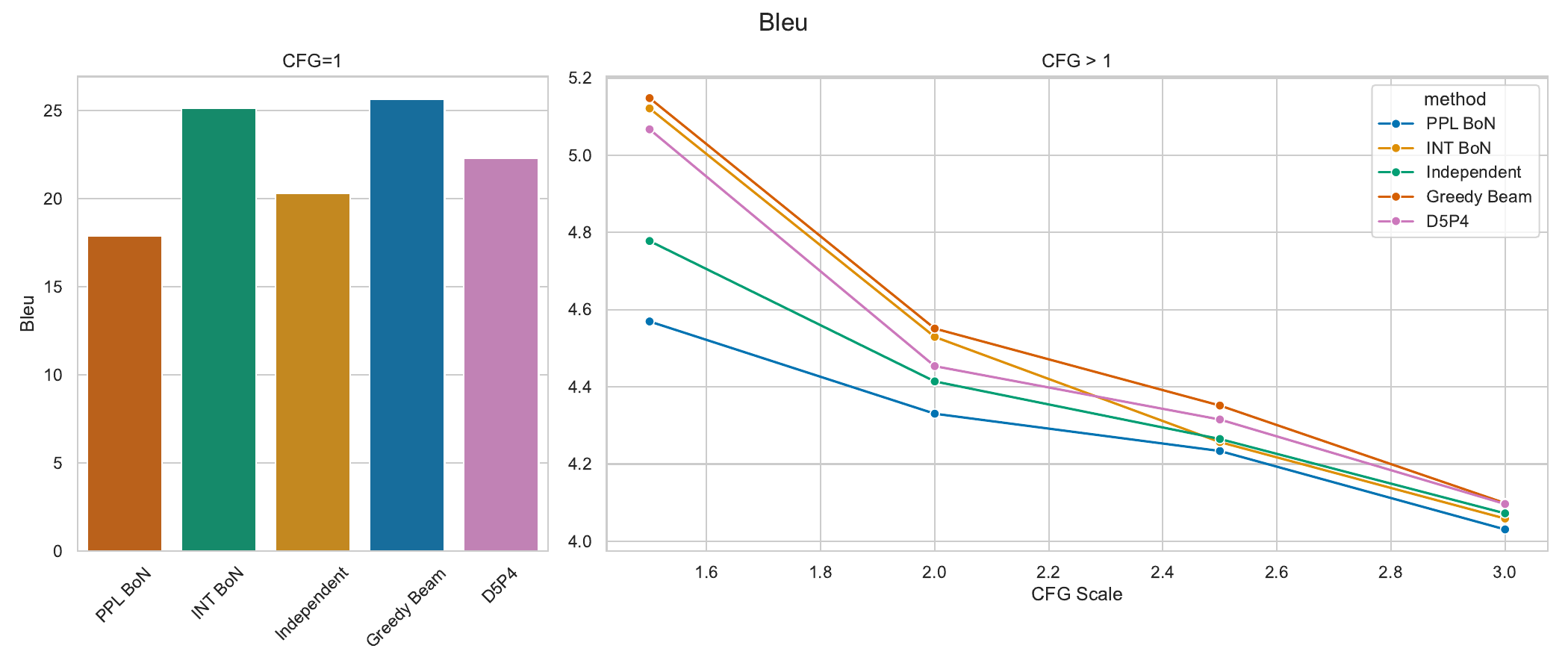}
  \end{subfigure}
  \hfill
  \begin{subfigure}{0.49\linewidth}
    \centering
    \includegraphics[width=\linewidth]{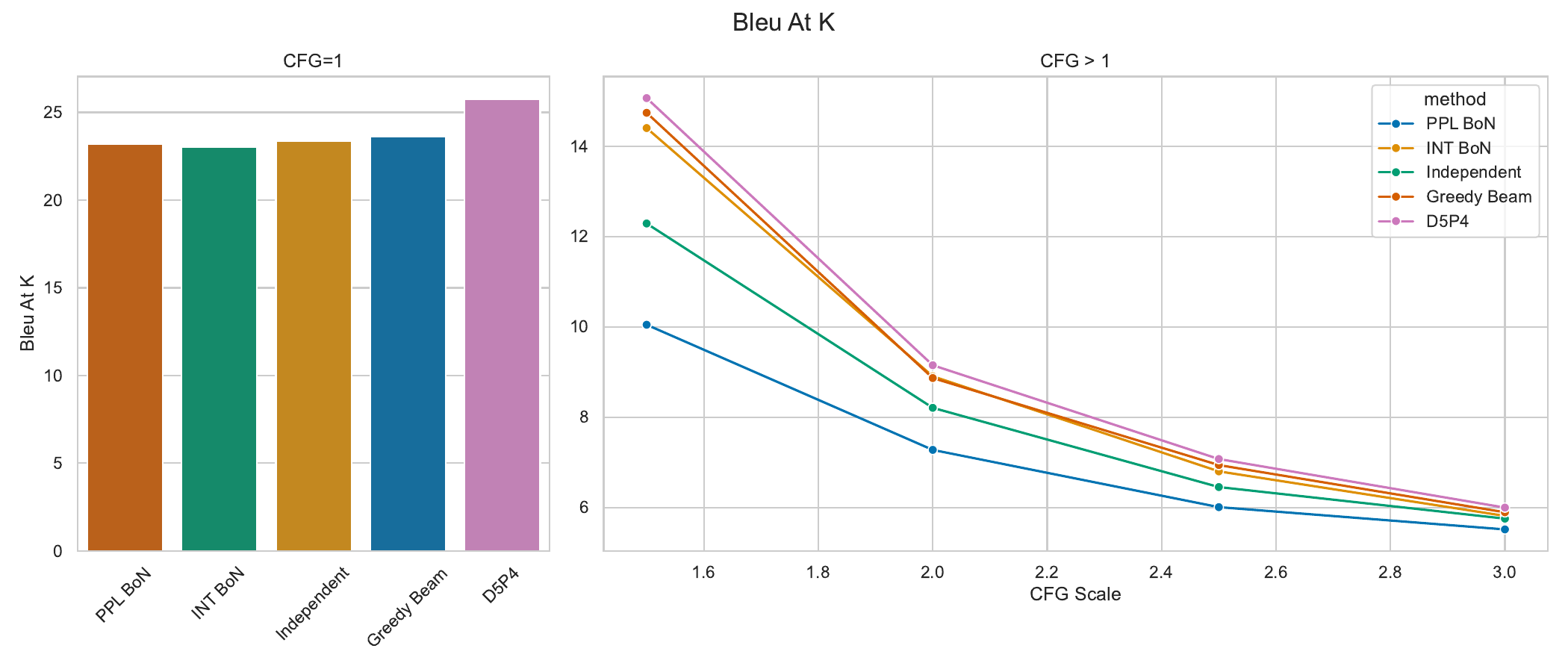}
  \end{subfigure}

  % \begin{subfigure}{0.49\linewidth}
  %   \centering
  %   \includegraphics[width=\linewidth]{plots/cfg_collapse/cos_at_k.pdf}
  % \end{subfigure}
  % \hfill
  % \begin{subfigure}{0.49\linewidth}
  %   \centering
  %   \includegraphics[width=\linewidth]{plots/cfg_collapse/wasserstein_distance.pdf}
  % \end{subfigure}

  %   \caption{Impact of CFG across quality and accuracy metrics: comparing baseline and best-of-$n$ methods with D5P4.}
  %   \label{fig:plots_quality}
  % \end{figure}

  % \begin{figure}[H]
  \ContinuedFloat
  \centering
  \begin{subfigure}{0.49\linewidth}
    \centering
    \includegraphics[width=\linewidth]{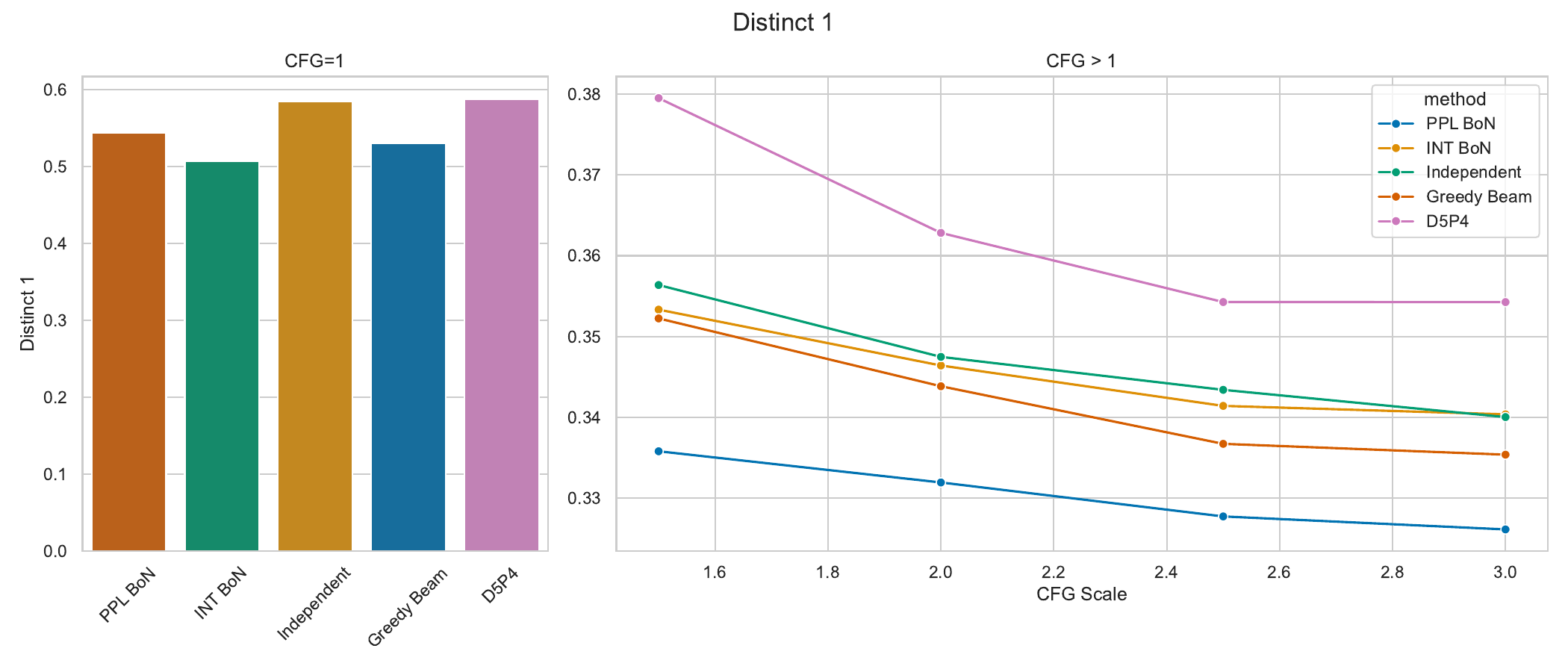}
  \end{subfigure}
  \hfill
  \begin{subfigure}{0.49\linewidth}
    \centering
    \includegraphics[width=\linewidth]{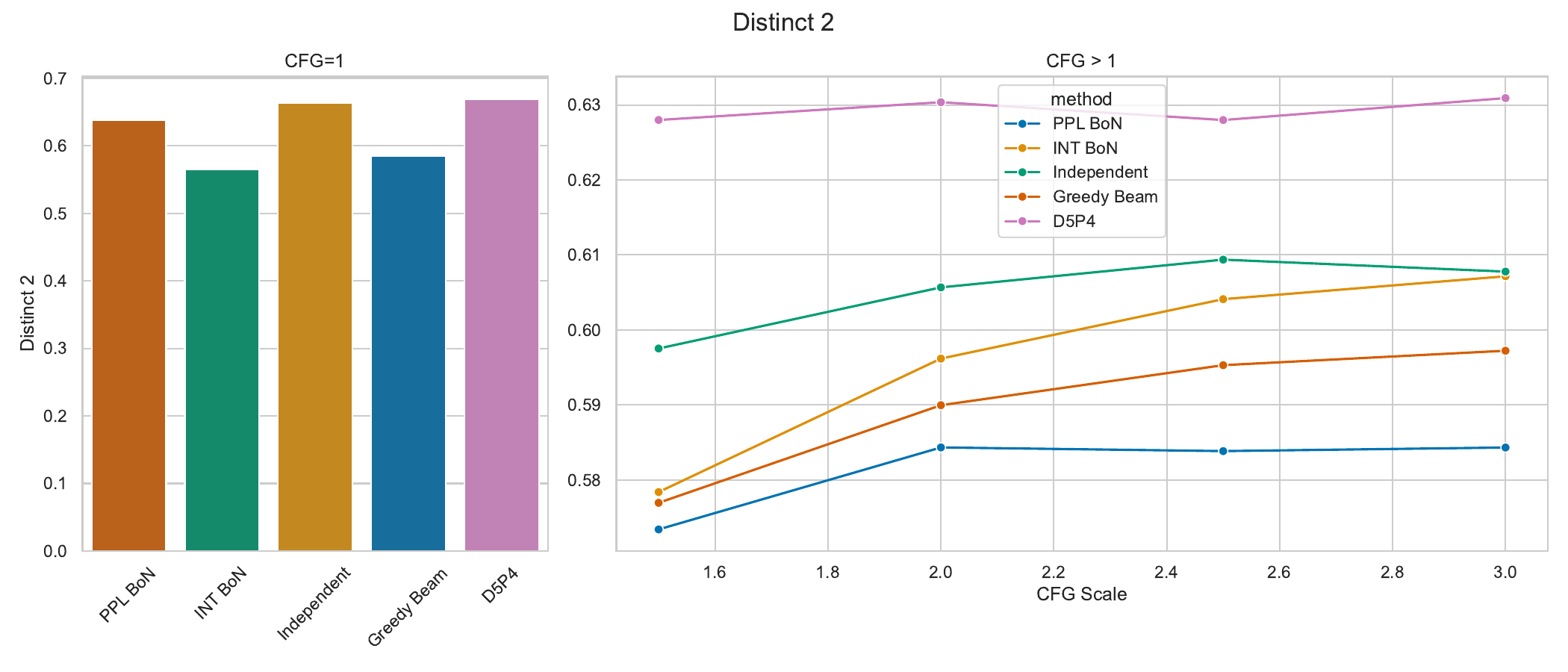}
  \end{subfigure}

  \begin{subfigure}{0.49\linewidth}
    \centering
    \includegraphics[width=\linewidth]{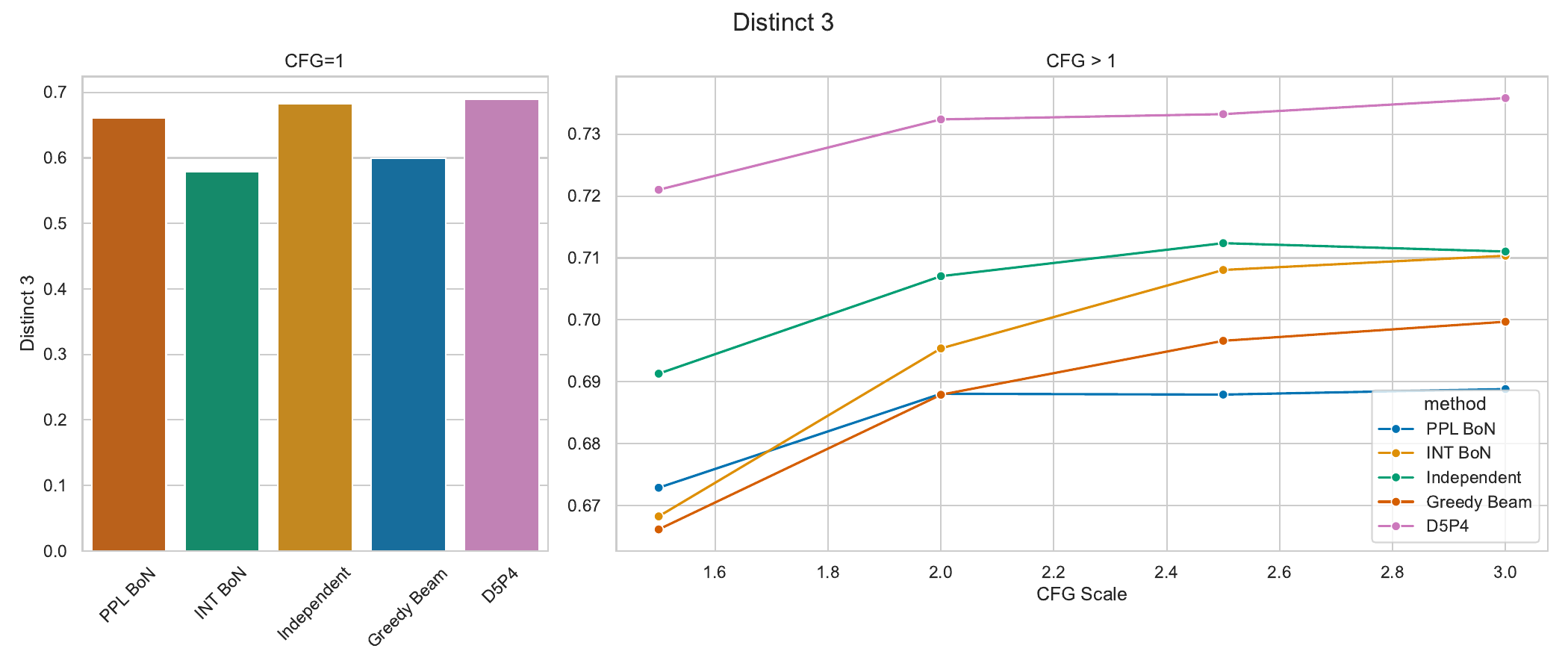}
  \end{subfigure}
  \hfill
  \begin{subfigure}{0.49\linewidth}
    \centering
    \includegraphics[width=\linewidth]{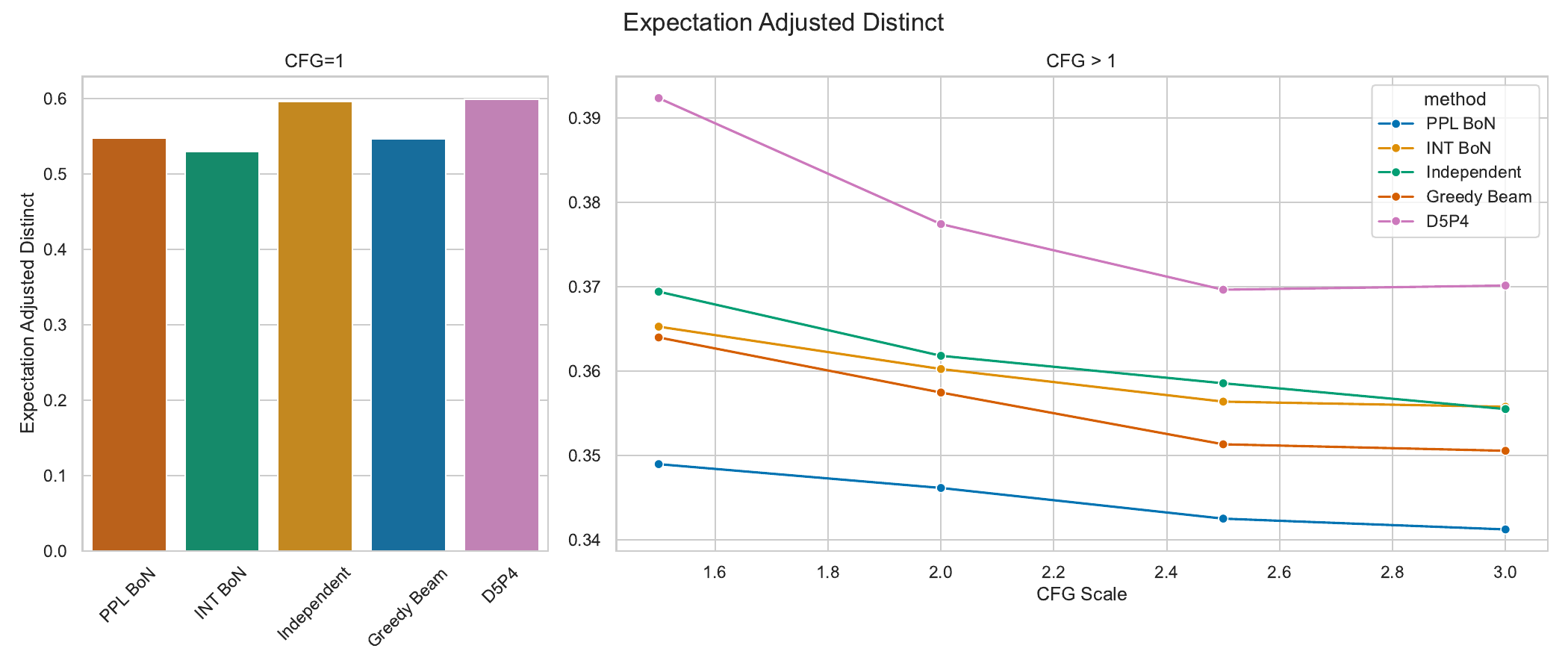}
  \end{subfigure}

  \begin{subfigure}{0.49\linewidth}
    \centering
    \includegraphics[width=\linewidth]{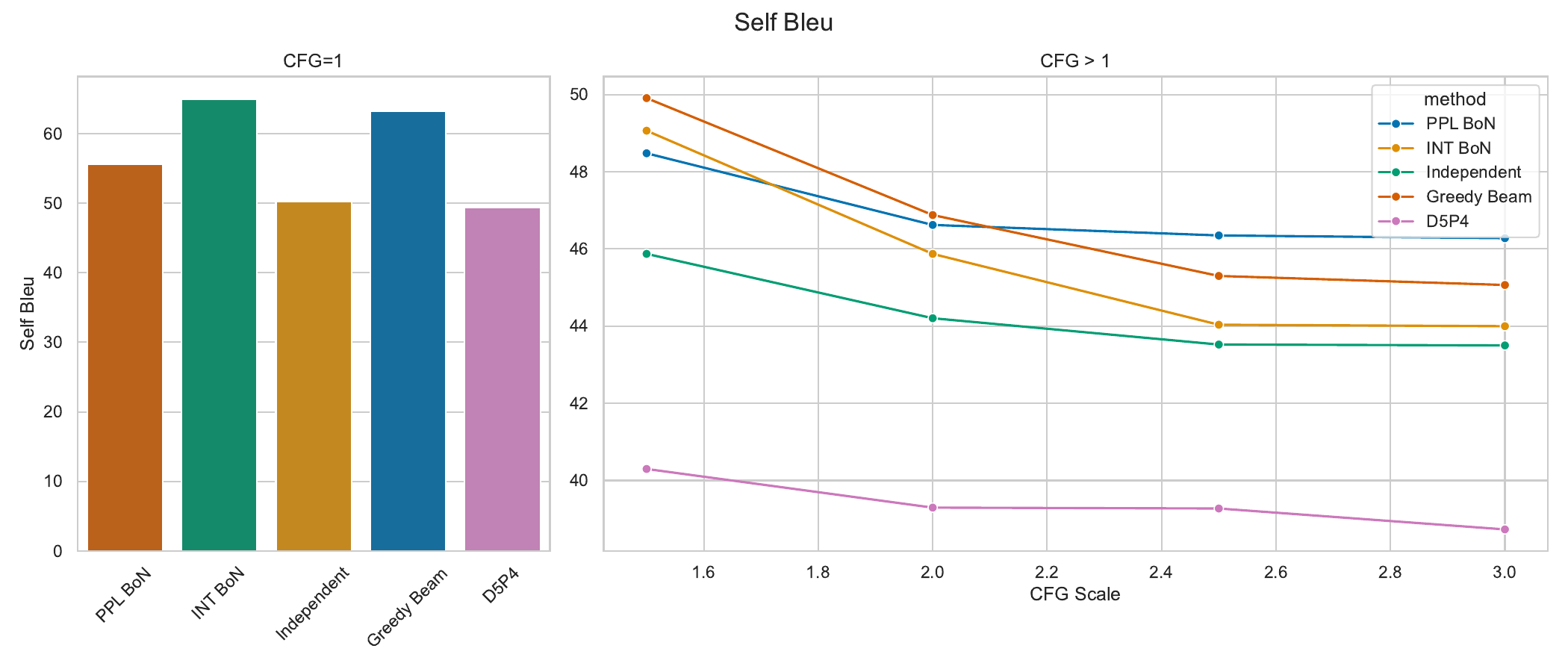}
  \end{subfigure}
  \hfill
  \begin{subfigure}{0.49\linewidth}
    \centering
    \includegraphics[width=\linewidth]{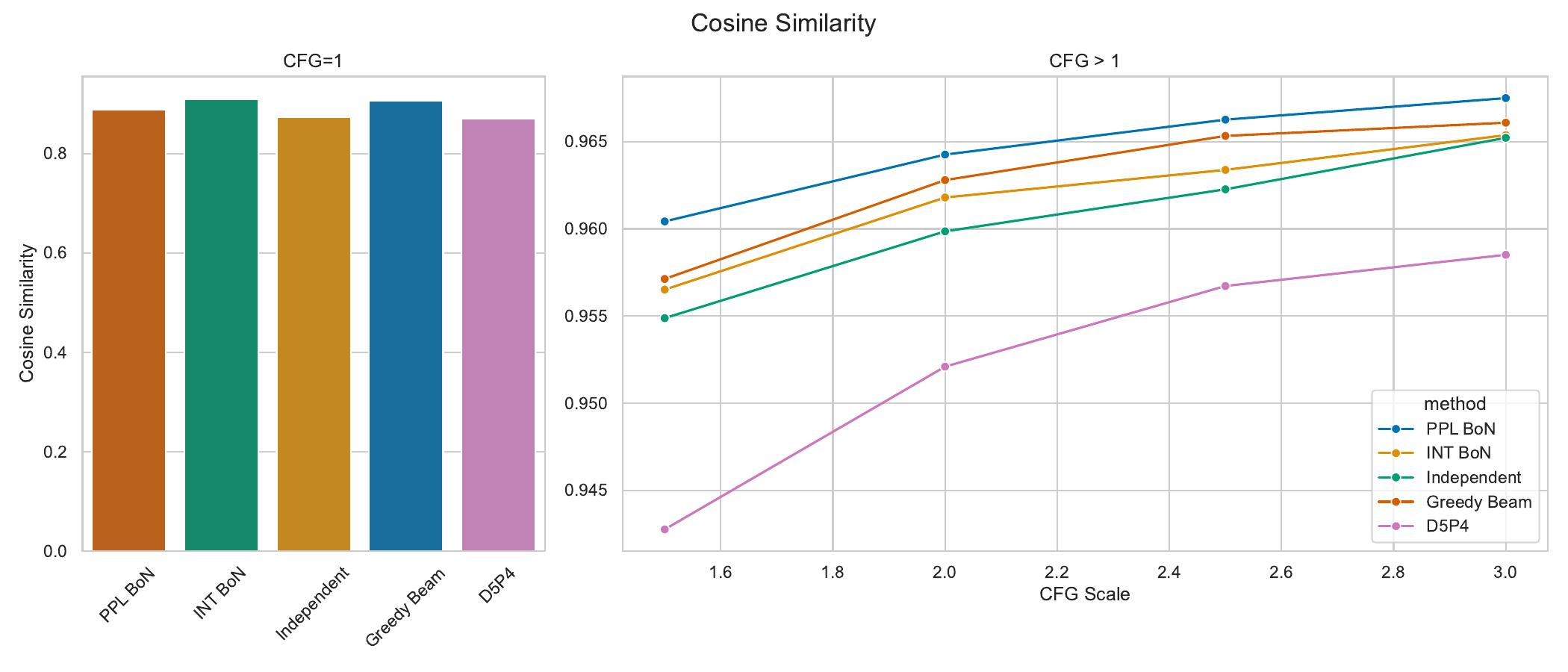}
  \end{subfigure}

  \caption{Impact of CFG across quality, accuracy, lexical and semantic diversity metrics.}
\end{figure}

\section{Ablation Studies}\label{app:ablations}

In this section, we conduct a series of ablation studies to empirically validate the core components of D5P4. We first assess whether
internal diffusion-model signals provide reliable quality
proxies for the external evaluators used in the main experiments. We then study the two design choices used to instantiate the DPP kernel:
the sequence-level scoring function and the representation pooling strategy used for diversity estimation.

\subsection{Validation of Internal Quality Signals}

\begin{table}[h]
  \centering
  \caption{Correlation between diffusion-model entropy and the autoregressive quality evaluators used in our experiments.}
  \begin{tabular}{c c c}
    Model & Entropy $\rho$ & Evaluator \\
    \toprule
    MDLM & 0.906 & GPT-2 \\
    LLaDA & 0.892 & LLaMA-3 \\
    \bottomrule
  \end{tabular}
  \label{tab:prelim_corr}
\end{table}

We first evaluate whether the internal uncertainty signals used by D5P4 align with external quality evaluators. Concretely, we compare
diffusion-model entropy with autoregressive likelihood-based signals, using GPT-2 for MDLM and LLaMA-3 for LLaDA. Table~\ref{tab:prelim_corr}
shows strong agreement in both cases, with Spearman correlations above $0.89$.

For MDLM, we additionally report the direct comparison between Monte-Carlo likelihood estimates and GPT-2 log-likelihood in
Figure~\ref{fig:mdlm_corr}. The strong linear relationship (Pearson $r=0.911$) provides a representative illustration of this alignment.
These measurements are computed on FineWeb~\citep{penedo2024finewebdatasetsdecantingweb} text for MDLM and on the TruthfulQA training set
for LLaDA, reflecting the respective domains and training regimes of the two models.

\begin{figure}[ht]
  \centering
  \begin{tikzpicture}
  \definecolor{mplblue}{HTML}{1F77B4}
  \pgfplotsset{set layers}

  \begin{axis}[
      width=\linewidth,
      height=0.5\linewidth,
      xlabel={MDLM log-likelihood},
      ylabel={GPT2 log-likelihood},
      axis line style={draw=black},
      grid=none,
      tick align=outside,
      legend pos=north west,
      legend style={draw=gray!30, fill=white, rounded corners=2pt},
    ]

    \addplot[
      only marks,
      mark=*,
      mark size=2.0pt,
      mark options={fill=mplblue, draw=Blues-9, line width=0.5pt},
      opacity=1,
    ] table[
      x={mdlm_ll},
      y={ref_ll},
      col sep=comma,
      each nth point=2,
    ] {data/correlation/likelihood.csv};
    \addlegendentry{Samples}

    \addplot[
      red,
      dashed,
      domain=-6.2:-1.8,
      samples=2,
      on layer=axis foreground
    ] {0.95*x + 0.17};
    \addlegendentry{Regression}

  \end{axis}
\end{tikzpicture}
  \caption{Correlation between MDLM-estimated log-likelihood and GPT-2 log-likelihood.}
  \label{fig:mdlm_corr}
\end{figure}

Overall, these results support the use of internal entropy-based signals as sequence-level quality proxies, without requiring external
evaluators at inference time. The representation choices used for diversity estimation are studied separately in the next subsection.

\subsection{Analysis of Scoring and Representation Choices}

Having established that internal signals can serve as effective proxies, we now compare the specific instantiations used in the DPP
kernel: the sequence-level scoring function for quality and the pooling strategy used to construct diversity representations.

\paragraph{Quality estimation.}

% We compare our entropy-based sequence score with the self-certainty measure proposed by \citet{kang_scalable_2025}. Both are evaluated by
% their correlation with GPT-2 perplexity, which serves here as a reference quality signal. The results in Table~\ref{tab:score} and
% Figure~\ref{fig:score_corr} show that entropy is substantially better aligned with the target metric, supporting its use as the quality
% term in D5P4.

We compare our entropy-based sequence score with the self-certainty measure proposed by \citet{kang_scalable_2025}. Both are
evaluated by their correlation with GPT-2 perplexity, which serves here as a reference quality signal. As illustrated in
Figure~\ref{fig:score_corr}, entropy exhibits a substantially stronger alignment with the target metric (Spearman $\rho = -0.776$) compared
to self-certainty ($\rho = -0.290$). These results empirically support the use of internal entropy as the primary quality term in the D5P4
objective.

\begin{figure}[H]
  \centering
  \begin{tikzpicture}
  \definecolor{mplblue}{HTML}{1F77B4}
  \pgfplotsset{set layers}

  \begin{groupplot}[
      group style={
        group size=2 by 1,
        horizontal sep=1.5cm,
      },
      width=0.48\linewidth,
      height=0.3\linewidth,
      axis line style={draw=black},
      grid=none,
      tick align=outside,
    ]

    % First subplot: Entropy Scores
    \nextgroupplot[
      xlabel={Entropy Score},
      ylabel={Reference PPL},
    ]
    \addplot[
      only marks,
      mark=*,
      mark size=1.5pt,
      mark options={fill=mplblue, draw=none, opacity=0.4},
    ] table[
      x={entropy_scores},
      y={ref_ppl},
      col sep=comma,
      each nth point=5,
    ] {data/correlation/scores.csv};

    % Second subplot: Self-Certainty Scores
    \nextgroupplot[
      xlabel={Self-Certainty Score},
      ylabel={Reference PPL},
    ]
    \addplot[
      only marks,
      mark=*,
      mark size=1.5pt,
      mark options={fill=mplblue, draw=none, opacity=0.4},
    ] table[
      x={self_certainty_scores},
      y={ref_ppl},
      col sep=comma,
      each nth point=5,
    ] {data/correlation/scores.csv};

  \end{groupplot}
\end{tikzpicture}
  \caption{Correlation between sequence scoring methods and the reference perplexity target.}
  \label{fig:score_corr}
\end{figure}

% \begin{table}[ht]
%   \centering
%   \caption{Correlation of sequence-level scoring methods with GPT-2 perplexity (reference quality signal).}
%   \begin{tabular}{cc}
%     Method & Correlation \\
%     \toprule
%     Self-certainty & -0.290 \\
%     Entropy & \textbf{-0.776} \\
%     \bottomrule
%   \end{tabular}
%   \label{tab:score}
% \end{table}

\paragraph{Diversity estimation.}

For the diversity term, we evaluate several pooling strategies for constructing sequence embeddings: mean pooling over all tokens, pooling
restricted to masked tokens, pooling restricted to non-masked tokens, and flattened sequence-level embeddings. We assess these choices
using Centered Kernel Alignment (CKA) with the Jina v2 embedding space, together with average cosine similarity (ACS) across masking ratios.
ACS complements CKA by revealing potential representation collapse that may not be fully reflected by alignment scores alone.

\begin{figure}[H]
  \centering
  \begin{subfigure}[b]{0.48\linewidth}
    \centering
    \resizebox{0.9\linewidth}{!}{\begin{tikzpicture}
  \definecolor{okabe_orange}{HTML}{E69F00}
  \definecolor{okabe_skyblue}{HTML}{56B4E9}
  \definecolor{okabe_bluishgreen}{HTML}{009E73}
  \definecolor{okabe_vermillion}{HTML}{D55E00}

  \begin{axis}[
      width=\linewidth,
      height=0.7\linewidth,
      axis line style={draw=black},
      grid=both,
      grid style={line width=.1pt, draw=gray!10},
      major grid style={line width=.2pt, draw=gray!20},
      tick align=outside,
      xlabel={Mask Ratio},
      ylabel={CKA Similarity},
      legend style={at={(0.5,-0.35)}, anchor=north, legend columns=2, font=\small},
      xmin=0, xmax=1,
    ]

    \addplot[
      color=okabe_skyblue,
      mark=o,
      mark size=1.5pt,
    ] table[
      x={mask_ratio},
      y={mean_cka},
      col sep=comma,
    ] {data/correlation/embeddings_mdlm.csv};
    \addlegendentry{Mean}

    \addplot[
      color=okabe_orange,
      mark=square*,
      mark size=1.2pt,
    ] table[
      x={mask_ratio},
      y={pool_non_masked_cka},
      col sep=comma,
    ] {data/correlation/embeddings_mdlm.csv};
    \addlegendentry{Pool Non-Masked}

    \addplot[
      color=okabe_bluishgreen,
      mark=triangle*,
      mark size=1.5pt,
    ] table[
      x={mask_ratio},
      y={pool_masked_cka},
      col sep=comma,
    ] {data/correlation/embeddings_mdlm.csv};
    \addlegendentry{Pool Masked}

    \addplot[
      color=okabe_vermillion,
      mark=diamond*,
      mark size=1.5pt,
    ] table[
      x={mask_ratio},
      y={flatten_cka},
      col sep=comma,
    ] {data/correlation/embeddings_mdlm.csv};
    \addlegendentry{Flatten}

  \end{axis}
\end{tikzpicture}}
    \caption{MDLM CKA}
    \label{fig:mdlm_cka}
  \end{subfigure}
  \hfill
  \begin{subfigure}[b]{0.48\linewidth}
    \centering
    \resizebox{0.9\linewidth}{!}{\begin{tikzpicture}
  \definecolor{okabe_orange}{HTML}{E69F00}
  \definecolor{okabe_skyblue}{HTML}{56B4E9}
  \definecolor{okabe_bluishgreen}{HTML}{009E73}
  \definecolor{okabe_vermillion}{HTML}{D55E00}

  \begin{axis}[
      width=\linewidth,
      height=0.7\linewidth,
      axis line style={draw=black},
      grid=both,
      grid style={line width=.1pt, draw=gray!10},
      major grid style={line width=.2pt, draw=gray!20},
      tick align=outside,
      xlabel={Mask Ratio},
      ylabel={Average Cosine Similarity},
      legend style={at={(0.5,-0.35)}, anchor=north, legend columns=2, font=\small},
      xmin=0, xmax=1,
    ]

    \addplot[
      color=okabe_skyblue,
      mark=o,
      mark size=1.5pt,
    ] table[
      x={mask_ratio},
      y={mean_acs},
      col sep=comma,
    ] {data/correlation/embeddings_mdlm.csv};
    \addlegendentry{Mean}

    \addplot[
      color=okabe_orange,
      mark=square*,
      mark size=1.2pt,
    ] table[
      x={mask_ratio},
      y={pool_non_masked_acs},
      col sep=comma,
    ] {data/correlation/embeddings_mdlm.csv};
    \addlegendentry{Pool Non-Masked}

    \addplot[
      color=okabe_bluishgreen,
      mark=triangle*,
      mark size=1.5pt,
    ] table[
      x={mask_ratio},
      y={pool_masked_acs},
      col sep=comma,
    ] {data/correlation/embeddings_mdlm.csv};
    \addlegendentry{Pool Masked}

    \addplot[
      color=okabe_vermillion,
      mark=diamond*,
      mark size=1.5pt,
    ] table[
      x={mask_ratio},
      y={flatten_acs},
      col sep=comma,
    ] {data/correlation/embeddings_mdlm.csv};
    \addlegendentry{Flatten}

  \end{axis}
\end{tikzpicture}}
    \caption{MDLM ACS}
    \label{fig:mdlm_acs}
  \end{subfigure}

  \vspace{0.5em}

  \begin{subfigure}[b]{0.48\linewidth}
    \centering
    \resizebox{0.9\linewidth}{!}{\begin{tikzpicture}
  \definecolor{okabe_orange}{HTML}{E69F00}
  \definecolor{okabe_skyblue}{HTML}{56B4E9}
  \definecolor{okabe_bluishgreen}{HTML}{009E73}
  \definecolor{okabe_vermillion}{HTML}{D55E00}

  \begin{axis}[
      width=\linewidth,
      height=0.7\linewidth,
      axis line style={draw=black},
      grid=both,
      grid style={line width=.1pt, draw=gray!10},
      major grid style={line width=.2pt, draw=gray!20},
      tick align=outside,
      xlabel={Mask Ratio},
      ylabel={CKA Similarity},
      legend style={at={(0.5,-0.35)}, anchor=north, legend columns=2, font=\small},
      xmin=0, xmax=1,
    ]

    \addplot[
      color=okabe_skyblue,
      mark=o,
      mark size=1.5pt,
    ] table[
      x={mask_ratio},
      y={mean_cka},
      col sep=comma,
    ] {data/correlation/embeddings_llada.csv};
    \addlegendentry{Mean}

    \addplot[
      color=okabe_orange,
      mark=square*,
      mark size=1.2pt,
    ] table[
      x={mask_ratio},
      y={pool_non_masked_cka},
      col sep=comma,
    ] {data/correlation/embeddings_llada.csv};
    \addlegendentry{Pool Non-Masked}

    \addplot[
      color=okabe_bluishgreen,
      mark=triangle*,
      mark size=1.5pt,
    ] table[
      x={mask_ratio},
      y={pool_masked_cka},
      col sep=comma,
    ] {data/correlation/embeddings_llada.csv};
    \addlegendentry{Pool Masked}

    \addplot[
      color=okabe_vermillion,
      mark=diamond*,
      mark size=1.5pt,
    ] table[
      x={mask_ratio},
      y={flatten_cka},
      col sep=comma,
    ] {data/correlation/embeddings_llada.csv};
    \addlegendentry{Flatten}

  \end{axis}
\end{tikzpicture}}
    \caption{LLaDA CKA}
    \label{fig:LLaDA_cka}
  \end{subfigure}
  \hfill
  \begin{subfigure}[b]{0.48\linewidth}
    \centering
    \resizebox{0.9\linewidth}{!}{\begin{tikzpicture}
  \definecolor{okabe_orange}{HTML}{E69F00}
  \definecolor{okabe_skyblue}{HTML}{56B4E9}
  \definecolor{okabe_bluishgreen}{HTML}{009E73}
  \definecolor{okabe_vermillion}{HTML}{D55E00}

  \begin{axis}[
      width=\linewidth,
      height=0.7\linewidth,
      axis line style={draw=black},
      grid=both,
      grid style={line width=.1pt, draw=gray!10},
      major grid style={line width=.2pt, draw=gray!20},
      tick align=outside,
      xlabel={Mask Ratio},
      ylabel={Average Cosine Similarity},
      legend style={at={(0.5,-0.35)}, anchor=north, legend columns=2, font=\small},
      xmin=0, xmax=1,
    ]

    \addplot[
      color=okabe_skyblue,
      mark=o,
      mark size=1.5pt,
    ] table[
      x={mask_ratio},
      y={mean_acs},
      col sep=comma,
    ] {data/correlation/embeddings_llada.csv};
    \addlegendentry{Mean}

    \addplot[
      color=okabe_orange,
      mark=square*,
      mark size=1.2pt,
    ] table[
      x={mask_ratio},
      y={pool_non_masked_acs},
      col sep=comma,
    ] {data/correlation/embeddings_llada.csv};
    \addlegendentry{Pool Non-Masked}

    \addplot[
      color=okabe_bluishgreen,
      mark=triangle*,
      mark size=1.5pt,
    ] table[
      x={mask_ratio},
      y={pool_masked_acs},
      col sep=comma,
    ] {data/correlation/embeddings_llada.csv};
    \addlegendentry{Pool Masked}

    \addplot[
      color=okabe_vermillion,
      mark=diamond*,
      mark size=1.5pt,
    ] table[
      x={mask_ratio},
      y={flatten_acs},
      col sep=comma,
    ] {data/correlation/embeddings_llada.csv};
    \addlegendentry{Flatten}

  \end{axis}
\end{tikzpicture}}
    \caption{LLaDA ACS}
    \label{fig:LLaDA_acs}
  \end{subfigure}
  \caption{CKA and Average Cosine Similarity (ACS) of pooling methods across masking ratios for MDLM and LLaDA.}
  \label{fig:mdlm_LLaDA_ablation}
\end{figure}

As shown in Table~\ref{tab:cka} and Figure~\ref{fig:mdlm_LLaDA_ablation}, flattened embeddings provide the strongest alignment for
both MDLM and LLaDA, which motivates the representation choice adopted in the main experiments.

\begin{table}[H]
  \centering
  \caption{Representation alignment (CKA) with the reference model for different pooling methods.}
  \begin{tabular}{ccc}
    Method & MDLM & LLaDA \\
    \toprule
    Mean & 0.777 & 0.482 \\
    Non-masked & 0.660 & 0.536 \\
    Masked & 0.710 & 0.435 \\
    Flatten & \textbf{0.821} & \textbf{0.667} \\
    \bottomrule
  \end{tabular}
  \label{tab:cka}
\end{table}

\section{Role of the Partition Constraint}\label{app:partition-ablation}

To isolate the individual contributions of our selection mechanism and our partition constraint, we evaluate an unconstrained variant of
our method, denoted \textbf{D5P3}. Relaxing the partition constraint yields an unconstrained interacting-particle selection procedure,
making the comparison to SMC methods more direct (provided the number of groups equals the number of elements per group). This structural
equivalence allows for a direct empirical comparison against the E-SMC baseline~\cite{chen2025optimizing} on the MDLM open-ended generation
task. To ensure a strictly fair comparison, we match the compute budget across both methods: E-SMC uses 4 particles, D5P3 uses 2 groups of
2 candidates. For E-SMC, we vary the resampling interval up to the limit of a single resampling per trajectory. For D5P3, we sweep the
interaction parameter $\beta$. Performance is measured using the perplexity and pairwise cosine similarity metrics established in
Section~\ref{subsec:open-ended-gen}.

\begin{figure}[H]
  \centering
  \includegraphics[width=\linewidth]{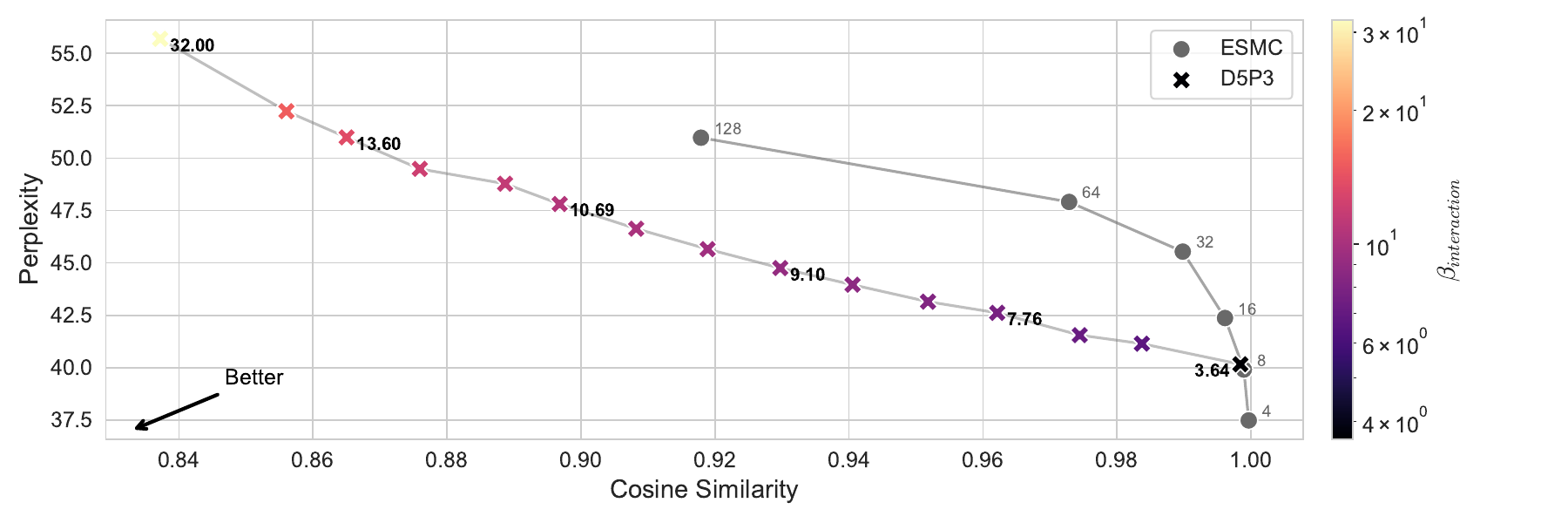}
  \caption{
    Quality--diversity trade-off comparison between E-SMC and our unconstrained variant (D5P3) on MDLM. While D5P3 dominates the
    non-collapsed regime, both methods ultimately suffer from lineage collapse at extreme parameters, highlighting the necessity of the
    partition constraint.
  }
  \label{fig:np_esmc_comparison}
\end{figure}

As shown in Figure~\ref{fig:np_esmc_comparison}, D5P3 Pareto-dominates E-SMC over the non-collapsed regime (cosine similarity $< 1$).
This indicates that even without the partition constraint, our diversity-aware selection mechanism is significantly more effective at balancing
the quality--diversity trade-off than standard E-SMC resampling. However, despite this improved efficiency, unconstrained search remains
fundamentally vulnerable to lineage collapse. The E-SMC frontier inherently includes a fully collapsed endpoint ($\text{cosine similarity}
= 1$), representing zero effective set diversity. We observe this exact structural failure in D5P3: low values of $\beta$ (low interaction)
similarly lead to a collapsed set of sequences with a cosine similarity close to 1. Furthermore, when matching the broader experiment
parameters from Section~\ref{subsec:open-ended-gen}, even a high interaction parameter ($\beta = 30$) fails to prevent the relaxed search
from converging to the same sequence across all beams, yielding a degraded perplexity of 22.24 and a cosine similarity of 0.7835. Together,
these results demonstrate that while our scoring mechanism constitutes a strict improvement over prior unconstrained methods, explicitly
enforcing the partition constraint (as in D5P4) is absolutely necessary to prevent lineage collapse and maintain reliable batch diversity.

\section{Algorithmic details of the Search Methods}\label{app:algorithms}

We summarize the main subset-selection rules considered in this work: a quality-only selector, an MMR-style diverse beam selector, and the
greedy MAP-DPP selector underlying D5P4. For clarity, Algorithms~\ref{alg:greedy_quality}--\ref{alg:greedy_map_partition} present the core
decision rules in their conceptual single-trajectory form under transversal constraints.

In the transversal setting, the candidate set is partitioned into groups and the selector must return exactly one item per group. We denote
by $s_i$ the quality score of candidate $i$, by $x_i \in \mathbb{R}^{LD}$ its flattened normalized embedding, by $g(i)$ its group index,
and by $K$ the number of groups to select.

\begin{algorithm}[H]
  \caption{Greedy Quality-Only Selection (partitioned)}
  \label{alg:greedy_quality}
  \begin{algorithmic}[1]
    \STATE \textbf{Input:} Scores $(s_i)_{i=1}^N$, group labels $(g(i))_{i=1}^N$, target size $K$
    \STATE $Y \leftarrow \emptyset$
    \FOR{$m = 1$ \textbf{to} $K$}
    \STATE $j \leftarrow \arg\max_{i:\, g(i)=m} s_i$
    \STATE $Y \leftarrow Y \cup \{j\}$
    \ENDFOR
    \STATE \textbf{Return:} $Y$
  \end{algorithmic}
\end{algorithm}

\begin{algorithm}[H]
  \caption{MMR-Style Diverse Beam Search (partitioned)}
  \label{alg:mmr_partition}
  \begin{algorithmic}[1]
    \STATE \textbf{Input:} Scores $(s_i)_{i=1}^N$, normalized embeddings $(x_i)_{i=1}^N$, trade-off $\alpha$, group labels
    $(g(i))_{i=1}^N$, target size $K$
    \STATE $Y \leftarrow \emptyset$
    \FOR{$t = 1$ \textbf{to} $K$}
    \STATE $Z_{\mathrm{valid}} \leftarrow \{i \mid g(i) \notin g(Y)\}$
    \IF{$t = 1$}
    \STATE $j \leftarrow \arg\max_{i \in Z_{\mathrm{valid}}} s_i$
    \ELSE
    \STATE $\bar{x}(Y) \leftarrow \frac{1}{|Y|}\sum_{\ell \in Y} x_\ell$
    \STATE $j \leftarrow \arg\max_{i \in Z_{\mathrm{valid}}} \left[s_i - \alpha \langle \bar{x}(Y), x_i \rangle \right]$
    \ENDIF
    \STATE $Y \leftarrow Y \cup \{j\}$
    \ENDFOR
    \STATE \textbf{Return:} $Y$
  \end{algorithmic}
\end{algorithm}

\begin{algorithm}[H]
  \caption{Greedy MAP Inference for Partition DPPs}
  \label{alg:greedy_map_partition}
  \begin{algorithmic}[1]
    \STATE \textbf{Input:} Kernel $\mathbf{L} \in \mathbb{R}^{N \times N}$, group labels $(g(i))_{i=1}^N$, target size $K$
    \STATE \textbf{Initialize:} $Y \leftarrow \emptyset$, $\mathbf{c}_i \leftarrow \emptyset$, $d_i^2 \leftarrow \mathbf{L}_{ii}$ for all $i$
    \FOR{$t = 1$ \textbf{to} $K$}
    \STATE $Z_{\mathrm{valid}} \leftarrow \{i \mid g(i) \notin g(Y)\}$
    \STATE $j \leftarrow \arg\max_{i \in Z_{\mathrm{valid}}} d_i^2$
    \STATE $Y \leftarrow Y \cup \{j\}$
    \IF{$t < K$}
    \FOR{$i \in Z_{\mathrm{valid}} \setminus \{j\}$}
    \STATE $e_i \leftarrow \bigl(\mathbf{L}_{ji} - \langle \mathbf{c}_j,\mathbf{c}_i \rangle \bigr)/\sqrt{d_j^2}$
    \STATE $\mathbf{c}_i \leftarrow [\mathbf{c}_i \;\; e_i]$
    \STATE $d_i^2 \leftarrow d_i^2 - e_i^2$
    \ENDFOR
    \ENDIF
    \ENDFOR
    \STATE \textbf{Return:} $Y$
  \end{algorithmic}
\end{algorithm}

\paragraph{Implementation note.}
The implementation used in our experiments replaces Algorithms~\ref{alg:mmr_partition} and~\ref{alg:greedy_map_partition} by
full-exploration parallel variants. Concretely, one trajectory is initialized from each possible first item, all trajectories are updated in
parallel on the GPU, and the final subset is chosen as the one with the highest cumulative MMR objective (for Diverse Beam Search) or the
highest cumulative log-determinant (for D5P4). This preserves the underlying local selection rule while improving solution quality in
practice. Since the computation is done on GPU, this multi-start exploration can be batched, effectively achieving a near-zero overhead.

\paragraph{Complexity of the selectors.}
We now summarize the asymptotic cost of the implemented selectors. Let $N$ denote the number of selector candidates, $k$ the number of
selected groups/items, $w$ the number of elements per group, $D$ the hidden dimension per token, and $L$ the number of token positions
retained in the selector embedding. In the transversal setting, the typical candidate count is $N = kw$.
Since the selector receives block-local embeddings, its effective flattened embedding dimension is $D_{\mathrm{eff}} = LD$,
where $L$ is the block length in blockwise diffusion and the full generated length in pure diffusion. Prompt tokens affect the model
forward sequence length, but not the selector embedding dimension.

Algorithm~\ref{alg:greedy_quality} has negligible selector overhead in the transversal setting once the quality scores have been computed,
since it reduces to independent per-group maximization.

For the full-exploration implementation of MMR-style Diverse Beam Search, the dominant costs are:
\[
  \text{flatten + normalize: } O(NLD),
  \text{per-step diversity penalty: } O(N^2LD),
  \text{$k$ selection steps: } O(kN^2LD).
\]
Therefore,
\[
  T_{\mathrm{DivBS}} = O(kN^2LD) = O(k^3 w^2 LD)
\]
after substituting $N = kw$. Its memory usage is mainly $O(NLD + N^2)$,
corresponding to flattened embeddings together with the search state and pairwise bookkeeping.

For the full-exploration implementation of D5P4, the dominant costs are:
\[
  \text{flatten + normalize: } O(NLD),
  \text{kernel construction: } O(N^2LD),
  \text{greedy MAP selection: } O(N^2k^2).
\]
Therefore,
\[
  T_{\mathrm{D5P4}} = O\!\bigl(N^2(LD + k^2)\bigr),
\]
which becomes
\[
  T_{\mathrm{D5P4}} = O\!\bigl(k^2 w^2 LD + k^4 w^2\bigr)
\]
after substituting $N = kw$. Memory is at least $O(N^2)$ for the kernel, with additional work buffers scaling roughly as $O(N^2k)$ in the
Triton full-exploration path.

This comparison clarifies the runtime differences observed in practice and verified in the next section. D5P4 pays the embedding
interaction cost once through explicit kernel construction, then performs greedy MAP updates on the resulting kernel. By contrast, Diverse
Beam Search avoids explicit kernel materialization but repeatedly recomputes embedding-based diversity penalties over $k$ selection steps.
As a result, when $LD \gg k^2$, the repeated interaction term can make Diverse Beam Search asymptotically more expensive than D5P4. This
regime is especially relevant in pure diffusion, where $L$ is the full generated sequence length rather than a short local block.

\section{Scaling Behavior of the Subset-Selection Algorithms}\label{app:scaling}

We study the scaling behavior of the subset-selection algorithms underlying D5P4 in a controlled synthetic setting. The goal is to isolate
the computational and optimization properties of the search procedure itself, independently of model forward-pass costs, and to evaluate
how these properties evolve as the combinatorial complexity of the partition-constrained selection problem increases.

We construct synthetic DPP kernels matching real distribution characteristics. Concretely, we estimate first- and
second-order statistics from MDLM hidden representations and associated quality scores collected from real decoding trajectories. We then
sample synthetic embeddings and quality values from these statistics, and build additive kernels combining a quality term with a
similarity-based interaction term, following the D5P4 formulation. This yields controlled instances that preserve the statistical structure
of realistic decoding states while allowing systematic scaling sweeps.

For each synthetic instance, we compare methods using three complementary metrics:
(i) the \emph{normalized objective value}, defined as the achieved subdeterminant normalized across methods for the same instance;
(ii) the \emph{average rank} of each method across trials; and
(iii) the \emph{execution time} of the selection step alone.
The first two quantify the quality of the selected subset under the DPP objective, while the third isolates the computational cost of the
search procedure.

We vary both the number of groups and the group size from 4 to 64, yielding increasingly difficult partition-constrained selection
problems. The corresponding combinatorial complexity is $\log |\mathcal{S}| = \log\!\bigl(\text{group size}^{\text{number of groups}}\bigr)$.

All results are averaged over several values of the interaction parameter controlling the strength of pairwise repulsion in the kernel and 500
independent trials.

We compare the following methods: Random selection, Greedy Beam Search, Diverse Beam Search (DivBS), a DPPy-based spectral DPP reference,
our standard D5P4 solver, and a Triton-optimized implementation of D5P4. All methods except DPPy are implemented on the GPU. Since DPPy
relies on NumPy and does not support transversal partition constraints, it is included only as a reference point rather than as a directly
comparable constrained solver.

Figure~\ref{fig:speed_scaling} summarizes the results. In terms of objective quality, DPPy and Greedy Beam Search remain close to the
random baseline, while DivBS is competitive only at relatively low combinatorial complexity. As complexity increases, D5P4 consistently
achieves stronger normalized objectives and better average ranks, indicating that its approximation remains effective in large
search spaces.

In terms of runtime, Random and Greedy Beam Search incur minimal overhead, whereas the CPU-based DPPy reference scales poorly and rapidly
becomes impractical. DivBS also becomes increasingly expensive as complexity grows. By contrast, the GPU implementation of D5P4 scales more
favorably, and the Triton-optimized variant further reduces selection time, highlighting the benefit of specialized kernels for large-scale
subset selection.

%Overall, this controlled scaling study shows that D5P4 combines strong subset-selection quality with favorable computational scaling,
%especially in the high-complexity regimes that are most relevant to large-batch decoding.

\begin{figure}[H]
  \centering
  \input{figures/method_bench}
  \caption{Scaling behavior of subset-selection methods as a function of log combinatorial complexity. We report normalized objective value
    and selection time across synthetic partitioned DPP instances constructed from MDLM statistics. D5P4 achieves the strongest objective
  values in high-complexity regimes while scaling more favorably than alternative diversity-aware baselines.}
  \label{fig:speed_scaling}
\end{figure}

\section{Theoretical link between DPP MAP and cosine similarity}\label{app:dpp-theory}

In this section, we provide a formal link between maximizing the DPP objective and minimizing average pairwise cosine similarity. While
finding the exact global maximum of the DPP objective is NP-hard, the fast greedy MAP algorithm provides a
$(1 - 1/e)$-approximation for submodular functions under certain conditions.

As exploring the full exponential state-space of possible trajectories is computationally intractable, we do not claim global optimality
across the entire denoising process. Instead, we establish a local connection between the determinant and similarity in the following regimes.

\subsection{Weak-correlation regime}

We first consider the interaction-only variant of D5P4+, obtained either by removing the quality term or, equivalently, in the limit where
the interaction coefficient tends to infinity. For a subset of size $k$, the corresponding kernel can be written as
\[
  K = I + C,
\]
where $C$ is symmetric, satisfies $c_{ii}=0$, and has off-diagonal entries $c_{ij} = \cos(x_i,x_j)$ for $i \neq j$.

To relate the determinant to pairwise similarity, we expand $\det(I+C)$ around $C=0$. Using
\[
  \det(I+C) = \exp\!\bigl(\operatorname{tr}(\log(I+C))\bigr),
\]
together with the Mercator expansion of the matrix logarithm, valid for $\|C\|_2 < 1$, yields the third-order expansion
\[
  \det(I+C)
  = 1 + \operatorname{tr}(C)
  + \frac{1}{2}\!\left(\operatorname{tr}(C)^2 - \operatorname{tr}(C^2)\right)
  + \frac{1}{6}\!\left(\operatorname{tr}(C)^3 - 3\operatorname{tr}(C)\operatorname{tr}(C^2) + 2\operatorname{tr}(C^3)\right)
  + \mathcal{O}(\|C\|_2^4).
\]

Since $c_{ii}=0$, we have $\operatorname{tr}(C)=0$, and therefore
\[
  \det(I+C)
  = 1 - \frac{1}{2}\operatorname{tr}(C^2) + \frac{1}{3}\operatorname{tr}(C^3) + \mathcal{O}(\|C\|_2^4).
\]
By symmetry and the vanishing diagonal,
\[
  \operatorname{tr}(C^2)
  = \sum_{i,j} c_{ij}c_{ji}
  = 2 \sum_{i<j} c_{ij}^2,
\]
which gives
\[
  \det(I+C)
  = 1 - \sum_{i<j} c_{ij}^2 + \frac{1}{3}\operatorname{tr}(C^3) + \mathcal{O}(\|C\|_2^4).
\]

In the weak-correlation regime, the cubic and higher-order terms are negligible, leading to the approximation
\[
  \sum_{i<j} c_{ij}^2 \approx 1 - \det(I+C).
\]

Let $m=\binom{k}{2}$ denote the number of unordered pairs. By Cauchy--Schwarz,
\[
  \frac{1}{m}\sum_{i<j} c_{ij}
  \leq
  \sqrt{\frac{1}{m}\sum_{i<j} c_{ij}^2}
  \approx
  \sqrt{\frac{1-\det(I+C)}{m}}.
\]

Hence, in this regime, maximizing $\det(I+C)$ minimizes an upper bound on the average pairwise cosine similarity. While this does not imply
exact minimization of the mean cosine in general, it establishes a direct local connection between the DPP objective and diversity as
measured by pairwise similarity.

\subsection{Accounting for shared structural tokens}

For MDLM, the weak-correlation assumption is not always satisfied empirically. A key reason is that all sequences share structural tokens
such as \texttt{<bos>} and \texttt{<eos>}, which induce a common component in the sequence embeddings and artificially increase raw cosine
similarities.

To account for this effect, we decompose each embedding as
\[
  e_i = \sqrt{\rho}\,u + \sqrt{1-\rho}\,z_i,
\]
where $u$ is a shared unit-norm component, $\rho \in [0,1]$ quantifies the contribution of the shared structure, and $z_i$ captures the
sequence-specific residual.

Under this decomposition, the kernel takes the form
\[
  K = B_\rho + (1-\rho)\tilde{C},
  \qquad
  B_\rho = (1-\rho)I + \rho\,\mathbf{1}\mathbf{1}^\top,
\]
where $\tilde{C}$ is symmetric, has zero diagonal, and contains the residual inner products
\[
  \tilde{c}_{ij} = z_i^\top z_j.
\]

We then isolate the shared component by defining
\[
  H := B_\rho^{-1/2}(1-\rho)\tilde{C}\,B_\rho^{-1/2},
\]
so that
\[
  K = B_\rho^{1/2}(I+H)B_\rho^{1/2}.
\]
Taking determinants yields
\[
  \log\det(K) = \log\det(B_\rho) + \log\det(I+H).
\]

Whenever $\|H\|_2 < 1$, we may expand
\[
  \log\det(K)
  =
  \log\det(B_\rho)
  + \operatorname{tr}(H)
  - \frac{1}{2}\operatorname{tr}(H^2)
  + \frac{1}{3}\operatorname{tr}(H^3)
  + \mathcal{O}(\|H\|_2^4).
\]

The first-order term admits a closed-form expression. By cyclicity of the trace,
\[
  \operatorname{tr}(H)
  =
  \operatorname{tr}\!\bigl(B_\rho^{-1}(1-\rho)\tilde{C}\bigr).
\]
Using the Sherman--Morrison formula,
\[
  B_\rho^{-1}
  =
  \frac{1}{1-\rho}I
  -
  \frac{\rho}{(1-\rho)(1+(k-1)\rho)}\mathbf{1}\mathbf{1}^\top,
\]
and therefore
\[
  \operatorname{tr}(H)
  =
  \operatorname{tr}(\tilde{C})
  -
  \frac{\rho}{1+(k-1)\rho}\operatorname{tr}(\mathbf{1}\mathbf{1}^\top \tilde{C}).
\]
Since $\tilde{C}$ has zero diagonal, $\operatorname{tr}(\tilde{C})=0$, and
\[
  \operatorname{tr}(\mathbf{1}\mathbf{1}^\top \tilde{C})
  =
  \sum_{i,j}\tilde{c}_{ij}
  =
  2\sum_{i<j}\tilde{c}_{ij},
\]
it follows that
\[
  \operatorname{tr}(H)
  =
  -\frac{2\rho}{1+(k-1)\rho}\sum_{i<j}\tilde{c}_{ij}.
\]

Substituting this expression into the expansion gives
\[
  \log\det(K)
  =
  \log\det(B_\rho)
  -
  \frac{2\rho}{1+(k-1)\rho}\sum_{i<j}\tilde{c}_{ij}
  -
  \frac{1}{2}\operatorname{tr}(H^2)
  +
  \mathcal{O}(\|H\|_2^3).
\]

This decomposition separates the contribution of the shared structural component, absorbed into $\log\det(B_\rho)$, from that of the
residual similarities. In particular, the leading subset-dependent term is a negative multiple of $\sum_{i<j}\tilde{c}_{ij}$, showing that,
once the shared structure has been factored out, the log-determinant penalizes large residual similarity at first order.

In our setting, the condition $\|H\|_2 < 1$ is empirically satisfied for MDLM generations up to approximately $45\%$ masked tokens,
supporting the validity of this expansion in the regime relevant to our experiments. Taken together, the two approximations above clarify
how DPP MAP selection is linked to cosine-based diversity both in the ideal weak-correlation regime and in the presence of shared
structural-token effects.

\section{Compute Resources and Existing Assets}

\subsection{Compute resources}

We report the compute used for the main inference and large-scale selection experiments. Experiments were run on GPU-accelerated nodes
equipped with NVIDIA A100, L40S, H100, and H200 GPUs. The budgets in Table~\ref{tab:compute_resources} are approximate and exclude
preliminary debugging runs, failed jobs, and small pilot sweeps. We report accelerator-hours, where one accelerator-day corresponds to
$24$ hours on a single GPU.

\begin{table}[H]
  \centering
  \caption{Approximate compute budget used for the main experiments.}
  \label{tab:compute_resources}
  \small
  \begin{tabularx}{\linewidth}{@{}lXlr@{}}
    \toprule
    Experiment & Configuration & Accelerator & Accelerator-hours \\
    \midrule
    Open-ended generation
    & $4$ methods $\times$ $2$ accelerator-days
    & mixed
    & $192$ \\
    E-SMC comparison
    & $3$ accelerator-days
    & L40S
    & $72$ \\
    Question answering
    & $3$ datasets $\times$ $4$ methods $\times$ $20$ accelerator-hours
    & A100
    & $240$ \\
    CFG sweep
    & $5$ CFG values $\times$ $3$ methods $\times$ $20$ accelerator-hours
    & A100
    & $300$ \\
    Mathematical reasoning
    & $4$ methods $\times$ $3$ remasking settings $\times$ $30$ accelerator-hours
    & H100
    & $360$ \\
    Synthetic DPP maximization
    & $1$ accelerator-day
    & H200
    & $24$ \\
    \midrule
    Total
    & --
    & --
    & $1188$ \\
    \bottomrule
  \end{tabularx}
\end{table}

Aggregated by accelerator type, this corresponds to $540$ A100-hours, $72$ L40S-hours, $360$ H100-hours, $24$ H200-hours, and $192$
additional mixed-GPU hours from the open-ended generation experiments, for a total of $1188$ accelerator-hours, or $49.5$ accelerator-days.

\subsection{Use of existing assets}

We used publicly available software, pretrained models, and benchmark datasets. Tables~\ref{tab:software_assets}--\ref{tab:public_assets}
list the main external assets used in the experiments, together with the corresponding public repository or model/dataset card and the
license reported there. We retain the original licenses and attribution notices for all third-party assets.

\begin{table}[H]
  \centering
  \caption{Main software assets used in the implementation and comparisons.}
  \label{tab:software_assets}
  \small
  \begin{tabularx}{\linewidth}{@{}lXl@{}}
    \toprule
    Asset & Link & Reported license \\
    \midrule
    Python
    & \href{https://docs.python.org/3/license.html}{\texttt{python.org}}
    & PSF License \\
    PyTorch
    & \href{https://github.com/pytorch/pytorch}{\texttt{pytorch/pytorch}}
    & BSD-3-Clause \\
    NumPy
    & \href{https://numpy.org/}{\texttt{numpy.org}}
    & BSD-3-Clause / modified BSD \\
    DPPy
    & \href{https://github.com/guilgautier/DPPy}{\texttt{guilgautier/DPPy}}
    & MIT \\
    \bottomrule
  \end{tabularx}
\end{table}

\begin{table}[H]
  \centering
  \caption{Pretrained models (generation and embedding) and datasets used in our experiments.}
  \label{tab:public_assets}
  \small
  \begin{tabularx}{\linewidth}{@{}lXl@{}}
    \toprule
    Model / dataset & Link & Reported license \\
    \midrule
    MDLM
    & \href{https://huggingface.co/kuleshov-group/mdlm-owt}{\texttt{kuleshov-group/mdlm-owt}}
    & Apache-2.0 \\
    LLaDA-8B-Base
    & \href{https://huggingface.co/GSAI-ML/LLaDA-8B-Base}{\texttt{GSAI-ML/LLaDA-8B-Base}}
    & MIT \\
    LLaMA~3 8B
    & \href{https://huggingface.co/meta-llama/Meta-Llama-3-8B}{\texttt{meta-llama/Meta-Llama-3-8B}}
    & Meta Llama~3 Community License \\
    GPT-2
    & \href{https://huggingface.co/openai-community/gpt2}{\texttt{openai-community/gpt2}}
    & MIT / modified MIT \\
    Jina Embeddings v2
    & \href{https://huggingface.co/jinaai/jina-embeddings-v2-base-en}{\texttt{jinaai/jina-embeddings-v2-base-en}}
    & Apache-2.0 \\
    \midrule
    TruthfulQA
    & \href{https://huggingface.co/datasets/truthfulqa/truthful_qa}{\texttt{truthfulqa/truthful\_qa}}
    & Apache-2.0 \\
    ARC-Challenge
    & \href{https://huggingface.co/datasets/allenai/ai2_arc}{\texttt{allenai/ai2\_arc}}
    & CC-BY-SA-4.0 \\
    CommonsenseQA
    & \href{https://huggingface.co/datasets/tau/commonsense_qa}{\texttt{tau/commonsense\_qa}}
    & MIT \\
    GSM8K
    & \href{https://huggingface.co/datasets/openai/gsm8k}{\texttt{openai/gsm8k}}
    & MIT \\
    \bottomrule
  \end{tabularx}
\end{table}

%\newpage
%\input{checklist.tex}

\end{document}